\documentclass[11pt]{article}

\usepackage{amsmath,amsthm} 
\usepackage{amssymb,mathrsfs} 
\usepackage{hyperref}
\usepackage{graphicx} 
\usepackage{color}
\usepackage{enumerate}  
\usepackage{fullpage}
\usepackage{subcaption}
\usepackage{todonotes}

\newtheorem{theorem}{Theorem}
\newtheorem{lemma}{Lemma}
\newtheorem{assumption}{Assumption}
\newtheorem{proposition}{Proposition}

\newtheorem{remark}[theorem]{Remark}

\newcommand{\dps}{\displaystyle } 
\newcommand{\rme}{\mathrm{e}} 
\newcommand{\eps}{\varepsilon} 
\newcommand{\R}{\mathbb{R}}

\newcommand{\cL}{\mathcal{L}}

\renewcommand{\leq}{\leqslant}
\renewcommand{\geq}{\geqslant}

\newcommand{\Tr}{\operatorname{Tr}}
\newcommand{\tht}{\theta}
\newcommand{\cLham}{\mathcal{L}_{\rm ham}}
\newcommand{\cLFD}{\mathcal{L}_{\rm FD}}
\newcommand{\cLa}{\mathcal{L}_{\rm 1}}
\newcommand{\cLb}{\mathcal{L}_{\rm 2}}
\newcommand{\cLc}{\mathcal{L}_{\rm 3}}
\newcommand{\cLd}{\mathcal{L}_{\rm 4}}
\newcommand{\Norm}{\mathcal{N}}
\newcommand{\var}{\mathrm{var}}  
\newcommand{\cov}{\mathrm{cov}}  
\newcommand{\Nd}{N_{\mathrm{data}}}  
\newcommand{\Ni}{N_{\mathrm{iter}}}  
\newcommand{\vectx}{\textbf{x}}  
\newcommand{\vecty}{\textbf{y}}
\newcommand{\vectz}{\textbf{z}}  
\newcommand{\PP}{ P_{\rm prior}}  
\newcommand{\PL}{P_\mathrm{likelihood}}  
\newcommand{\PLe}{P_{\mathrm{elem}}}  
  
\newcommand{\gradT}{\nabla_\tht}

\newcommand{\dt}{\Delta t}

\begin{document}

\title{Quantifying the mini-batching error in Bayesian inference for Adaptive Langevin dynamics}


\author{I. Sekkat$^{1}$ and G. Stoltz$^{1,2}$ \\
	{\small $^{1}$ CERMICS, Ecole des Ponts, Marne-la-Vall\'ee, France} \\
	{\small $^{2}$ MATHERIALS team-project, Inria Paris, France} \\
}
\maketitle

\abstract{
Bayesian inference allows to obtain useful information on the parameters of models, either in computational statistics or more recently in the context of Bayesian Neural Networks.
The computational cost of usual Monte Carlo methods for sampling posterior laws in Bayesian inference
scales linearly with the number of data points. One option to reduce it to a fraction of this cost
is to resort to mini-batching in conjunction with unadjusted discretizations of Langevin dynamics, in which case
only a random fraction of the data is used to estimate the gradient. However, this leads to an additional noise
in the dynamics and hence a bias on the invariant measure which is sampled by the Markov chain. We advocate
using the so-called Adaptive Langevin dynamics, which is a modification of standard inertial Langevin dynamics
with a dynamical friction which automatically corrects for the increased noise arising from mini-batching. We
investigate the practical relevance of the assumptions underpinning Adaptive Langevin (constant
covariance for the estimation of the gradient, Gaussian minibatching noise), which are not satisfied in typical models of Bayesian inference, and quantify the bias induced by minibatching in this case. We also suggest a possible extension of AdL to further reduce the bias on the posterior distribution, by considering a dynamical friction depending on the current value of the parameter to sample.
}

\section{Introduction}

Bayesian modeling allows to determine the distribution of parameters in statistical models, and hence to estimate functions of these parameters and their uncertainties. The Bayesian approach to neural networks in particular has recently gained some attention from the Machine Learning community; see for instance~\cite{WY20,BNN} and references therein. Since the distributions of parameters are given by (possibly very) high dimensional probability measures, Markov Chain Monte Carlo (MCMC) techniques (see~\cite{RC04,MR2742422}) are the default methods to sample these target measures. In this work, we restrict ourselves to probability distributions having a density with respect to the Lebesgue measure. Quantities of interest are then expectations with respect to the target distribution, which are approximated by Monte Carlo estimates based on ergodicity results for the Markov chains under consideration. Let us however mention, that, beyond sampling from distributions of parameters, MCMC methods can also be used for optimization when run at small target temperature, which may be beneficial to explore complex non-convex energy landscapes (see for instance~\cite{MCJFJ19,LMV19,DDB20}), or to ensure some form of regularization in an attempt to avoid overfitting (as already noted in~\cite{Welling:2011:BLV:3104482.3104568}).

Two major classes of MCMC techniques can be distinguished. The first one gathers methods based on the Metropolis--Hastings algorithm (see~\cite{MRRTT53, Hastings70}). The second class of MCMC techniques relies on discretizations of stochastic differential equations (SDEs) which are ergodic with respect to the target measure. The SDEs used to sample from a probability distribution were originally introduced in molecular dynamics, where atomic configurations in a system are typically distributed according to the Boltzmann--Gibbs distribution $\pi(d\tht) = Z^{-1}\exp\left(- V(\tht)\right)\,d\tht$, with~$Z$ the normalization constant and~$V$ the potential energy function of the system; see~\cite{FrenkelSmit,Tuckerman,29acd3d494044594aea0829ef236aad6,AT17}. Two prominent examples of such dynamics are Langevin dynamics (which we consider in this work as being the underdamped, or kinetic version) and its overdamped limit. In practice, methods from the two classes can be blended, as for the Metropolis-adjusted Langevin algorithm (see~\cite{bj/1178291835, RDF78, MR1625691}), which is a Metropolis--Hastings algorithm whose proposal function is provided by a Euler--Maruyama discretization of overdamped Langevin dynamics. We focus in this work on the second class of techniques, namely the discretization of SDEs. Langevin-like dynamics are gaining increasing popularity in Machine Learning and related application fields (see for instance~\cite{pmlr-v32-cheni14,MMWBJ21,GGHZ21} to quote just a few works).

Both Metropolis--Hastings methods and discretization of SDEs can be computationally expensive in the context of Bayesian inference since the log-likelihood and/or its gradient have to be evaluated at each step, either to perform a Metropolis test or to compute the forces in the dynamics. The cost of computing the log-likelihood and its gradient scales as~$\mathcal{O}(\Nd)$, where $\Nd$ is the size of the data set, which may be large. Some variations of Metropolis-like algorithms, based on estimates of the log-likelihood obtained from a random subsample of the data, have been introduced to reduce the computational cost of one iteration; see for example~\cite{Korattikara:2014:AML:3044805.3044827,bardenet:hal-01254232,MR3963184}. These methods however require some prior knowledge on the target measure, and/or introduce some bias on the measure actually sampled by the algorithm. A similar approach was proposed for discretizations of SDEs by~\cite{Welling:2011:BLV:3104482.3104568}, who suggested to use a mini-batch of the data to construct an estimator of the gradient of the log-likelihood, leading to the so-called Stochastic Gradient Langevin Dynamics (SGLD).

SGLD however also induces biases on the sampled invariant measure, which have two origins: the finiteness of the time step and the fact that the mini-batch size~$n$ is smaller than~$\Nd$. It is possible to remove the bias by using decreasing time steps, as initially suggested in~\cite{Welling:2011:BLV:3104482.3104568} and mathematically analyzed in~\cite{MR3482927}, but this is not practical for the convergence of longtime averages, in particular when there is some metastability in the system (\emph{i.e.} in situations when the posterior probability measure is multimodal, and the numerical methods remain temporarily stuck in one of the modes before finding their way to another one. Multimodality is generic in Bayesian inference as soon as some parameters are exchangeable. It can be argued that numerical methods which are able to switch between equivalent modes have good exploration properties, which allows them to find possibly unanticipated local modes; see~\cite{chopin-lelievre-stoltz-12}). An analysis of the asymptotic bias of SGLD for fixed step sizes is provided in~\cite{JMLR:v17:15-494} (together with some analysis of the non-asymptotic bias, following the approach developed in~\cite{Mattingly908convergenceof}). The results show that SGLD has a weak error of order one, similarly to the Euler-Maruyama discretization of overdamped Langevin dynamics, but with an extra term in the leading error arising from the mini-batching procedure. In fact, the magnitude of this term makes it the dominant one, unless~$n$ is very close to~$\Nd$.

Various extensions and refinements of SGLD were proposed, in an attempt to improve the performance of the method and/or to reduce its bias. A first trend is to apply the mini-batching philosophy to dynamics which are more efficient in terms of sampling than overdamped Langevin dynamics, in particular Langevin dynamics (see~\cite{2018arXiv180508863M}), Hybrid Monte Carlo algorithms (see~\cite{pmlr-v32-cheni14}) and piecewise deterministric Markov processes (see~\cite{PGCP17,BFR19}). A second trend is to rely on control variate techniques to reduce the covariance of the stochastic estimator of the gradient of the log-likelihood, computed in practice by Gaussian approximations of the modes of the target probability measure; see for instance~\cite{articleN17, brosse:hal-01934291, MR3969063}. Of course, both trends can (should) be combined.

Our emphasis in this work is on the adaptive Langevin dynamics (AdL), introduced in~\cite{PMID:21895177, NIPS2014_5592}, which provides a way to reduce, and hopefully even almost remove the bias arising from mini-batching in SGLD. This dynamics is a modification of Langevin dynamics where the friction is considered as a dynamical variable that adjusts itself so that the distribution of the velocities is correct. In addition to correcting for the mini-batching bias, it can also be used to train neural networks; see~\cite{LMV19}. AdL was mathematically studied in~\cite{MR4099815}, and further tested from a numerical viewpoint in~\cite{Shang2015}. The method completely removes the mini-batching bias when the covariance matrix~$\Sigma_\vectx(\tht)$ of the estimator of the gradient is constant in the range of parameters explored by the method. This assumption is satisfied for Gaussian posterior distributions, as obtained when considering a Gaussian prior and Gaussian likelihoods, or when the number of data points is large enough and the posterior distribution concentrates around a Gaussian distribution according to the Bernstein--von Mises theorem (see for instance Section~$10.2$ in~\cite{vdV98}). There are however situations where this assumption does not hold (as we highlight on a numerical example in Section~\ref{mix_adl}), in which case AdL fails to fully correct for the bias arising from mini-batching. 

In this paper, we consider the case of extreme mini-batching procedures, corresponding to~$n$ as small as~1, as in initial works on stochastic approximation in~\cite{Robbins&Monro:1951}. In this situation, no central limit theorem can be invoked to precisely characterize the statistical properties of the stochastic gradient estimator. This is in contrast with various works which assume the mini-batching noise to be Gaussian. This is however not needed to make precise the weak error of numerical methods, and hence the bias on the invariant probability measure, as we show in our estimates on the bias for SGLD and Langevin dynamics with stochastic gradient estimators. One of our contributions is to precisely quantify how the bias on the invariant measure depends on the type (with or without replacement) and amount of mini-batching, through some key parameter~$\eps(n)$ defined in Section~\ref{sec:minbatch_procedure}. 
We also carefully study the covariance matrix of the stochastic gradient estimator in illustrative numerical examples, highlighting that this matrix may exhibit substantial variations in the range of parameters explored by the dynamics under consideration, so that the key assumption underlying AdL is not satisfied in general. Nonetheless, our numerical analysis explains why AdL still succeeds to substantially reduce the bias compared to SGLD and Langevin dynamics. We finally introduce an extended version of AdL allowing to even further reduce the bias incurred by non constant covariance matrices.

One of our main contributions is to prove that, for all the Langevin-type SDEs considered in this paper (standard Langevin, adaptive Langevin, extended adaptive Langevin), the bias introduced by the minibatching procedure is controlled by
\begin{equation}
\eps(n) \dt \min_{S \in \mathscr{S}} \| \Sigma_\vectx - S\|_{L^2(\pi)} = \eps(n) \dt \| \Sigma_\vectx - S^* \|_{L^2(\pi)},
\label{eq:erreur_gen}
\end{equation}
where~$\dt$ is the time step used to discretize the SDE at hand, and~$\pi$ is the target posterior distribution. The vector space of matrix valued functions~$\mathscr{S}$ depends on the chosen dynamics. For discretizations of standard Langevin dynamics, $S^* = 0$ (see Sections~\ref{sec:eff_SGLD} and~\ref{sec:effective_Langevin_dynamics}, in particular~\eqref{eq:f_mb_Langevin}), whereas $S^*$ corresponds to the average of~$\Sigma_\vectx$ with respect to~$\pi$ for AdL (see Section~\ref{sec:bias_adl}, in particular~\eqref{eq:f_mb_AdL} and Remark~\ref{rem:adl_xi_cases}) and to the $L^2(\pi)$-projection of~$\Sigma_\vectx$ onto the vector space of symmetric matrices generated by some basis of functions for the extended version of AdL we introduce (see Section~\ref{sec:disEADL}, in particular~\eqref{eq:proj_sigma}). We numerically verify that the reduction in the bias is proportional to the quality of the approximation of the covariance matrix on a basis of functions (\emph{e.g.} constant matrices, piecewise constant scalar functions, ...). 

The paper is organized as follows. We start in Section~\ref{sec:sto_grad_MCMC} by reviewing various results related to the numerical analysis of SDEs and the quantification of the bias on the invariant measure sampled by SGLD and Langevin dynamics with stochastic gradient estimators. We next turn to AdL in Section~\ref{sec:AdL}: we illustrate that some residual bias remains present due to the fact that the covariance of the gradient estimator is not constant in general, and we quantify it. Section~\ref{sec:eAdL} is dedicated to the introduction of an extended version of AdL that allows to accomodate non constant covariance matrices for the gradient estimator and further reduce the bias. 
Some conclusions and perspectives are gathered in Section~\ref{sec:discussion_perspectives}.

\section{Stochastic gradient Markov Chain Monte Carlo}
\label{sec:sto_grad_MCMC}

We consider a Bayesian inference problem where we denote by~$\vectx = (x_i)_{i=1, ..., \Nd} \in \mathcal{X}^{\Nd}$ a set of~$\Nd$ data points, with~$\mathcal{X} \subset \R^{d_{\mathrm{data}}}$. The data points are assumed to be independent and identically distributed (i.i.d.) with respect to an elementary likelihood probability measure~$\PLe(\cdot|\tht)$, parameterized by $\tht \in \Theta = \R^d$. The likelihood of~$\vectx$ is then given by
\[\PL(\vectx|\tht) = \prod_{i=1}^{\Nd}  \PLe(x_i|\tht).\]
In the Bayesian framework, a prior distribution~$\PP$ is considered on the parameters. The aim is to sample from the posterior distribution 
\begin{equation}
\pi(\tht|\vectx)\propto \PP(\tht)\PL(\vectx|\tht),
\label{pi_bayes}
\end{equation} 
in order, for instance, to compute expectations with respect to this distribution. Sampling is usually done with MCMC methods. This however requires, at each iteration, to compute either the log-likelihood~$\pi(\cdot|\vectx)$ or its gradient
\begin{equation}
\nabla_\tht (\log \pi(\tht|\vectx)) = \nabla_\tht (\log \PP(\tht)) + \sum\limits_{i=1}^{\Nd}  \nabla_\tht( \log \PLe(x_i|\tht)) .
\label{grad}
\end{equation}
The cost of computing these quantities scales as~$\mathcal{O}(\Nd)$, and is usually the computational bottleneck in implementations of MCMC algorithms. As discussed in the introduction, we focus here on stochastic gradient dynamics, which are discretizations of stochastic dynamics admitting~$\pi(\cdot|\vectx)$ as invariant probability measure, and where the cost of evaluating~\eqref{grad} is reduced by approximating the gradient through some estimator based on a mini-batch of the complete data set. 

We first review in Section~\ref{sec:Review_error_analysis} elements on the error analysis of discretization of SDEs, with some emphasis on the bias induced on the invariant measure, as these results will be repeatedly used throughout this paper, and are key to understanding the performance and limitations of all the methods we discuss. We then recall in Section~\ref{sec:minbatch_procedure} the mini-batching method introduced in~\cite{Welling:2011:BLV:3104482.3104568} and focus on the case where the size of the mini-batch is small, possibly limited to a single point (as done in stochastic approximation algorithms in~\cite{Robbins&Monro:1951}). We then recall two classical SDEs upon which our method is based: overdamped Langevin dynamics in Section~\ref{sec:SGLD} (known as SGLD when using mini-batching), and then Langevin dynamics in Section~\ref{sec:Langevin}. In each of these sections, we analyze how mini-batching affects the posterior distribution and quantify the bias incurred on it by studying the associated effective dynamics. These results are illustrated by numerical examples in Section~\ref{sec:numEff}.


\subsection{Some elements on error analysis for discretizations of SDEs}
\label{sec:Review_error_analysis}

Consider a SDE~$(\theta_t)_{t \geq 0} \subset \Theta$ with generator~$\cL$ admitting~$\pi(\cdot|\vectx)$ as invariant probability measure: For any smooth function~$\phi$ with compact support, 
\[
\int_\Theta \cL \phi \, d\pi(\cdot|\vectx) = 0.
\]
Denote by~$(\theta^m)_{m \geq 0}$ a time discretization of the SDE with a fixed time step~$\Delta t$ (so that~$\theta^m$ is an approximation of~$\theta_{m \Delta t}$). We assume that the Markov chain corresponding to the time discretization of the SDE admits a unique invariant probability measure, denoted by~$\pi_{\Delta t}$. This is for instance the case for Langevin-type dynamics when the drift of the dynamics is globally Lipschitz or when Lyapunov conditions are satisfied; see~\cite{MR1931266}. For a given observable~$\phi$, the target expectation
\[
\mathbb{E}_\pi(\phi) = \int_{\Theta} \phi(\tht) \pi(\tht|\vectx)\,d  \tht
\]
is approximated by~$\mathbb{E}_{\pi_{\Delta t } }(\phi)$, which is itself typically estimated by the trajectory average 
\[
\widehat{\phi}_{\dt, \Ni} = \frac{1}{\Ni} \sum\limits_{m=1}^{\Ni} \phi(\tht^m) .
\] 
The total error on averages with respect to $\pi(\cdot |\vectx)$ can then be written as:
\[
\widehat{\phi}_{\dt, \Ni}  -  \mathbb{E}_\pi(\phi)   = (\mathbb{E}_{\pi_{\Delta t } }(\phi) - \mathbb{E}_\pi (\phi) ) + \left(\widehat{\phi}_{\dt, \Ni} - \mathbb{E}_{\pi_{\Delta t } }(\phi) \right)  .
\]
The first term on the right hand side corresponds to the bias on the invariant probability measure resulting from taking finite step sizes. The second term in the error has two origins: (i) a bias coming from the initial distribution of $\tht^0$ when this random variable is not distributed according to~$\pi_{\dt}$; (ii) a statistical error, which is dictated by the central limit theorem for~$\Ni$ large.

We focus in this work on the bias on the invariant probability measure, which can be bounded using the weak order of the scheme, provided some ergodicity conditions are satisfied. Recall that a numerical scheme is of weak order~$s$ if for any smooth and compactly supported function~$\phi$ and final time $T > 0$, there exists~$C \in \R_+$ such that
\begin{equation}
\label{eq:weak_error}
\forall m \in \{1,\dots,\lceil T/\dt \rceil\},
\qquad
\left|\mathbb{E}[\phi(\tht_{m \Delta t} )] - \mathbb{E}[\phi(\tht^{m})]\right| \leq C \Delta t^s.
\end{equation}
When this condition holds, and under appropriate ergodicity conditions (see~\cite{talay-tubaro-90} for a pioneering work, as well as~\cite{Talay02,MR1931266,MR3229658,MR3463433, MR2608370} for subsequent works on Langevin-like dynamics), the following bound is obtained on the bias on the invariant probability measure of the numerical scheme: For any smooth and compactly supported function~$\phi$, there exist~$\dt_\star>0$ and~$L$ such that 
\begin{equation}
\label{eq:error_invariant_measure_general}
\forall \dt \in (0,\dt_\star],
\qquad
\left| \mathbb{E}_{\pi_{\Delta t } }(\phi) - \mathbb{E}_\pi (\phi) \right| \leq L \Delta t^s.
\end{equation}
In order to write a sufficient local consistency condition to obtain an estimate such as~\eqref{eq:weak_error}, we introduce the evolution operator~$P_{\Delta t }$ associated with the numerical scheme at hand, defined as follows: For any smooth and compactly supported function~$\phi$,
\[
\left(P_{\Delta t}\phi\right)(\tht) = \mathbb{E}\left[\phi\left(\tht^{m+1}\right) \, \middle| \, \tht^m= \tht\right].
\]
Under appropriate technical conditions, including moment conditions on the iterates of the numerical scheme (see Theorem~2.1 in~\cite{milstein2013stochastic} for a precise statement), a sufficient condition for~\eqref{eq:weak_error} to hold is 
\begin{equation}
\label{eq:local_consistency_general}
P_{\dt} =  \mathrm{e}^{\Delta t\cL}+ \mathcal{O}(\Delta t ^{s+1}),
\end{equation}
where~$(\mathrm{e}^{t\cL}\phi)(\theta) = \mathbb{E}[\phi(\theta_t) \, | \, \theta_0=\theta]$ is the evolution operator associated with the underlying SDE. Here and in the sequel, the above equality has to be understood as follows: For any smooth and compactly supported function~$\phi$, there exist~$\dt_\star>0$ and~$K \in \mathbb{R}_+$ such that, for any~$\dt \in (0,\dt_\star]$, there is a function~$R_{\phi,\dt}$ for which 
\begin{equation}
  \label{eq:write_remainders_rigorously}
  P_{\dt}\phi =  \mathrm{e}^{\Delta t\cL}\phi + \Delta t^{s+1} R_{\phi,\dt},
  \qquad
  \sup_{\dt \in (0,\dt_\star]} \sup_{\theta \in \Theta} |R_{\phi,\dt}(\theta)| \leq K;
\end{equation}
see for instance Section~3.3 in~\cite{lelievre_stoltz_2016} for a more precise discussion of this point.

Let us conclude this section by introducing the concept of effective dynamics (also called modified SDEs in~\cite{Shardlow06,MR2783188}), a tool inspired by backward numerical analysis. The idea is to construct a continuous dynamics, parameterized by the time step, which better coincides with the numerical scheme than the reference dynamics. More precisely, given a numerical scheme of weak order~$s$, characterized by the local consistency condition~\eqref{eq:local_consistency_general}, we look for a modified SDE with generator~$\cL_{\mathrm{mod}, \dt}$ such that 
\begin{equation}
\label{eq:evolution_op_modified_dynamics}
P_{\dt} - \mathrm{e}^{\dt \cL_{\mathrm{mod}, \dt}} =  \mathcal{O}(\dt^{s+2}).
\end{equation}
An alternative reformulation is the following: denoting by~$(\widetilde{\theta})_{t \geq 0}$ the solution to the SDE with generator~$\cL_\mathrm{mod, \dt}$, then
\[
\mathbb{E}[\phi(\theta^1) \, | \, \theta_0=\theta] = \mathbb{E}[\phi(\widetilde{\theta}_\dt) \, | \, \theta_0=\theta] + \mathcal{O}(\dt^{s+2}),
\]
while
\[
\mathbb{E}[\phi(\theta^1) \, | \, \theta_0=\theta] = \mathbb{E}[\phi(\theta_\dt) \, | \, \theta_0=\theta] + \mathcal{O}(\dt^{s+1}).
\]
The prime interest of effective dynamics in our work is to provide some interpretation of the behavior of numerical schemes. Following the standard philosophy of backward analysis, it is indeed possible to obtain information on the behavior of numerical schemes by studying the properties of the effective dynamics, in particular their invariant probability measures. This is also useful for instance to understand the impact of mini-batching on SGLD and Langevin dynamics (see respectively~\cite{JMLR:v17:15-494} and Section~\ref{sec:Langevin}), and is also the foundation of Adaptive Langevin dynamics. In particular, when~\eqref{eq:evolution_op_modified_dynamics} holds with~$\cL_{\mathrm{mod}, \dt} = \cL + \dt^s \mathcal{A}_{\rm mod}$, then~\eqref{eq:error_invariant_measure_general} can be rewritten more precisely in the form of a Talay--Tubaro expansion (see~\cite{talay-tubaro-90}):
\begin{equation}
\label{eq:Talay_Tubaro_like}
\forall \dt \in (0,\dt_\star],
\qquad
\left| \mathbb{E}_{\pi_{\Delta t } }(\phi) - \mathbb{E}_\pi (\phi) - \dt^s \mathbb{E}_\pi(f\phi) \right| \leq \widetilde{L} \Delta t^{s+1},
\end{equation}
with $f = \left(-\cL^*\right)^{-1}\mathcal{A}_\mathrm{mod}^* \mathbf{1}$, adjoints being taken on~$L^2(\pi)$ (provided the inverse of~$\mathcal{L}$ can be defined on an appropriate subspace of~$L^2(\pi)$, see again Section~3.3 in~\cite{lelievre_stoltz_2016}). The formula~\eqref{eq:Talay_Tubaro_like} allows to make precise the dominant term in the error~$\mathbb{E}_{\pi_{\Delta t } }(\phi) - \mathbb{E}_\pi (\phi)$.


\subsection{Mini-Batching procedure}
\label{sec:minbatch_procedure}

The idea of using mini-batching in the context of SDEs has first been introduced through SGLD in~\cite{Welling:2011:BLV:3104482.3104568} to reduce the calculation time of one step of the discretization of the overdamped Langevin dynamics (see~\eqref{EulerM} below) from~$\mathcal{O}(\Nd)$ to a fraction of this cost. It is inspired from the classical stochastic gradient descent method in~\cite{Robbins&Monro:1951} in the sense that a consistent approximation of~$\gradT ( \log \pi(\cdot| \vectx))$ is used in the discretization of overdamped Langevin dynamics; see Section~\ref{sec:SGLD} below.

Mini-batching relies on computing at each iteration an approximation of the gradient of the log-likelihood using a random subset of size~$n$ of the data points, which reduces the cost from~$\mathcal{O}(\Nd)$ to~$\mathcal{O}(n)$. More precisely, the gradient of the posterior distribution is approximated by
\begin{equation}
\widehat{F_n}(\tht) =  \gradT (\log \PP(\tht) )+ \frac{\Nd}{n} \sum\limits_{i\in I_n}  \gradT (\log \PLe(x_i|\tht)),
\label{eq:gradSto}
\end{equation}
where~$I_n$ is a random subset of size $n$ generated by sampling uniformly indices from $\{1,..., \Nd \}$, with or without replacement. Sampling with or without replacement produces similar results when $n \ll \Nd$ with~$\Nd$ large enough. Sampling with replacement however leads to estimators~\eqref{eq:gradSto} with larger variances for a given value of~$n$ (as quantified by~\eqref{eq:eps} below), and should therefore be avoided when larger batches are considered. In any case, it is easily shown that~$\widehat{F_n}(\tht)$ in~\eqref{eq:gradSto} is a consistent approximation of~$\gradT( \log \pi(\cdot | \vectx))$ given by~\eqref{grad}.

\paragraph{Properties of $\widehat{F_n}(\tht)$.} In view of~\eqref{eq:gradSto}, the covariance matrix of the stochastic gradient, which is nonnegative and symmetric, is given by:
\begin{equation}
\cov\left(\widehat{F_n}(\tht) \right) = \eps(n) \Sigma_\vectx(\tht),
\label{eq:sigmadef}
\end{equation}
with $\Sigma_\vectx (\tht)$ the empirical covariance of the gradient estimator for~$n=1$ (\emph{i.e.} with expectations computed with respect to the random variable~$\mathcal{I}$ uniformly distributed in~$\{1, ..., \Nd\}$):
\begin{equation}
\begin{aligned}
\Sigma_\vectx (\tht) &= \cov_{\mathcal{I}}\left[\gradT (\log \PLe(x_{\mathcal{I}}|\tht))\right]\\
&= \frac{1}{\Nd-1}\sum\limits_{i =1}^{\Nd }\left[ \gradT (\log\PLe(x_i|\tht)) - \mathscr{F}_\vectx(\tht) \right]\left[\gradT (\log \PLe(x_i|\tht)) - \mathscr{F}_\vectx(\tht)\right]^T,
\label{eq:sigma_somme}
\end{aligned}
\end{equation}
where~$uv^T$ is the matrix~$(u_i v_j)_{1 \leq i,j \leq d} \in \R^{d \times d}$ for~$u,v \in \R^d$, the average force reads
\[
\mathscr{F}_\vectx(\tht) = \frac{1}{\Nd}\sum\limits_{i =1}^{\Nd }\gradT (\log \PLe(x_i|\tht)),
\]
and
\begin{equation}
\eps(n) =
\left\{ \begin{aligned}
& \frac{\Nd(\Nd-1)}{n}, \quad &&  \mathrm{for \ sampling \  with \ replacement,} \\
&  \frac{\Nd (\Nd-n)}{n}, \quad  &&\mathrm{for \ sampling \  without \ replacement.} 
\end{aligned} \right.
\label{eq:eps}
\end{equation}
We refer for instance to~\cite{chaudhari2018stochastic} for more details. We will see in the next sections that, for all the methods we consider, the bias due to mini-batching is controlled by~$\eps(n)$ whether sampling is performed with our without replacement. It is really this parameter (in fact, $\eps(n)\Delta t$) which needs to be small in order for the asymptotic analysis on the bias to be correct. Note that~$\eps(n) \geq \Nd-1$ for sampling with replacement, which means that a bias is necessarily observed in this case. Values of~$\eps(n)$ of order~$\Nd$ are obtained only when the mini-batch size~$n$ is a fraction of the total number of data points~$\Nd$. To obtain values of order~1, sampling has to be done without replacement with~$n$ close to~$\Nd$. In contrast, $\eps(n)$ is of order~$\Nd^2$ when the mini-batch size is small, whether sampling is performed with or without replacement.

\begin{figure}
	\centering
	\begin{minipage}{0.31\textwidth}
		\centering
		\includegraphics[width=\textwidth]{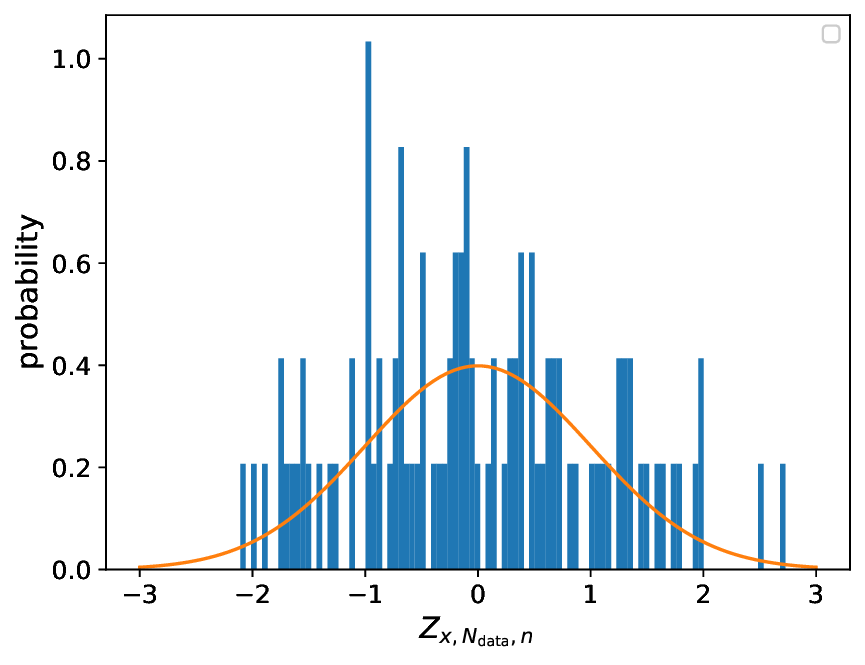}
		\subcaption[first caption.]{$n=1$}
	\end{minipage}
	\begin{minipage}{0.31\textwidth}
		\centering
		\includegraphics[width=\textwidth]{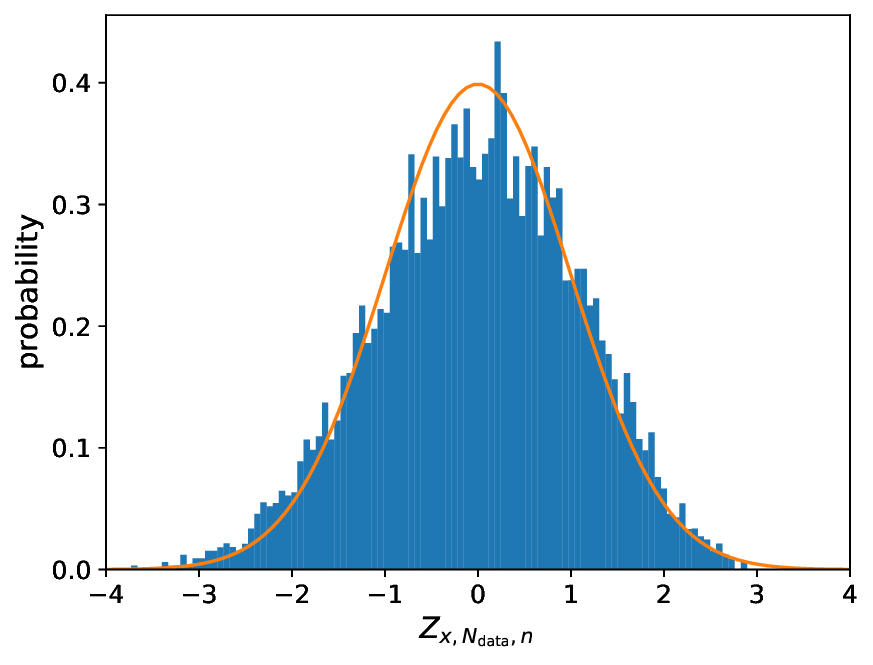}
		\subcaption[second caption.]{$n=2$}
	\end{minipage}
	\begin{minipage}{0.31\textwidth}
		\centering
		\includegraphics[width=\textwidth]{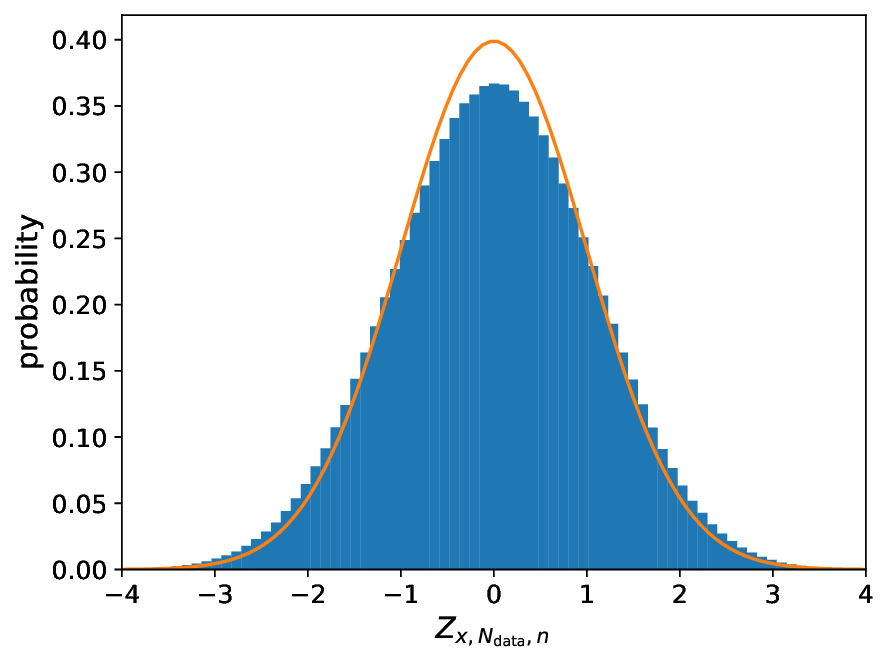}
		\subcaption[first caption.]{$n=30$}
	\end{minipage}
	\caption{Histogram of $ Z_{\vectx, \Nd,  n}$ when sampling is performed without replacement for the Gaussian model of Section~\ref{sec:gaussian_model}, with $\Nd=100$, $\sigma_\theta = 1$, $\sigma_x = 1$, $\mu = 0$, $\tht = 0.5$. The reference standard Gaussian distribution is superimposed as a continuous line.} \label{fig:Z_x_gaussian} 
\end{figure}

\begin{figure}
	\centering
	\begin{minipage}{0.45\textwidth}
		\centering
		\includegraphics[width=\textwidth]{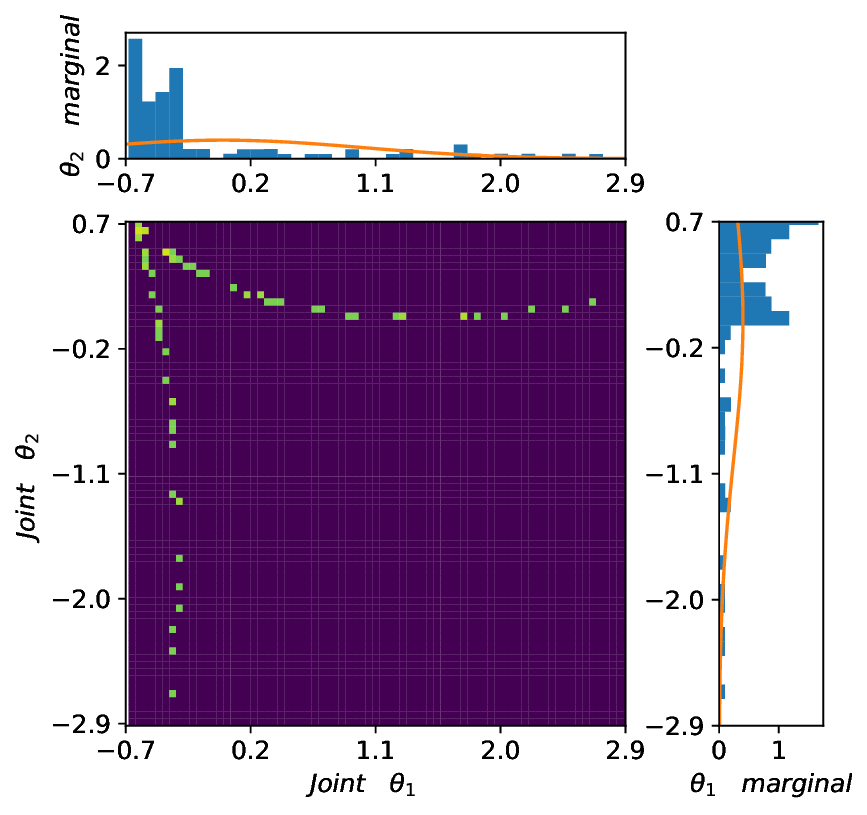}
		\subcaption[first caption.]{$n=1$}
	\end{minipage}
	\begin{minipage}{0.45\textwidth}
		\centering
		\includegraphics[width=\textwidth]{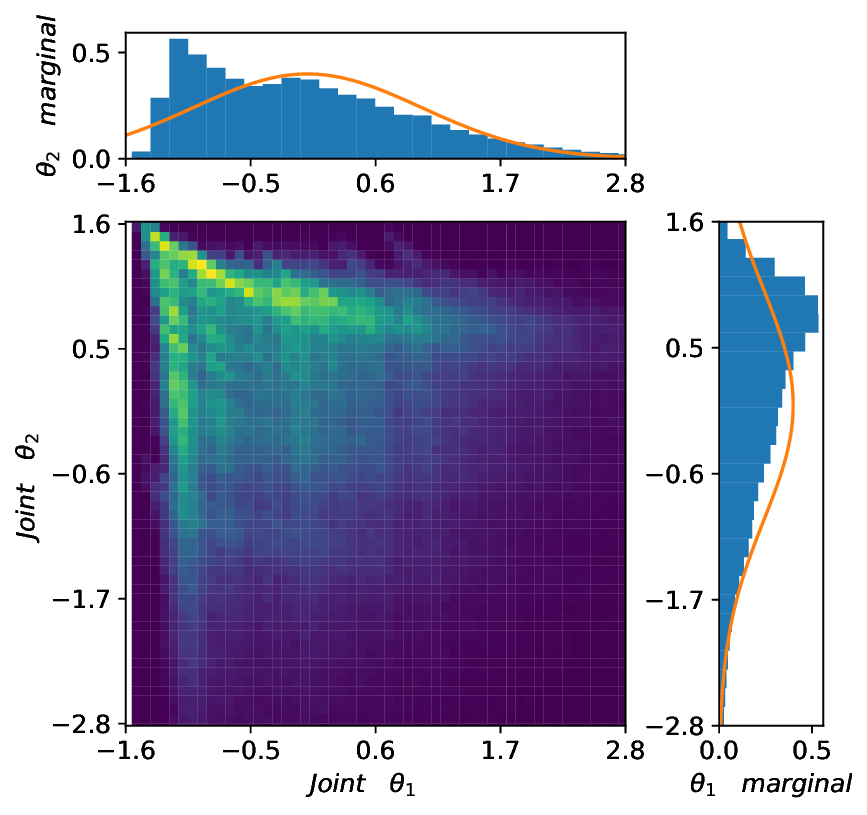}
		\subcaption[second caption.]{$n=5$}
	\end{minipage}
	\begin{minipage}{0.45\textwidth}
		\centering
		\includegraphics[width=\textwidth]{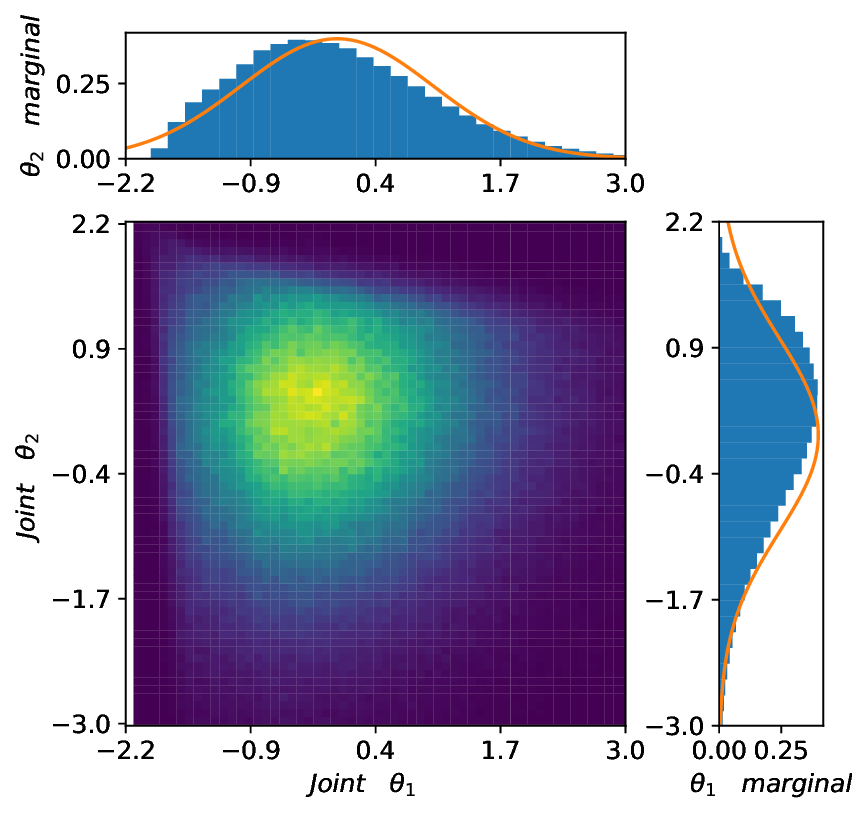}
		\subcaption[first caption.]{$n=10$}
	\end{minipage}
	\begin{minipage}{0.45\textwidth}
		\centering
		\includegraphics[width=\textwidth]{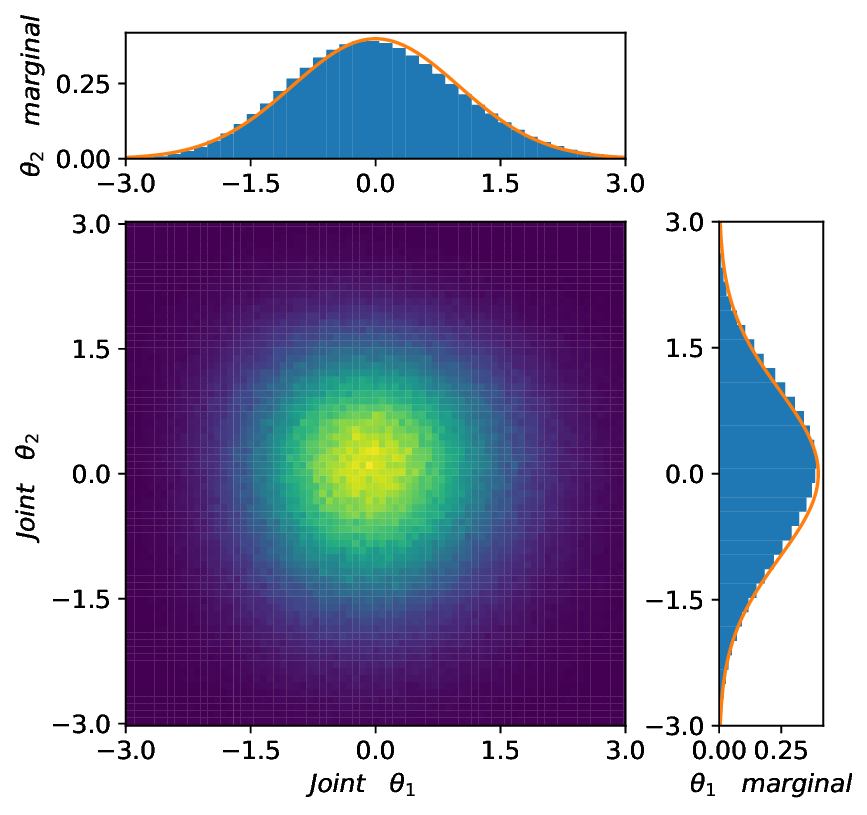}
		\subcaption[first caption.]{$n=30$}
	\end{minipage}
	\caption{Histogram of $Z_{\vectx, \Nd,  n}$ for~$\theta = (1,0.2)$, when sampling is performed without replacement for the model of Section~\ref{sec:mixture_model} with $\Nd=100$ data points sampled from a mixture of Gaussians with parameters~$\sigma_1 = \sigma_2 = 0.4$, $w = 0.5$, $\tht_1 = 0.2$ and $\tht_2 = 1$. The reference standard Gaussian distribution is superimposed as a continuous line on the marginal distributions in~$\tht_1,\tht_2$.} \label{fig:Z_x_mixture}
\end{figure}

The above definitions allow to rewrite~\eqref{eq:gradSto} as
\begin{equation}
\widehat{F_n}(\tht) =\gradT (\log\pi(\tht|\vectx)) + \sqrt{\eps(n)}\Sigma_\vectx^{\frac{1}{2}} (\tht)Z_{\vectx, \Nd,  n} ,
\label{eq:hypothese}
\end{equation}
where~$Z_{\vectx, \Nd,  n}$ is by construction a centered random variable with identity covariance. If~$n$ and~$\Nd$ are large enough, and~$n\ll \Nd$, the central limit theorem holds so that~$Z_{\vectx, \Nd,  n} $ asymptotically follows a centered reduced normal distribution (as assumed for instance in~\cite{pmlr-v32-cheni14, 10.5555/3042573.3042799, pmlr-v97-zhu19e}). Most of the works in the literature consider the regime where the central limit theorem holds. This is however unnecessary when it comes to quantifying the weak error of the numerical scheme and hence the error on the invariant measure. In our work, we consider explicitly the case when~$n$ is of order~$1$ and~$\Nd$ is (moderately) large, so that $Z_{\vectx, \Nd,  n}$ is not necessarily close to a Gaussian distribution. This is illustrated in Figures~\ref{fig:Z_x_gaussian} and~\ref{fig:Z_x_mixture}, where we plot the distribution of~$Z_{\vectx, \Nd,  n}$ for various values of~$n$ for the models introduced in Section~\ref{sec:numEff}, where the elementary likelihood is either a Gaussian or a mixture of Gaussians. It is clear that the distribution of~$Z_{\vectx, \Nd,  n}$ is far from being Gaussian for small values of~$n$, but is nonetheless centered and with covariance~$\mathrm{I}_d$. Distributions close to Gaussian are obtained for~$n=20-30$ for the situation considered in Figure~\ref{fig:Z_x_mixture}.

\subsection{Stochastic Gradient Langevin Dynamics}
\label{sec:SGLD}

We introduce in this section the SGLD algorithm (see Section~\ref{sec:description_SGLD}) and analyze the bias due to replacing the gradient~\eqref{grad} by the stochastic estimator~\eqref{eq:gradSto} using the effective dynamics associated with the numerical method at hand (see Section~\ref{sec:eff_SGLD}).

\subsubsection{Description of the method}
\label{sec:description_SGLD}

A popular SDE to sample from $\pi(\cdot |\vectx)$ is the overdamped Langevin dynamics:
\begin{equation}
d\tht_t =  \gradT (\log \pi(\tht_t|\vectx)) \,dt + \sqrt{2  } \,dW_t, 
\label{eq:OverdampedL}
\end{equation}
where $W_t$ is a standard $d$-dimensional Wiener process. Its generator is given by
\begin{equation}
\cL_\mathrm{ovd}= \nabla_\tht (\log\pi(\cdot |\vectx))^T \gradT+  \Delta_\tht.
\label{L_ovd}
\end{equation}
It is well known that the process~\eqref{eq:OverdampedL} is irreducible and admits~$\pi(\cdot |\vectx)$ as a unique invariant probability measure, so that expectations can be approximated by trajectory averages (see for instance~\cite{MR885138}). In general, one cannot directly simulate the overdamped Langevin dynamics~\eqref{eq:OverdampedL}. To numerically approximate the solution, the widely used Euler-Maruyama scheme can be considered: 
\begin{equation}
\tht^{m+1} = \tht^m +\dt \nabla_\tht (\log\pi(\tht^m|\vectx))+ \sqrt{2 \dt}\, G^m,
\label{EulerM}
\end{equation}
where $\dt > 0$ is the step size, and $(G^m)_{m \geq 0}$ is a vector of i.i.d.~standard $d$-dimensional Gaussian random variables. It can be shown that the bias $\mathbb{E}_{\pi_{\Delta t } }(\phi) - \mathbb{E}_\pi (\phi)$ for overdamped Langevin dynamics is of order~$\mathcal{O}(\dt)$; see for instance~\cite{doi:10.1080/17442509008833606} and Theorem~7.3~ in~\cite{MR1931266} for pioneering works on ergodic properties of discretization of SDEs and error estimation for globally Lipschitz vector fields. 

Using the stochastic estimator~\eqref{eq:gradSto}, SGLD corresponds to the following numerical scheme:
\begin{equation} 
\tht^{m+1} = \tht^m +\Delta t\widehat{F_n}(\tht^m)+ \sqrt{2  \Delta t} \,G^m.
\label{SGLD}
\end{equation}
We assume in the sequel that this Markov chain admits a unique invariant probability measure~$\pi_{\Delta t,n}$ (we omit the dependence on~$\mathbf{x}$ to simplify the notation).

\begin{remark}
	It is suggested in~\cite{Welling:2011:BLV:3104482.3104568} to use decreasing time steps as this allows to ultimately eliminate the bias without resorting to a Metropolis Hastings scheme. However, considering decreasing time steps means that we need more iterations for a fixed final time~$T$, which increases the computational cost of the algorithm and results in a possibly large variance in the estimation of averages at fixed computational cost. In practice, it is more customary to use a small finite time step. As discussed in~\cite{JMLR:v17:15-494}, the bias arising from mini-batching dominates the one resulting from the time step discretization. Our focus in this work is therefore on the bias arising from mini-batching and not on the bias coming from the use of finite time steps.
\end{remark}

\subsubsection{Effective SGLD}
\label{sec:eff_SGLD}

We recall in this section how to prove that the bias between the invariant measure~$\pi_{\Delta t,n}$ of SGLD~\eqref{SGLD} and~$\pi(\cdot|\vectx)$ is of order $\mathcal{O}((1+\eps(n))\Delta t)$ -- by which we mean that error estimates such as~\eqref{eq:error_invariant_measure_general} hold with~$\Delta t^s$ replaced by~$(1+\eps(n))\Delta t$ on the right hand side, namely: for any smooth and compactly supported function~$\phi$, there exist~$\dt_\star>0$ and~$L$ such that 
\[
\left|\int_\Theta \phi \, d\pi_{\Delta t,n} - \int_\Theta \phi \, d\pi \right| \leq L(1+\eps(n))\Delta t. 
\]
Such error estimates were already obtained in~\cite{JMLR:v17:15-494}. This analysis is important since we will repeatedly rely on it to quantify the bias on the invariant probability measures of numerical schemes, in particular for Adaptive Langevin dynamics and their extensions, which is why we insist on presenting the overall strategy.

From a technical viewpoint, the analysis relies on effective dynamics, discussed at the end of Section~\ref{sec:Review_error_analysis}. To derive an effective dynamics for the SGLD, we follow the general framework of backward error analysis for SDEs considered in~\cite{Shardlow06,MR2783188,MR2970763,ACVZ12}, which is based on building a perturbation to the generator associated with the original dynamics~\eqref{L_ovd}, constructed so that SGLD evolved over one step is closer to the SDE associated with the modified generator than to the original dynamics. Recalling the definition~\eqref{eq:write_remainders_rigorously} for the notation we use for remainder terms, the results in~\cite{JMLR:v17:15-494} show that (see also Appendix~\ref{app_1} for a short proof allowing to make precise the dependence of the remainder term on~$n$)
\begin{equation}
\widehat{P}_{\Delta t,n} = \mathrm{e}^{\Delta t (\cL_\mathrm{ovd} + \Delta t \mathcal{A}_{\mathrm{ovd},n} )} + \mathcal{O}\left((1+\eps(n)^{3/2})\Delta t^3\right),
\qquad 
\mathcal{A}_{\mathrm{ovd},n} = \eps(n) \mathcal{A}_{\mathrm{mb}} + \mathcal{A}_{\mathrm{disc}},
\label{T_op}
\end{equation}
where $\widehat{P}_{\dt,n}$ is the evolution operator associated with the SGLD scheme~\eqref{SGLD}, and the operators~$\mathcal{A}_{\mathrm{mb}}$ and~$\mathcal{A}_{\mathrm{disc}}$ respectively encode the perturbations arising from mini-batching and time discretization:
\[
\begin{aligned}
\mathcal{A}_{\mathrm{mb}}\phi  & =\frac{1}{2}  \Sigma_\vectx : \nabla_\tht^2\phi, \\
\mathcal{A}_{\mathrm{disc}}\phi  & = - \gradT^2 (\log\pi(\cdot |\vectx)) : \nabla _\tht^2 \phi - \frac{1}{2} \nabla_\tht \Delta( \log\pi(\cdot |\vectx))^T \nabla_\tht  \phi\\
& \ \ -  \frac{1}{2}  \nabla_\tht (\log\pi(\cdot |\vectx))^T \nabla_\tht^2( \log\pi(\cdot |\vectx)) \nabla_\tht \phi.
\end{aligned}
\]
In the latter expressions, we denote by $:$ the Frobenius inner product of two square matrices, \textit{i.e.}
\[
\forall M^1, M^2 \in \R^{d \times d},
\qquad 
M^1:M^2 = \sum\limits_{i,j=1 }^d M^1_{i,j} M^2_{i,j}.
\]
In fact, as discussed in Remark~\ref{rem:erreur_sgld_Z_moment} of Appendix~\ref{app_1}, the error estimate in~\eqref{T_op} can be improved to $\mathcal{O}((1+\eps(n))\dt^3)$ when $\mathbb{E}[Z_{\vectx, \Nd, n}^3] = 0$. Using results from Remark~5.5 in~\cite{lelievre_stoltz_2016}, or from~\cite{JMLR:v17:15-494}, and following the general strategy of Talay--Tubaro estimates such as~\eqref{eq:Talay_Tubaro_like}, we deduce that the bias between~$\pi_{\Delta t,n}$ and~$\pi$ is of order $\mathcal{O}\left( (1+\eps(n))\Delta t\right)$ provided that~$\dt $ and~$\eps(n)\dt$ are small enough.

\begin{proposition}
  \label{prop:SGLD_error}
  Assume that the SGLD scheme~\eqref{SGLD} admits a unique invariant probability measure~$\pi_{\Delta t,n}$, and that~$\log \pi(\cdot|\vectx)$ satisfies the assumption of~\cite{Kopec14}. Introduce the smooth functions
  \begin{equation}
    \label{eq:f_disc_f_mb}
    f_\mathrm{disc} = \left(-\cL_\mathrm{ovd}\right)^{-1} \mathcal{A}_{\mathrm{disc}}^* \mathbf{1},
    \qquad
    f_\mathrm{mb} = \left(-\cL_\mathrm{ovd}\right)^{-1} \mathcal{A}_{\mathrm{mb}}^* \mathbf{1}.
  \end{equation}
  Then, for any smooth function~$\varphi$ with compact support, there is~$\dt_\star>0$ and~$K \in \R_+$ for which  
  \[
  \int_\Theta \varphi \, d\pi_{\Delta t,n} - \int_\Theta \varphi \, d\pi = \Delta t \int_\Theta \varphi \, (f_\mathrm{disc}+ \varepsilon(n)f_\mathrm{mb})\, d\pi + \varepsilon(n)^{3/2} \Delta t^2 R_{\varphi,n,\Delta t}, 
  \]
  with
  \[
  \sup_{1 \leq n \leq \Nd} \sup_{\dt \in (0,\dt_\star]} \left|R_{\varphi,n,\Delta t}\right| \leq K. 
  \]
\end{proposition}

The bias is determined by the norm of the correction functions~$f_\mathrm{disc},f_\mathrm{mb}$ in appropriate functional spaces. These functions are well defined and belong to~$L^2(\pi)$ in the framework of~\cite{Kopec14}, so that the bias term can be bounded as
\begin{equation}
  \label{eq:bias_SGLD_made_precise}
  \left| \Delta t \int_\Theta \varphi \, (f_\mathrm{disc}+ \varepsilon(n)f_\mathrm{mb})\, d\pi \right| \leq (1+\varepsilon(n))\Delta t \left( \|f_\mathrm{disc}\|_{L^2(\pi)} + \|f_\mathrm{mb}\|_{L^2(\pi)} \right) \|\varphi\|_{L^2(\pi)} .
\end{equation}
The bias is expected to be larger when~$\Sigma_\vectx$ is larger, because~$\mathcal{A}_{\mathrm{mb}}^* \mathbf{1}$ in~\eqref{eq:f_disc_f_mb} is larger -- although this statement is not as clear cut as for Langevin dynamics since the correction function~$f_\mathrm{mb}$ involves derivatives of~$\Sigma_\vectx$, in contrast to estimates obtained for underdamped and adaptive Langevin dynamics.

Note that the bias can be decreased by either decreasing~$\dt$ or increasing~$n$, hence decreasing~$\eps(n)$ in view of~\eqref{eq:eps}. However, as already discussed in~\cite{JMLR:v17:15-494}, the mini-batching error usually dominates the discretization error by orders of magnitude since~$\eps(n)$ is proportional to~$\Nd^2$ unless~$n$ is a fraction of~$\Nd$, in which case it scales as~$\Nd$.

\begin{remark}
  \label{rmk:SGLD_estimation_Sigma}
  To reduce the mini-batching bias, it is suggested in~\cite{JMLR:v17:15-494} to renormalize the magnitude of the injected noise by a quantity involving~$\Sigma_{\emph{\vectx}}(\tht)$ (the so-called modified SGLD scheme). Since~$\Sigma_{\emph{\vectx}}(\tht)$ is usually unknown, this requires estimating this matrix, which is however computationally expensive since the estimation has to be repeated for each new value of~$\tht$, and may cancel the gain provided by mini-batching in the first place. Our focus in this work is to reduce the mini-batching error without the need to estimate~$\Sigma_{\emph{\vectx}}(\tht)$.
\end{remark}

\subsection{Langevin dynamics with mini-batching}
\label{sec:Langevin}

It has been observed in practice that a better sampling of probability measures can be provided by Langevin dynamics, both in the literature on computational statistical physics (see for instance~\cite{cances-le-goll-stoltz-07}) and more recently in the machine learning literature, see for instance~\cite{DRD20}. We first present in Section~\ref{sec:description_Langevin_mb} the numerical scheme obtained by discretizing Langevin dynamics with a splitting scheme and replacing the gradient of the log-likelihood by its estimator~\eqref{eq:gradSto}, and then analyze the bias induced on the invariant probability measure of the numerical scheme in Section~\ref{sec:effective_Langevin_dynamics}.

\subsubsection{Standard Langevin dynamics}
\label{sec:description_Langevin_mb}

Langevin dynamics introduces some inertia in the evolution of $\tht$, through an extended configuration space with a momentum vector~$p$ conjugated to~$\tht$. It can be seen as a perturbation of the Hamiltonian dynamics where some fluctuation/dissipation mechanism is added to the evolution of the momenta, and reads
\begin{equation}
\left\{ \begin{aligned}
d\tht_t & = p_t \,dt, \\
dp_t & = \gradT( \log\pi(\tht_t|\vectx))\,dt - \Gamma  p_t \,dt + \sqrt{2} \Gamma^{1/2}\,dW_t,
\end{aligned} \right.
\label{Langevin}
\end{equation}
where~$\Gamma \in \R^{d \times d}$ is a positive definite symmetric matrix. It would be possible, as in molecular dynamics, to attach a mass to each degree of freedom, but we set here for simplicity this mass to~$1$, the generalization to non trivial mass matrices being straightforward. The generator of Langevin dynamics~\eqref{Langevin} is given by
\begin{equation}
\cL_{\mathrm{lan}} = \gradT (\log\pi(\cdot|\vectx))^T \nabla_p + p^T \nabla_\tht - p^T \Gamma \nabla_p + \Gamma : \nabla_p^2.
\label{gen_lan}
\end{equation}

The diffusion constant~$ \sqrt{2} \Gamma^{1/2}$ in front of the Wiener process ensures that the following probability distribution is invariant:
\[
\mu(d\tht\,dp|\vectx) = \pi(\tht|\vectx)\tau(dp) \,d\tht ,
\]
where 
\begin{equation}
\tau(dp) = (2\pi)^{-d/2} \rme^{-p^2/2} \, dp,
\label{eq:tau}
\end{equation}
see for instance~\cite{pavliotis2014stochastic,29acd3d494044594aea0829ef236aad6,lelievre_stoltz_2016}. In particular, the marginal distribution of~$\mu(\cdot|\vectx)$ in the~$\tht$ variable is indeed the target distribution~\eqref{pi_bayes}.  It can even be shown that time averages along solutions of~\eqref{Langevin} almost surely converge to averages with respect to~$\mu$ since the generator of the dynamics is hypoelliptic; see~\cite{MR885138}. 

To numerically approximate the solution of Langevin dynamics, we use a numerical integrator based on a second order Strang splitting, as encoded by the following evolution operator (although there are various other choices of orderings, see for instance~\cite{LM12,MR3463433,29acd3d494044594aea0829ef236aad6}):
\begin{equation}
P_{\Delta t} = \mathrm{e}^{\Delta t \cLc/2}\mathrm{e}^{\Delta t \cLb/2} \mathrm{e}^{\Delta t \cLa}\mathrm{e}^{\Delta t \cLb/2}\mathrm{e}^{\Delta t \cLc/2},
\label{opLan_P}
\end{equation}
where 
\begin{equation}
\cLa = \gradT \log(\pi(\cdot|\vectx))^T \nabla_p, \qquad
\cLb = p^T \nabla_\tht, \qquad
\cLc = -p^T \Gamma \nabla_p + \Gamma : \nabla^2.  
\label{eq:L_op}
\end{equation}
The elementary generators~$\cLa$ and~$\cLb$ are respectively associated with the elementary differential equations~$dp_t  = \gradT \log(\pi(\tht_t|\vectx)\,dt $ and~$d\tht_t  = p_t \, dt$, which can be analytically integrated. The elementary generator~$\cLc$ is associated with the Ornstein--Uhlenbeck process~$dp_t  = - \Gamma p_t \, dt + \sqrt{2 } \Gamma^{1/2} \, dW_t$, which can also be analytically integrated as:
\begin{equation}
p_t = \mathrm{e}^{-\Gamma t} p_0 + \sqrt{2} \int\limits_{0}^t  \mathrm{e}^{-(t-s) \Gamma} \Gamma^{1/2} \,dW_s \sim \Norm(\alpha_{t} p_0, \mathrm{I}_d-\alpha_{2t}), \qquad \alpha_{t} = \mathrm{e}^{-\Gamma t}.
\end{equation}
Finally, the numerical scheme encoded by~\eqref{opLan_P} reads:
\begin{equation}
\left\{ \begin{aligned}
p^{m+\frac{1}{3}}  & =\alpha_{\Delta t/2}\, p^{m}  + \left( \mathrm{I}_d-\alpha_{\Delta t} \right)^{1/2}\,G^m,\\
\tht^{m+ \frac{1}{2}} &= \tht^m + \frac{\Delta t}{2}   \,p^{m+\frac{1}{3}}, \\
p^{m+\frac{2}{3}} & = p^{m+\frac{1}{3}}  + \Delta t\gradT \left[\log\pi\left(\tht^{m+ \frac{1}{2}} \middle|\vectx\right)\right],\\
\tht^{m+1} & = \tht^{m+ \frac{1}{2}} +   \frac{\Delta t}{2}   p^{m+\frac{2}{3}}, \\
p^{m+1}  & =\alpha_{\Delta t/2}\,   p^{m+\frac{2}{3}} + \left( \mathrm{I}_d-\alpha_{\Delta t} \right)^{1/2}\,G^{m+ \frac{1}{2}},
\end{aligned} \right.
\label{GLA}
\end{equation}
where~$(G^m)_{m \geq 0}$ and~$(G^{m+ \frac{1}{2}})_{m \geq 0}$ are two independent families of i.i.d.~standard $d$-dimensional Gaussian random variables. Moreover, it is proved in~\cite{MR3463433, MR3229658, DEMS21} that the Markov chain generated by~\eqref{GLA} admits a unique invariant probability measure~$\mu_{\Delta t}$ and that there exists~$C \in \R+ $ such that, for any smooth function~$\phi$ with compact support,
\[
\left|\int_{\Theta}\phi (\tht, p)\mu_{\Delta t}(d\tht \, d p) -  \int_{\Theta}\phi(\tht, p) \mu(d\tht \, d p|\mathbf{x})\right| \leq C \Delta t^2.
\]
The key element to prove this statement is the fact that the numerical scheme~\eqref{GLA} is an approximation of weak order~$2$ of Langevin dynamics (see~\cite{MR3463433, MR3229658})
\begin{equation}
P_{\dt} = \mathrm{e}^{\dt \cL_{\mathrm{lan}}}+ \mathcal{O}(\dt^3).
\label{eq:p_ovd}
\end{equation}

\subsubsection{Error estimates for Langevin dynamics with mini-batching}
\label{sec:effective_Langevin_dynamics}
In order to analyze the error on the posterior measure sampled by a discretization of Langevin dynamics used in conjunction with mini-batching, we derive here the effective dynamics associated with the numerical method~\eqref{GLA} when the gradient of the log-likelihood is replaced by its stochastic estimator~\eqref{eq:gradSto} (similar results are obtained for other Strang splittings). This corresponds to the following numerical scheme: 
\begin{equation}
\left\{ \begin{aligned}
p^{m+\frac{1}{3}}  & =\alpha_{\Delta t/2}\, p^{m}  + \left( \mathrm{I}_d-\alpha_{\Delta t} \right)^{1/2}\,G^m,\\
\tht^{m+ \frac{1}{2}} &= \tht^m + \frac{\Delta t}{2}   \,p^{m+\frac{1}{3}}, \\
p^{m+\frac{2}{3}} & = p^{m+\frac{1}{3}}  + \Delta t \widehat{F_n}\left(\tht^{m+ \frac{1}{2}}\right),\\
\tht^{m+1} & = \tht^{m+ \frac{1}{2}} +   \frac{\Delta t}{2}   p^{m+\frac{2}{3}}, \\
p^{m+1}  & =\alpha_{\Delta t/2}\,   p^{m+\frac{2}{3}} + \left( \mathrm{I}_d-\alpha_{\Delta t} \right)^{1/2}\,G^{m+ \frac{1}{2}}.
\end{aligned} \right.
\label{GLA_minibatched}
\end{equation}
When using mini-batching, we are in fact replacing $\mathrm{e}^{\Delta t \cLa}$ in~\eqref{opLan_P} by the elementary evolution operator $Q_{\Delta t}^{\cLa}$ acting as
\[
\left(Q_{\Delta t}^{\cLa}\varphi\right)(\tht,p) = \mathbb{E}\left[\varphi\left(\tht,p + \Delta t \widehat{F_n}(\tht)\right)\right].
\]
A simple computation provided in Appendix~\ref{app_2} shows that 
\begin{equation}
Q_{\Delta t}^{\cLa} = \mathrm{e}^{\Delta t \cLa} + \mathcal{O}(\eps(n)\dt^2).
\label{eq:op_Q}
\end{equation}
Denoting by~$\widehat{P}_{\Delta t,n}$ the evolution operator of the numerical scheme~\eqref{GLA_minibatched}, and by~$\mu_{\dt,n}$ its invariant probability measure (assuming that it exists and it is unique), it can then be proved that (see Appendix~\ref{app_2})
\begin{equation}
\widehat{P}_{\Delta t,n} = \mathrm{e}^{\dt\cL_{\mathrm{lan}}} + \mathcal{O}\left((\eps(n)+\dt)\Delta t^2\right).
\label{eq:op_lan_mini}
\end{equation}
This allows to formulate the following result on the bias of~$\mu_{\dt,n}$. 

\begin{proposition}
  \label{prop:Langevin_error}
  Assume that the Markov chain~\eqref{GLA_minibatched} admits a unique invariant probability measure~$\mu_{\Delta t,n}$, and that~$\log \pi(\cdot|\vectx)$ satisfies the assumption of~\cite{KopecLangevin}. Then, there exist smooth functions~$f_\mathrm{disc},f_\mathrm{mb}$ such that, for any smooth function~$\varphi$ with compact support, there is~$\dt_\star>0$ and~$K \in \R_+$ for which  
  \[
  \int_{\Theta \times \R^d} \varphi \, d\mu_{\Delta t,n} - \int_{\Theta \times \R^d} \varphi \, d\mu = \Delta t \int_{\Theta \times \R^d} \varphi \, (\dt f_\mathrm{disc}+ \varepsilon(n)f_\mathrm{mb})\, d\mu + (\dt+\varepsilon(n)^{3/2})\dt^2 R_{\varphi,n,\Delta t}, 
  \]
  with
  \[
  \sup_{1 \leq n \leq \Nd} \sup_{\dt \in (0,\dt_\star]} \left|R_{\varphi,n,\Delta t}\right| \leq K. 
  \]
\end{proposition}

Note that the factor~$\dt$ in the term~$\dt+\varepsilon(n)$ of the remainder is useful only when~$n = \Nd$, in which case~$\varepsilon(n) = 0$. Proposition~\ref{prop:Langevin_error} makes precise the statement that the bias between $\mu$ and $\mu_{\dt,n}$ is of order $\mathcal{O}\left((\eps(n)+\dt)\dt\right)$. Indeed, $f_\mathrm{disc},f_\mathrm{mb}$ belong to~$L^2(\pi)$ in the framework of~\cite{Kopec14} (see also the discussion in~\cite{lelievre_stoltz_2016}), so that the bias term can be bounded as
\begin{equation}
  \label{eq:bias_Langevin_made_precise}
  \left| \Delta t \int_{\Theta \times \R^d} \varphi \, (\dt f_\mathrm{disc}+ \varepsilon(n)f_\mathrm{mb})\, d\mu \right| \leq \dt \left( \dt \|f_\mathrm{disc}\|_{L^2(\mu)} + \varepsilon(n) \|f_\mathrm{mb}\|_{L^2(\mu)} \right) \|\varphi\|_{L^2(\mu)}.
\end{equation}
It is clear that the error~$\eps(n)\dt$ coming from mini-batching dominates the error~$\dt^2$ due to time discretization since~$\eps(n)$ is much larger than~$\dt$ unless~$n$ is very close to~$\Nd$. Comparing this equality to~\eqref{eq:p_ovd} highlights the fact that mini-batching degrades the consistency estimate~\eqref{eq:p_ovd} by one order in~$\dt$ with respect to Langevin dynamics~\eqref{Langevin}. The bias is of the same order of magnitude as for SGLD, see~\eqref{eq:bias_SGLD_made_precise}.

We do not provide at this stage functional estimates on the function~$f_\mathrm{mb}$ which appears in the dominant term of the bias in~\eqref{eq:bias_Langevin_made_precise}. This is discussed after~\eqref{eq:EfLangevin} below.

\begin{remark}
  To remove the bias on the invariant probability measure at dominant order in~$\eps(n)\dt$, it is suggested in~\cite{2018arXiv180508863M} to modify the integration of the Ornstein--Uhlenbeck part on the momenta. This requires however an estimation of the covariance matrix, which can again be computationally prohibitive (as discussed in Remark~\ref{rmk:SGLD_estimation_Sigma}). 
\end{remark}

\subsubsection{Effective dynamics for Langevin dynamics with mini-batching}

We now construct an effective dynamics which coincides at order~$3$ in~$\Delta t$ over one time step with the numerical scheme~\eqref{GLA_minibatched} even when using mini-batching, in order to obtain an equality similar to~\eqref{eq:p_ovd}. This effective dynamics is the key building block to understand Adaptive Langevin dynamics in Section~\ref{sub:AdL}. A straightforward computation, following the lines of~\cite{2018arXiv180508863M}, shows that (see Appendix~\ref{app_2})
\begin{equation}
\label{eq:widehatP_Langevin_mb}
\widehat{P}_{\Delta t,n} = \mathrm{e}^{\dt(\cL_{\mathrm{lan}}+ \Delta t \eps(n) \mathcal{A}_\mathrm{lan})}+ \mathcal{O}\left((1+ \eps(n)^{3/2})\Delta t^3\right), \qquad \mathcal{A}_\mathrm{lan} =\frac{1}{2} \dps\Sigma_\vectx: \nabla_p^2.
\end{equation}
The effective dynamics associated with $\cL_{\mathrm{lan}}+ \Delta t \eps(n) \mathcal{A}_\mathrm{lan}$ is
\begin{equation}
\left\{ \begin{aligned}
d\widetilde{\tht_t} & = \widetilde{p_t} \,dt, \\
d\widetilde{p_t} & = \nabla_\tht\left[ \log\pi\left(\widetilde{\tht_t}\middle|\vectx\right)\right]dt - \Gamma  \widetilde{p_t} \,dt + \left( 2\Gamma + \eps(n)\dt  \Sigma_\vectx(\widetilde{\tht_t})\right)^{1/2} dW_t,
\end{aligned} \right.
\label{eq:EfLangevin}
\end{equation}
where $2\Gamma +  \eps(n) \Delta t \Sigma_\vectx(\tht)$ is a positive definite matrix since $\Sigma_\vectx$ is positive (even though the latter matrix is unknown). This allows to identify the correction function~$f_\mathrm{mb}$ in~\eqref{eq:bias_Langevin_made_precise} as
\[
f_\mathrm{mb} = \left(-\cL_{\mathrm{lan}}^*\right)^{-1}\mathcal{A}_\mathrm{lan}^* \mathbf{1} = \frac12 \left(-\cL_{\mathrm{lan}}^*\right)^{-1} \Sigma_\vectx: (\nabla_p^*)^2 \mathbf{1},
\]
where the adjoints of the operator are taken on~$L^2(\mu)$. Thanks to estimates on~$\cL_{\mathrm{lan}}$ obtained from hypocoercive estimates (see~\cite{herau:hal-00004498,MR2576899,MR3324910}), we deduce that there exists a constant~$C \in \R_+$ such that
\begin{equation}
  \label{eq:f_mb_Langevin}
  \left\| f_\mathrm{mb} \right\|_{L^2(\mu)} \leq C  \left\| \Sigma_\vectx: (\nabla_p^*)^2 \mathbf{1} \right\|_{L^2(\mu)} = C \left\| (\nabla_p^*)^2 \mathbf{1} \right\|_{L^2(\tau)} \left\| \Sigma_\vectx \right\|_{L^2(\pi)}.
\end{equation}
Here and in the sequel, we assume implicitly that~$\Sigma_\vectx$ has all its entries in~$L^2(\pi)$. The bias on the invariant probability measure sampled by the numerical scheme is therefore, at dominant order, of order~$\eps(n)\dt \| \Sigma_\vectx \|_{L^2(\pi)}$, in accordance with the general estimate~\eqref{eq:erreur_gen} (choosing~$\mathscr{S} = \{0\}$ hence~$S^* = 0$). 

\begin{remark}
  \label{remark1}
  Consider the simple case when the covariance of the gradient estimator is contant, namely~$\Sigma_\emph{\vectx} = \sigma^2 \mathrm{I}_d \in \R^{d \times d}$, and the friction is isotropic, namely~$\Gamma = \gamma  \mathrm{I}_d \in \R^{d \times d}$, with $\sigma, \gamma>0 $. In this situation, the modified Langevin dynamics~\eqref{eq:EfLangevin} samples the invariant probability measure with a density proportional to~$\mu(\theta)^{\beta_{\rm eff}}$, with
  \begin{equation}
    \beta_{\rm eff}= \left(1+\frac{\eps(n)\sigma^2\Delta t}{2\gamma}\right)^{-1} < 1.
    \label{eq:beta_eff}
  \end{equation}
  Since the exponent~$\beta_\mathrm{eff}$ is smaller than~1, the modified Langevin dynamics samples a tempered version of~$\mu$.
\end{remark}


\subsection{Numerical illustration}
\label{sec:numEff}

We illustrate in this section the results on the bias of the posterior measure introduced by mini-batching when the elementary likelihoods are given by either a Gaussian or a mixture of Gaussians. The aim is to numerically quantify the bias in the non asymptotic regime $n = \mathcal{O}(1)$ (for which $Z_{\vectx, \Nd, n}$ is not Gaussian), and to have a benchmark on the bias to compare with AdL.

\subsubsection{Gaussian posterior}
\label{sec:gaussian_model}

We first suppose that the elements of the data set are normally distributed, namely $x_i|\theta \sim \Norm(\theta, \sigma_x^2)$, where~$\tht$ is the parameter to estimate. The prior distribution on~$\tht$ is a centered normal distribution with variance~$\sigma_\theta^2$. In this case, simple computations show that the posterior distribution on $\tht$ is also Gaussian with mean $\mu_{\rm post}$ and variance $\sigma_{\rm post}^2$, where (see for instance~\cite{JMLR:v17:15-494})
\begin{equation}
\mu_{\rm post}= \left(\frac{\sigma_x^2}{ \sigma_\theta^2}+\Nd\right)^{-1} \sum\limits_{i=1}^{\Nd}  x_i, \qquad \sigma_{\rm post}^2 =  \left(\frac{1}{ \sigma_\theta^2}+ \frac{\Nd}{ \sigma_x^2 }\right)^{-1}.
\label{eq:mu_sigma_post}
\end{equation}
Moreover,
\[
\widehat{F_n}(\theta) = - \frac{\theta }{ \sigma_\theta^2} + \frac{\Nd}{n}\sum\limits_{i\in I_n} \frac{x_i-\theta}{\sigma_x^2}.
\]
The variance of $\widehat{F_n}(\theta) $ as a function of $\theta$ can therefore be analytically computed using~\eqref{eq:sigma_somme} (see again~\cite{JMLR:v17:15-494} for instance):
\begin{equation}
\Sigma_{\vectx}(\tht) =  \mathrm{var}_\mathcal{I}\left[ \gradT (\log \PLe(x_\mathcal{I} |\theta))\right] = \frac{\mathrm{var}(\vectx)}{  \sigma_x^4} ,
\label{eq:var_hat_F}
\end{equation}
where $\mathcal{I}$ is a random variable uniformly distributed in $\{1, ..., \Nd\}$, and
\[
\var(\vectx) = \dps\frac{1}{\Nd-1} \sum_{i=1}^{\Nd} \left( x_i - \frac{1}{\Nd} \sum_{j=1}^{\Nd} x_j\right)^2
\]
is the empirical variance of the data. In this case, the covariance of the force~$\Sigma_{\vectx}$ is constant and does not depend on the parameter $\tht$.

Let us first discuss SGLD. As shown in Section~2.1 in~\cite{JMLR:v17:15-494}, there is no bias on the mean of the posterior distribution~$\pi_{\Delta t,n}$ of the Markov chain associated with SGLD. The latter distribution is however not strictly Gaussian, because the random variable~$Z_{\vectx,\Nd,n}$ is not Gaussian. It can however be shown that the non Gaussianity arises from higher order terms, so that the shape of the the distribution is Gaussian at dominant order in~$(1+\varepsilon(n))\Delta t$. In order to characterize the dominant order of the shape of the distribution, we write the effective dynamics with generator~$\cL_\mathrm{ovd} + \Delta t \mathcal{A}_{\mathrm{ovd},n}$ in~\eqref{T_op}, which turns out to be an Ornstein--Uhlenbeck process:
\[
d\tht_t = \left[-\left( 1 + \frac{\dt}{2 \sigma_{\rm post}^2} \right) \frac{\tht - \mu_{\rm post}}{\sigma_{\rm post}^2}\right] dt +\sqrt{2} \left(1  + \frac{\Delta t\eps(n)}{2 \sigma_x^4} \mathrm{var}(\vectx) + \frac{\Delta t}{\sigma_{\rm post}^2}\right)^{1/2} dW_t.
\]
The probability measure of the latter process is a Gaussian distribution with mean~0 and variance 
\[
\sigma_{\rm post}^2 \left(1 + \frac{\dt}{2 \sigma_{\rm post}^2}\right)^{-1} \left( 1 + \frac{\eps(n)\dt}{2 \sigma_{x}^4}\var(\vectx)+ \frac{\dt}{\sigma_{\rm post}^2} \right).
\]
This shows that the invariant probability measure of SGLD is, at dominant order in~$\dt$ and~$\eps(n)\dt$, a Gaussian distribution with mean~0 and variance
\begin{equation}
\sigma_{\rm post}^2 \left[1 + \frac{\dt}{2}\left(\frac{\eps(n)}{\sigma_{x}^4}\var(\vectx) + \frac{1}{\sigma_{\rm post}^2} \right)\right] + \mathcal{O}\left((1+\eps(n)^2)\dt^2\right).
\label{eq:sigme_post_erreur_SGLD}
\end{equation}

For Langevin dynamics with mini-batching, there is (as for SGLD) no bias on the mean of the posterior distribution~$\pi_{\Delta t,n}$, as proved in Appendix~\ref{app:unbiased_Langevin}. As for SGLD, the latter distribution is also not strictly Gaussian when the random variable~$Z_{\vectx,\Nd,n}$ is not Gaussian. It is however close to a Gaussian distribution since the marginal posterior distribution in the~$\tht$ variable obtained for the effective Langevin dynamics is, according to Remark~\ref{remark1}, a Gaussian distribution with mean~$\mu_{\rm post}$ and variance~$\sigma_{\rm post}^2/ \beta_{\rm eff}$. This means that the variance for the discretization of Langevin dynamics~\eqref{GLA_minibatched} is, at dominant order in~$\dt$ and~$\eps(n)\dt$, and when~$\Gamma = \gamma \mathrm{I}_d$,  
\begin{equation}
\sigma_{\rm post}^2 \left(1+ \frac{\eps(n)\dt}{2 \gamma \sigma_{x}^4}\var(\vectx) \right) + \mathcal{O}\left((1+\eps(n)^2) \dt ^2 \right).
\label{eq:sigme_post_erreur}
\end{equation}
Note that this expression coincides with~\eqref{eq:sigme_post_erreur_SGLD} when~$\gamma=1$ and~$\eps(n) \gg 1$.

\begin{figure}
	\centering
	\begin{minipage}{0.49\textwidth}
		\centering
		\includegraphics[width=\textwidth]{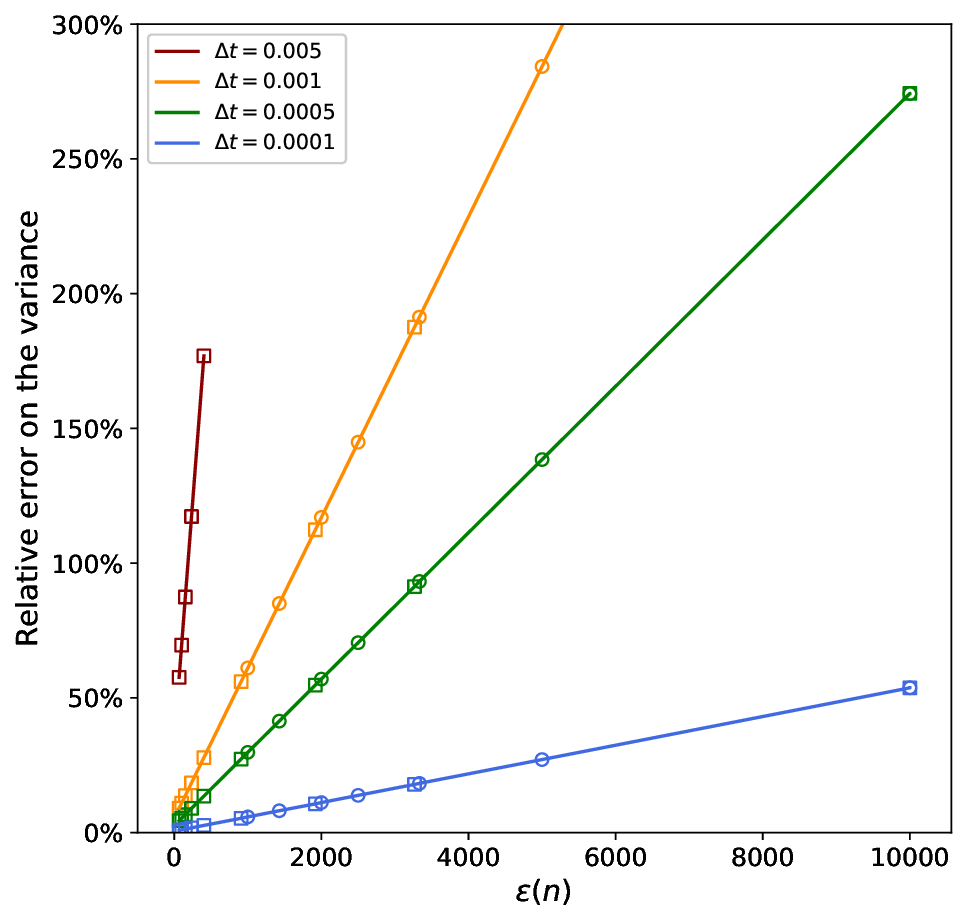}
		\subcaption[first caption.]{SGLD}
	\end{minipage}
	\begin{minipage}{0.49\textwidth}
		\centering
		\includegraphics[width=\textwidth]{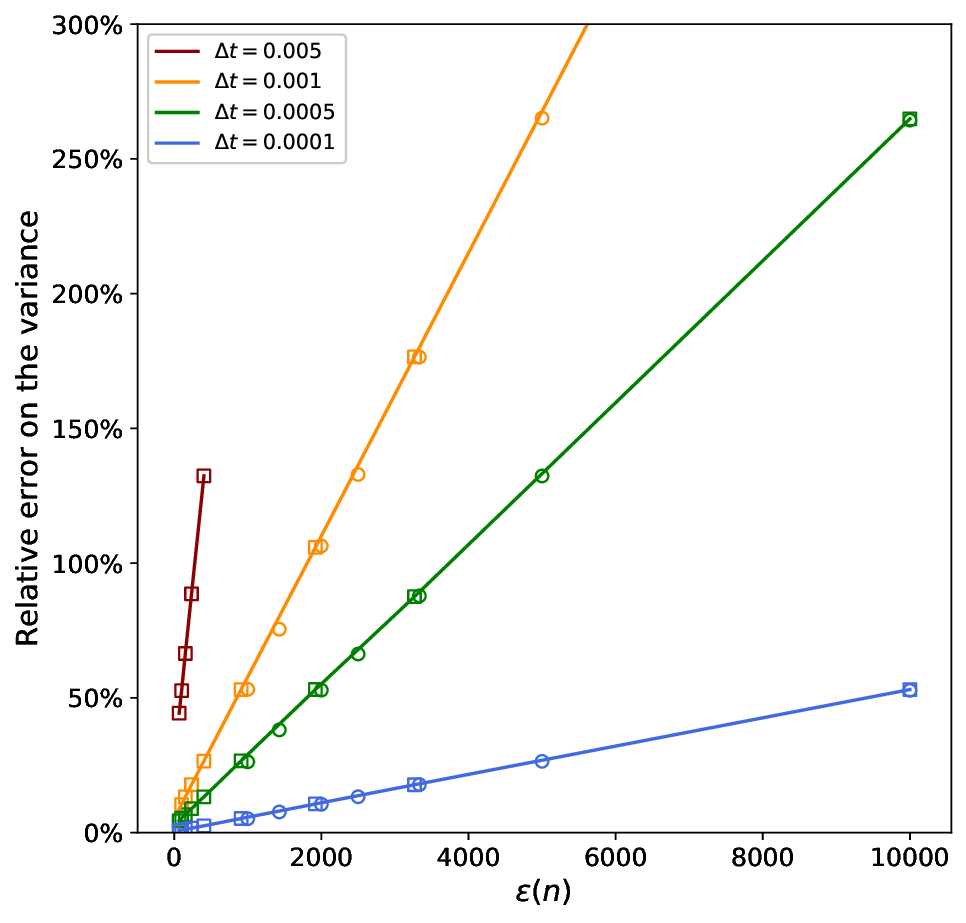}
		\subcaption[second caption.]{Langevin dynamics}
	\end{minipage}
	\caption{Relative error on the variance of the posterior distribution for various values of $\Delta t$ and $n$ when the elementary likelihood is a Gaussian distribution, when sampling with (circles) and without replacement (squares).} \label{fig:SGLD_gauss}
\end{figure}

To perform the numerical experiments, we generate a dataset of $\Nd = 100$ according to a Gaussian distribution with mean $\tht_0 = 0$ and variance~$\sigma_{x} = 1$. We also set~$\sigma_{\tht} = 1$ in the prior distribution. We run the SGLD scheme~\eqref{SGLD} and Langevin dynamics with the numerical scheme~\eqref{GLA_minibatched} with $\Gamma = 1$ for a final time~$T = 10^{6}$ and various values of~$\Delta t$ (which corresponds to $\Ni = T/\Delta t$ time steps). We also consider various values of~$n$, the subsampling of the data points being done with and without replacement. We report in Figure~\ref{fig:SGLD_gauss} the relative bias on the variance, given in the limit $\Ni\to+\infty$ by the ratio of
\[
\left| \left[\int_{\R}\tht^2\pi_{\Delta t,n}(\tht|\vectx) \,d\tht -\left(\int_{\R}\tht\pi_{\Delta t,n}(\tht|\vectx) \,d\tht \right)^2 \right]- \left[ \int_{\R} \tht^2\pi(\tht|\vectx) \,d\tht  -  \left(\int_{\R} \tht\pi(\tht|\vectx) \,d\tht  \right)^2     \right]\right| ,
\]
and~$\sigma^2_{\rm post}$, where $\pi_{\Delta t,n}(\cdot|\vectx)$ denotes the invariant measure in the~$\theta$ variable of the numerical scheme under consideration (for~\eqref{GLA_minibatched}, this corresponds to the marginal measure of~$\mu_{\dt,n}(\tht, p|\vectx)$ in the~$\tht$ variable). 

Several conclusions can be drawn from the results presented in Figure~\ref{fig:SGLD_gauss}. First, note that the bias is determined by the value of~$\eps(n)$ irrespectively of the fact that sampling is performed with or without replacement. Second, the bias is indeed affine in~$\eps(n)$, with a slope which proportional to~$\dt$. In fact, all curves would be superimposed if we were plotting the error as a function of~$\eps(n)\dt$, which demonstrates that the error is indeed determined at dominant order by this parameter. We do not report results for $\eps(n) = 0$, which corresponds to $n=\Nd$ and sampling without replacement, since the corresponding error is anyway much smaller than the one arising from mini-batching. Finally, errors are very similar for SGLD and the scheme~\eqref{GLA_minibatched} associated with Langevin dynamics, as anticipated from the comparison of~\eqref{eq:sigme_post_erreur_SGLD} and~\eqref{eq:sigme_post_erreur} when choosing $\gamma=1$.

\subsubsection{Mixture of Gaussians} 
\label{sec:mixture_model}

\begin{figure}
	\centering
	\begin{minipage}{0.49\textwidth}
		\centering
		\includegraphics[width=\textwidth]{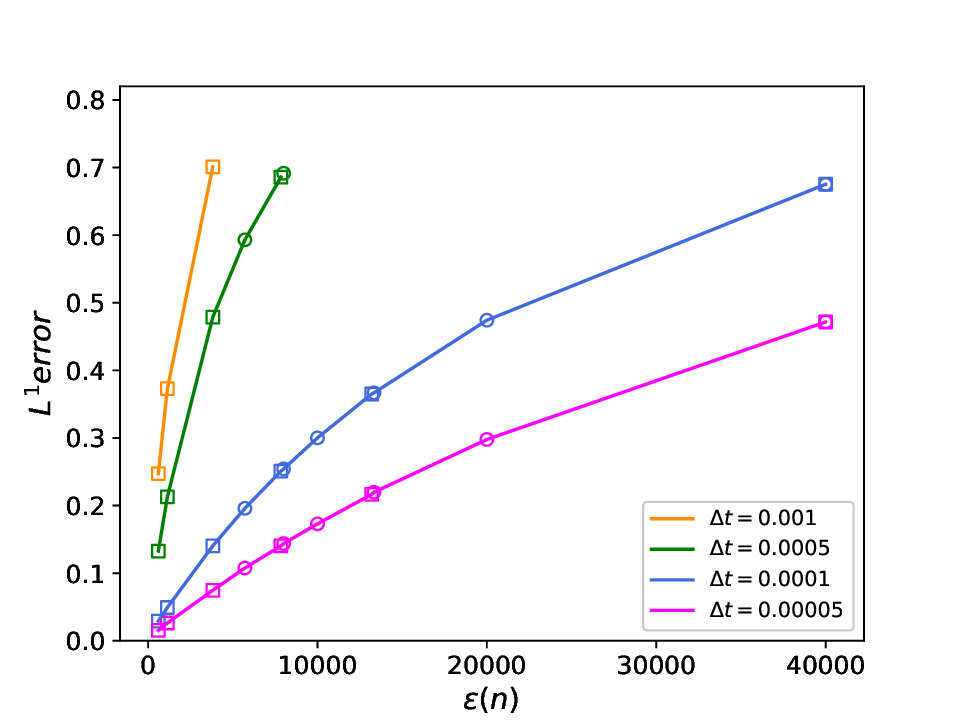}
		\subcaption[first caption.]{SGLD}
	\end{minipage}
	\begin{minipage}{0.49\textwidth}
		\centering
		\includegraphics[width=\textwidth]{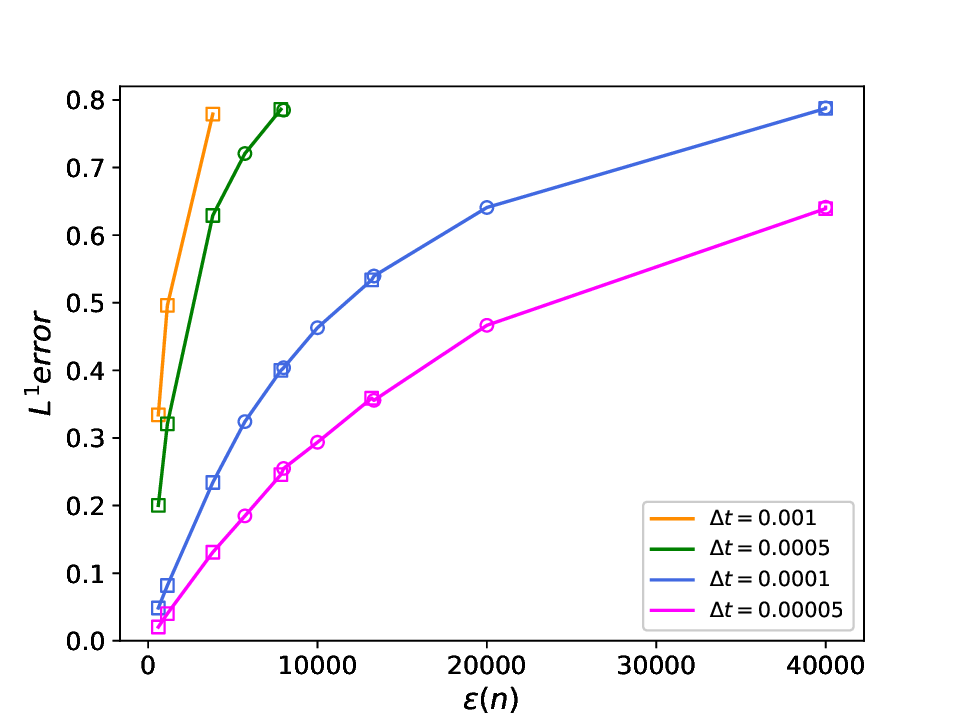}
		\subcaption[second caption.]{Langevin dynamics}
	\end{minipage}
	\caption{$L^1$ error on the  $\tht_1$ marginal of the posterior distribution for various values of $\Delta t$ and $n$ when the elementary likelihoods are mixtures of Gaussians, when sampling with (circles) and without replacement (squares).} \label{fig:SGLD}
\end{figure}

We next consider a more realistic case where the data points are distributed according to a mixture of Gaussians:
\begin{equation}
\PLe(x_i|\tht)  = \frac{w}{\sqrt{2\pi }\sigma_1}\exp\left( - \frac{(x_i-\mu_1)^2}{2 \sigma_1^2} \right) + \frac{1-w}{\sqrt{2\pi }\sigma_2}\exp\left( - \frac{(x_i-\mu_2)^2}{2 \sigma_2^2} \right), 
\label{eq:mixture}
\end{equation}
where $w \in[0,1]$, $\mu_1, \mu_2 \in \R$ and $\sigma_1, \sigma_2 \in (0,+\infty)$. We consider the case when the parameters to estimate are the centers of the Gaussians $\tht= (\mu_1, \mu_2)$, whereas $\sigma_1$, $\sigma_2$ and $w$ are given. The prior distribution on the vector of parameters $\tht$ is chosen to be a centered normal distribution with covariance matrix~$\mathrm{I}_2$. To perform numerical simulations, we fix $\mu_1 = 1$, $\mu_2 = 0.5$, $\sigma_{1} = \sigma_{2}  = 0.4$, $w=0.4 $ and $\Nd = 200$ to generate the dataset according to~\eqref{eq:mixture}. We run again the numerical schemes~\eqref{SGLD} and~\eqref{GLA_minibatched}, with $\Gamma = \mathrm{I}_2$ and an integration time~$T = 10^6$. We compute the $L^1$ error on the marginal distribution in the $\tht_1 = \mu_1$ variable, given in the limit~$\Ni\to+\infty$ by
\begin{equation}
\int_{\R} \left|\int_{\R} \pi_{\Delta t,n}(\tht |\vectx)\, d \tht_2- \int_{\R}  \pi(\tht|\vectx)\, d \tht_2 \right| d \tht_1.
\label{eq:erreur_L1}
\end{equation}
We plot this error with respect to $\eps(n)$ for various values of $\Delta t$ in Figure~\ref{fig:SGLD}. We use numerical quadratures to approximate the integral with respect to~$\tht_1$ in~\eqref{eq:erreur_L1} (approximating the marginals as piecewise constant functions over a grid of 500~bins over the interval~[0,1.4]), and also to compute the integral with respect to~$\tht_2$ for~$\pi(\cdot |\vectx)$.

The interpretation of the results presented in Figure~\ref{fig:SGLD} is quite similar to the discussion of the results of Figure~\ref{fig:SGLD_gauss}. They confirm that the bias is determined by the value of~$\eps(n)$ irrespectively of the fact that sampling is performed with or without replacement. Note also that the errors are quite similar for SGLD and Langevin dynamics with~$\Gamma= \mathrm{I}_2$. Moreover, the $L^1$ error~\eqref{eq:erreur_L1} is affine in~$\eps(n)$ provided~$\eps(n)\dt$ is sufficiently small so that the asymptotic analysis of the bias is indeed valid. From a quantitative viewpoint, the affine regime is observed for $L^1$ errors below~0.1. This regime is relevant for time steps~$\dt \leq 10^{-4}$ for the values of~$n$ considered in our simulations, but not for simulations with larger time steps.

\section{Adaptive Langevin dynamics}
\label{sec:AdL}

Using SGLD or Langevin dynamics with mini-batching introduces a bias on the posterior distribution. We recall in Section~\ref{sub:AdL} the Adaptive Langevin (AdL) dynamics introduced in~\cite{PMID:21895177} and~\cite{NIPS2014_5592}. The aim of this dynamics is to remove, or at least substantially reduce the bias due to mini-batching. Under the key assumption that the covariance matrix $\Sigma_\vectx(\tht)$  defined in~\eqref{eq:sigma_somme} is constant (but unknown), it can indeed be proved that Adaptive Langevin dynamics samples the target distribution (see Section~\ref{sec:AdL_cst}). However, we demonstrate in Section~\ref{subsec:Covariance} that this assumption may not hold in practice, \emph{e.g}. for models for which the elementary likelihood is a mixture of Gaussians, which motivates the construction of an extended version of AdL to tackle such cases. We also quantify the bias on the invariant measure due to the fact that~$\Sigma_\vectx$ is not constant by deriving an error estimate involving~\eqref{eq:erreur_gen}, henceforth explaining why AdL reduces the minibatching bias compared to SGLD or Langevin dynamics. Our analysis also allows to understand why AdL with a scalar friction can provide results of almost the same quality as AdL with a genuine friction matrix. This is illustrated for a model of logistic regression for MNIST data in Section~\ref{sec:log_reg}, and for Bayesian neural networks in Section~\ref{sec:BNN}.

\subsection{General formulation of Adaptive Langevin dynamics}
\label{sub:AdL}

AdL was initially introduced to address the issue of the gradient of the energy not being exactly computed in molecular dynamics in~\cite{PMID:21895177}. It was then considered for Bayesian inference in~\cite{NIPS2014_5592} where it allows to remove the bias arising from mini-batching under the assumption that the covariance~$\Sigma_\vectx$ is constant. As discussed in Section~\ref{sec:Langevin}, the effect of the stochastic estimator~\eqref{eq:hypothese} of the force in Langevin dynamics is, at dominant order, to add an unknown contribution to the diffusion coefficient in front of the Brownian motion, which we denote by $\sqrt{2} A_{\dt,n}(\tht)^{1/2}$, with
\begin{equation}
A_{\dt,n}(\tht) = \Gamma + \frac{\eps(n) \dt}{2} \Sigma_\vectx(\tht).
\label{eq:A}
\end{equation}
Note that  $A_{\dt,n}(\tht) \in \R^{d\times d}$ is an unknown positive definite symmetric matrix. The idea behind AdL is to modify the effective Langevin dynamics so that it admits~$\pi(\cdot|\vectx)$ as an invariant probability measure (more precisely, that it admits an invariant probability measure whose marginal distribution in the~$\tht$ variable is~$\pi(\cdot|\vectx)$). The friction matrix, denoted here by $\xi \in \R^{d\times d}$, is no longer a constant, but a dynamical variable that adjusts itself to the effective noise resulting from mini-batching. The unknowns in the dynamics are therefore~$(\theta,p,\xi) \in \Xi = \Theta \times \R^d \times \R^{d \times d}$. Denoting by $[\xi]_{i,j}$ the components of~$\xi$, AdL can be written as
\begin{equation}
\left\{ \begin{aligned}
d\tht_t & = p_t \,dt, \\
dp_t & =\left(\nabla_\theta( \log\pi(\tht_t|\vectx) - \xi_t  p_t \right)dt +\sqrt{2} A_{\dt,n}(\tht_t)^{1/2} dW_t, \\
d [\xi_{t}]_{i,j} & = \frac{1}{\eta} \left(p_{i,t} p_{j,t} -  \delta_{i,j} \right)dt, \quad 1 \leq i , j \leq d, 
\end{aligned} \right.
\label{AdL}
\end{equation}
where $\eta$ is a positive scalar which sets the timescale for the evolution of the friction matrix (see Remark~\ref{remark:eta} below for more general choices). Let us emphasize once again that we consider that each degree of freedom has mass $1$ for simplicity, but our analysis can be easily generalized to non trivial mass matrices, or even to non-quadratic kinetic energies as the ones considered in~\cite{ST18}. 

\begin{remark}
	\label{rem:xi}
	If the initial condition $\xi_0$ is symmetric, then the matrix $\xi_t$ is symmetric for all $t$. In any case, the quantities $\xi_{i,j,t} - \xi_{j,i,t}$ are constant. This shows that it suffices to introduce $(d+1)d/2$ new variables $([\xi]_{i,j})_{1 \leq i  \leq j \leq d}$ to simulate AdL. However, for Lemma~\ref{lemmeADL} below, it is more convenient to write out statements and proofs with all variables $([\xi]_{i,j})_{1 \leq i  , j \leq d}$.
\end{remark}

\begin{remark}
	\label{remark:eta}
	It is possible to formulate AdL for with different timescale parameters for the evolution of the components of the friction matrix, as defined by a symmetric matrix $\eta = \eta^T  \in \R^{d\times d}$ with positive entries. In this case, each element $i,j$ of $\xi$ follows the dynamics
	\[
	d[\xi_{t}]_{i,j} = \dps\frac{1}{\eta_{i,j}} \left(p_{i,t}p_{j,t} - \delta_{i,j} \right) dt.
	\]
\end{remark}

Let us conclude this general presentation of AdL by recalling the motivation for the dynamics on the friction variable. We assume for this discussion that $d=1$ in order to simplify the presentation. One way to understand the intuition behind AdL is to see~\eqref{AdL} as the superposition of a Hamiltonian dynamics (which preserves the energy $-\log\pi(\tht|\vectx)+ |p |^2/2$) and the following elementary dynamics:
\[
\left\{ \begin{aligned}
dp_t & = -\xi_t p_t \,dt + \sqrt{2}A_{\dt,n}^{1/2} dW_t,\\
d \xi_t & = \frac{1}{\eta}\left(p_t^2 -  1 \right) dt.
\end{aligned} \right.
\]
Note that we write~$A_{\dt,n}^{1/2}$ here instead of~$A_{\dt,n}(\tht)^{1/2}$ since the value of~$\tht$ does not change for this subdynamics. Knowing that the marginal distribution over the variable $p$ of the invariant probability measure should be a Gaussian of variance $1$ (see~\cite{NIPS2014_5592}), the idea is to keep the right balance between the friction $\xi$ and the fluctuation $A_{\dt,n}$. Given that the strength of the fluctuation is unknown, the friction is adjusted so that the average kinetic energy is fixed to its target value, \textit{i.e.} $\mathbb{E}(|p|^2) = 1$. More precisely, if $|p|^2 > 1$, the kinetic energy is larger than what it should be, so the friction is increased, which ends up decreasing $p$. We can use the same line of argument for the opposite case. In fact, the dynamics of the additional variable $\xi$ follows a negative feedback loop control as in the Nosé--Hoover thermostat; see~\cite{Nose84} and~\cite{Hoover85}.

\subsection{Adaptive Langevin dynamics for gradient estimators with constant covariance}
\label{sec:AdL_cst}

We discuss more precisely in this section the properties of AdL under the crucial assumption that the covariance of the gradient estimator is constant~:
\begin{equation}
\Sigma_\vectx(\tht) = \Sigma_\vectx.
\label{eq:assConst}
\end{equation}
This implies in particular that the matrix~$A_{\dt,n}$ defined in~\eqref{eq:A} is constant. The limitations of this assumption are discussed more precisely in Section~\ref{subsec:Covariance}.

We start by discussing invariant probability measures of AdL in Section~\ref{sec:ppties_AdL_cst_covariance}. We next present in Section~\ref{sec:numerical_scheme_AdL} numerical schemes to integrate AdL, together with error estimates on the bias of their invariant measures. These results are numerically illustrated in Section~\ref{sec:numerics_AdL} for the Gaussian model of Section~\ref{sec:gaussian_model}, for which the fundamental assumption~\eqref{eq:assConst} is satisfied. Let us emphasize here again that, in contrast to various works in the literature, we do not necessarily assume that~$n$ is much larger than~1 and that the random variable in~\eqref{eq:hypothese} follows a Gaussian distribution.

\subsubsection{Invariant probability measure of AdL}
\label{sec:ppties_AdL_cst_covariance}

The generator of the stochastic dynamics~\eqref{AdL} can be decomposed as
\[
\mathcal{L}_{\rm AdL, \Sigma_\vectx}  =  \cLham +   \cLFD  +  \mathcal{L} _{\mathrm{NH}},\]
where
\[ \mathcal{L} _{\mathrm{NH}} =-p^T(\xi - A_{\dt,n}) \nabla_p+  \frac{1}{\eta}\sum\limits_{1 \leq i , j\leq d}(p_i p_j - \delta_{i,j}) \partial_{[\xi]_{i,j}} ,\]
and 
\[ \cLham = p^T  \nabla_\tht+ \nabla_\tht (\log \pi(\cdot|\vectx) )^T \nabla_p, \qquad \cLFD = A_{\dt,n}:\nabla^2_p - p^TA_{\dt,n}\nabla_p.\]
We recall the following result on the invariant probability measure of~\eqref{AdL}; see Section~2 in~\cite{MR4099815}.

\begin{lemma}
  \label{lemmeADL}
  Suppose that~\eqref{eq:assConst} holds. Then the dynamics~\eqref{AdL} admits the invariant probability measure 
  \begin{equation}
    \nu(d\tht\,dp\,d \xi) = \pi(\tht| \emph{\vectx})\tau(dp)\rho(d\xi)\, d\tht ,
    \label{piAdL}
  \end{equation}
  where $ \tau(d p)$ is defined in~\eqref{eq:tau} and
  \[
  \rho(d \xi) = \prod\limits_{1 \leq i , j \leq d} \sqrt{\frac{\eta}{2 \pi}} \exp\left(- \displaystyle\frac{\eta}{2}\left(\xi_{i,j} - \left[A_{\dt,n}\right]_{i,j}\right)^2 \right) \,d[\xi]_{i,j},
  \]
  where~$\left[A_{\dt,n}\right]_{i,j}$ is the~$(i,j)$ component of~$A_{\dt,n}$.
\end{lemma}

Lemma~\ref{lemmeADL} suggests that, as long as assumption~\eqref{eq:assConst} is satisfied, sampling a probability measure using Adaptive Langevin dynamics is not affected by the mini-batching procedure to estimate the gradient of the log-likelihood in the~$\tht$ variable. The marginal distribution of~\eqref{piAdL} in the variable $\tht$ is indeed the target distribution $\pi$, whatever the value of~$A_{\dt,n}$. This shows that AdL can indeed adjust the friction in order to compensate fluctuations of arbitrary constant magnitude. We recall the proof of Lemma~\ref{lemmeADL} because we use similar computations in the proof of Theorem~\ref{th:1} below.

\begin{remark}
	\label{rem:nu_ergodic}
	Let us emphasize that Lemma~\ref{lemmeADL} states that~$\nu$ is invariant by AdL. However, the dynamics cannot be ergodic for this measure since $[\xi_t]_{i,j} - [\xi_t]_{j,i}$ remains constant, whereas~$[\xi]_{i,j}$ and~$[\xi_t]_{j,i}$ are independent under the probability measure~\eqref{piAdL}. The dynamics can therefore at best be ergodic for the restriction of $\nu$ onto the sub-manifold
	\[
	\mathscr{S}(\xi_0) = \Theta \times \R^d \times \left\{ \xi \in \R^{d \times d} \, \Big| \, [\xi]_{i,j} - [\xi]_{j,i} = [\xi_0]_{i,j} -[\xi_0]_{j,i} \right\},
	\]
	which is determined by the initial condition~$\xi_0$. Such an ergodicity result is however not trivial at all since there are~$d(d+1)$ independent degrees of freedom in the symmetric matrix~$\xi_t$, while the noise acting on the momentum variable~$p$ is only of dimension~$d$. The stochastic dynamics is therefore highly degenerate. In any case, the important point here is that the marginal in the $\tht$-variable of the projected measure is~$\pi(\cdot |\emph{\vectx})$ whatever the distribution~$\rho$ in~\eqref{piAdL}.
\end{remark}

\begin{proof}
	We follow the approach in Section~2 of~\cite{MR4099815}. It suffices to show that, for all smooth and compactly supported functions $\phi$,  \begin{equation}
	\int_{\Xi} \mathcal{L}_{\rm AdL, \Sigma_\vectx} \phi \,d\nu = \int_{\Xi} \phi \left(\cL_{\rm AdL, \Sigma_\vectx}^* \mathbf{1}\right) d\nu =0, 
	\label{eq:invar}
	\end{equation}
	where adjoints are taken in $L^2(\nu)$. Simple computations based on integration by parts show that $\partial^*_{\tht_i} = - \partial_{\tht_i} - \partial_{\tht_i}( \log \pi(\tht|\vectx))$, $\partial^*_{p_i} = - \partial_{p_i} + p_i$, $\partial^*_{ [\xi]_{i,j} } = - \partial_{ [\xi]_{i,j} } +   \eta \left( [\xi]_{i,j}  -[A_{\dt,n}]_{i,j}\right)$. We can then rewrite the generators~$\cLham$ and~$\cLFD$ as (see~\cite{MR4099815}):
	\begin{align}
	\cLham & = \sum\limits_{i=1}^d \partial_{p_i}^* \partial_{\tht_i} - \partial_{\tht_i}^*\partial_{p_i},
	\label{eq:lham} \\
	\cLFD & = - \nabla^*_pA_{\dt,n}\nabla_p = -\sum\limits_{1 \leq i ,j  \leq d} [A_{\dt,n}]_{i,j} \partial^*_{p_i}\partial_{p_j}.  
	\label{eq:cLfd}
	\end{align}
	Moreover, for $\phi,\psi$ two smooth and compactly supported functions, 
	\begin{equation}
	\begin{aligned}
	\int_\Xi (\mathcal{L} _{\mathrm{NH}} \phi ) \psi \, d\nu & = \sum\limits_{1 \leq i ,j  \leq d}  \int_\Xi -p_i([\xi]_{i,j} - [A_{\dt,n}]_{i,j}) \left(\partial_{p_j} \phi\right) \psi +  \frac{1}{\eta} (p_ip_j - \delta_{i,j}) \left(\partial_{[\xi]_{i,j}} \phi\right) \psi \, d\nu\\
	&=   \sum\limits_{1 \leq i ,j  \leq d}  \int_\Xi  - ([\xi]_{i,j} - [A_{\dt,n}]_{i,j}) \partial_{p_j}^* \left( p_i\psi\right) \phi  +  \frac{1}{\eta}  (p_ip_j - \delta_{i,j}) \left( \partial_{[\xi]_{i,j}}^* \psi \right) \phi \, d \nu\\
	& = -  \int_\Xi (\mathcal{L} _{\mathrm{NH}} \phi ) \psi \, d\nu.
	\end{aligned}
	\label{eq:cLnh}
	\end{equation}
	It is then clear that $\cLham$ and $\mathcal{L} _{\mathrm{NH}}$ are antisymmetric, while $\cLFD$ is symmetric. Moreover, $\mathcal{L} _{\mathrm{NH}}  \mathbf{1} = \cLham \mathbf{1}= \cLFD  \mathbf{1}= 0$. The invariance of $\nu$ therefore follows from~\eqref{eq:invar}.
\end{proof}

The mathematical properties of AdL are investigated in~\cite{herzog2018exponential} and~\cite{MR4099815} in the case when $\Sigma_{\vectx} = \sigma^2 \mathrm{I}_d$ with~$\sigma^2$ constant, and the friction is scalar valued (in which case the invariant probability measure provided by Lemma~\ref{lemmeADL} is in fact the only invariant probability measure). The main contributions of~\cite{MR4099815} are the following: (i) the exponential convergence of the law of the process encoded by the convergence of the semi-group $\mathrm{e}^{t\cL}$ is proved using the hypocoercive approach of~\cite{herau:hal-00004498} and~\cite{MR2576899,MR3324910}; (ii) a central limit theorem for time averages along one realization of the dynamics is derived with bounds on the asymptotic variance depending on the parameters~$\eta$ and~$\Gamma$ of the dynamics. If $A_{\dt,n} \approx \Gamma$ (which is the case when $\Delta t$ is sufficiently small and~$n$ is sufficiently large), the mathematical analysis suggests to fix~$\Gamma$ and~$\eta$ of order~1.

\subsubsection{Numerical scheme}
\label{sec:numerical_scheme_AdL}

We now construct a numerical scheme for which AdL in~\eqref{AdL} is the effective dynamics at dominant order in~$\dt$ and~$\eps(n)\dt$. Concretely, we replace the matrix $A_{\dt,n}$ by its expression~\eqref{eq:A} in the SDE~\eqref{AdL} and consider the symmetric splitting scheme introduced in~\cite{MR4099815}, which is based on decomposing~\eqref{AdL} into the following four elementary SDEs (the variables which are evolved are indexed by~$t$, while the ones which remain constant do not have any subscript):
\begin{align}
d\tht_t & = p \,dt, \label{eq:elementary_SDE_tht} \\
dp_t & = - \xi p_t \,dt + \sqrt{2} \Gamma^{1/2} \, dW_{1,t}, \label{eq:elementary_SDE_OU} \\
dp_t & = \gradT (\log \pi(\tht|\vectx)) \,dt +\sqrt{\eps(n) \dt} \Sigma_{\vectx}(\tht)^{1/2} \, dW_{2,t}, \label{eq:elementary_SDE_mb} \\
d[\xi_t]_{i,j}  & = \frac{1}{\eta} \left(p_{i} p_{j} -  \delta_{i,j} \right)dt, \qquad 1 \leq i \leq j \leq d, \label{eq:elementary_SDE_xi}
\end{align}
where~$W_{1,t},W_{2,t}$ are two independent standard $d$-dimensional Brownian motions. The elementary ordinary differential equations~\eqref{eq:elementary_SDE_tht} and~\eqref{eq:elementary_SDE_xi} can be trivially integrated. The elementary SDE~\eqref{eq:elementary_SDE_OU} is an Ornstein--Uhlenbeck process that can be analytically integrated in law as
\[
p^{m+1} = \mathrm{e}^{-\dt \xi} p^m + \sigma_{\xi, \Gamma, \dt} G^m,
\qquad
\sigma_{\xi, \Gamma, \dt}^2  = 2 \int\limits_{0}^{\dt} \mathrm{e}^{-s \xi} \Gamma \mathrm{e}^{-s \xi } \,ds,
\]
where $(G^m)_{m \geq 0}$ is a family of i.i.d.~standard $d$-dimensional Gaussian random variables. If there exists $M$ such that $\xi M + M \xi = \Gamma$, then $\sigma_{\xi, \Gamma, \dt}^2 = 2\left( M - \mathrm{e}^{-\dt \xi } M  \mathrm{e}^{-\dt \xi }\right).$ In particular, for $\Gamma = \gamma \mathrm{I}_d$ and when~$\xi$ is invertible, one can choose $M = \gamma \xi^{-1}/2$ in which case $\sigma_{\xi, \Gamma, \dt}^2  = \gamma\xi^{-1}(\mathrm{I}_d - \mathrm{e}^{-2 \dt \xi})$. When~$\xi$ is singular or close to singular, the latter formula has to be understood through a limiting procedure based on spectral calculus, see~\cite{MR4099815}.

The elementary SDE~\eqref{eq:elementary_SDE_mb} is integrated as
\begin{equation}
\label{eq:elementary_scheme_mb}
p^{m+1} = p^m + \dt \widehat{F_n} \left( \tht \right) = p^m + \dt \gradT (\log \pi (\tht |\vectx)) + \dt \sqrt{\eps(n)} \Sigma_{\vectx}(\tht)^{1/2} Z_{\vectx, \Nd, n}.
\end{equation}
Since the random variable~$Z_{\vectx, \Nd, n}$ has mean~0 and identity covariance by construction, the equality~\eqref{eq:estim_Q} in Appendix~\ref{app_2} shows that the numerical scheme~\eqref{eq:elementary_scheme_mb} is weakly consistent with~\eqref{eq:elementary_SDE_mb}, with an error of order~$(1+\eps(n)^{3/2})\dt^3$ over one time step, even if $Z_{\vectx, \Nd, n}$ is not Gaussian. In the case where $\mathbb{E}[Z_{\vectx, \Nd, n}^3] = 0$, the error over one time step is of order $(1+\eps(n))\dt^3$.

The numerical scheme we consider is finally obtained by a Strang splitting where~\eqref{eq:elementary_SDE_mb} is updated in the central step of the algorithm, in order to compute the force only once per time step and avoid its storage. The order in which the remaining elementary dynamics are integrated is unimportant for our purposes, although some orderings may be better than others, in particular in some limiting regimes where one of the parameters goes to~0 or infinity; see~\cite{LeSh16} for an extensive discussion in the context of AdL. Fixing $\Gamma = \gamma \mathrm{I}_d$, the numerical scheme reads as follows:
\begin{equation}
\left\{ \begin{aligned}
p^{m+\frac{1}{2}} & = \rme^{-\Delta t \xi^m/2} p^m + \left[ \gamma( \xi^m)^{-1}\left(\mathrm{I}_d-\rme^{- \Delta t \xi^m} \right) \right]^{1/2} G^m, \\
\xi^{m+\frac{1}{2}} & = \xi^m +\frac{ \Delta t }{2 \eta} \left( p^{m+\frac{1}{2}}\left(p^{m+\frac{1}{2}}\right)^T - \mathrm{I}_d \right),\\
\tht^{m+\frac{1}{2}} & = \tht^m + \frac{ \Delta t }{2} p^{m+\frac{1}{2}}, \\
\widetilde{p}^{m+\frac{1}{2}} & = p^{m+\frac{1}{2}} + \dt \widehat{F_n} \left( \tht^{m+\frac{1}{2}} \right),\\
\tht^{m+1} & =  \tht^{m+\frac{1}{2}}  +   \frac{ \Delta t }{2} \widetilde{p}^{m+\frac{1}{2}}, \\
\xi^{m+1} & =  \xi^{m+\frac{1}{2}} + \frac{ \Delta t }{2 \eta} \left(\widetilde{p}^{m+\frac{1}{2}} \left(\widetilde{p}^{m+\frac{1}{2}}\right)^T -  \mathrm{I}_d \right),\\
p^{m+1} & = \rme^{-\Delta t \xi^{m+1}/2} \widetilde{p}^{m+\frac{1}{2}} + \left[\gamma(\xi^{m+1})^{-1} \left(\mathrm{I}_d-\rme^{-\Delta t \xi^{m+1}}\right)\right]^{1/2}  G^{m+\frac{1}{2}},
\end{aligned} \right.
\label{eq:adl_scheme}
\end{equation}
where $(G^m)_{m \geq 0}$ and $(G^{m+\frac{1}{2}})_{m \geq 0}$ are two independent families of i.i.d.~standard Gaussian random variables. Note that it is possible to work only with the additional variables $(\xi_{i,j})_{1 \leq i  \leq j \leq d}$ since the updates of~$\xi_{i,j}$ and~$\xi_{j,i}$ in~\eqref{AdL} are the same. The initial conditions for~$\xi^0$ is~$\gamma \mathrm{Id}$, while the components of~$\tht^0,p^0$ are initialized to~0. Of course, more educated choices can be considered depending on the system at hand (\emph{e.g.} restarting from values sampled by SGLD or some Langevin-like dynamics).

\begin{remark}
  \label{rem:other_version_num}
  Other versions of the numerical scheme~\eqref{eq:adl_scheme} can be considered, in particular the one used in~\cite{MR4099815}, for which~$\xi \in \R$. In this case, \eqref{eq:elementary_SDE_xi} should be replaced by 
  \begin{equation}
    \label{eq:xi_scalar}
    d\xi_t= \frac{1}{\eta} \left(p^T p -  d \right)dt. 
  \end{equation}
  Another option to reduce the number of additional variables is to consider the variable~$\xi$ as a diagonal matrix with entries~$\xi_i$. In this case, \eqref{eq:elementary_SDE_xi} has to be replaced with 
  \begin{equation}
    \label{eq:xi_diag}
    d[\xi_t]_{i}= \frac{1}{\eta} \left(p_{i}^2  -  1 \right)dt, \qquad 1 \leq i \leq d. 
  \end{equation}
  The numerical schemes for these two cases can be obtained by an obvious modification of the right hand side of the update formulas for~$\xi^{m+\frac12}$ and~$\xi^{m+1}$ in~\eqref{eq:adl_scheme}.
\end{remark}

Using the Baker--Campbell--Hausdorff formula (see for instance Section~III.4.2 in~\cite{hairer-lubich-wanner-06}), as done in~\cite{MR3463433} and~\cite{LeSh16} for instance, one can prove that the evolution operator~$\widehat{P}_{\dt,n}$ of the numerical scheme~\eqref{eq:adl_scheme} satisfies (see Appendix~\ref{app_3})
\begin{equation}
\widehat{P}_{\dt,n} = \mathrm{e}^{\dt \mathcal{L}_{\rm AdL, \Sigma_\vectx} } + \mathcal{O}\left((1+\eps(n)^{3/2})\dt^3\right).
\label{eq:P_adL_erreur}
\end{equation}
Then, by following the strategy of proof of Theorem~3.3 in~\cite{lelievre_stoltz_2016}, the following error estimate holds.

\begin{proposition}
  \label{prop:AdL_constant_Sigma}
  Assume that~\eqref{AdL} and~\eqref{eq:adl_scheme} both admit a unique invariant probability measure, respectively denoted by~$\nu_{\xi_0}$ and~$\nu_{\xi_0,\dt,n}$. Then, for any smooth function~$\varphi$ with compact support, there exist~$C \in \R_+$ and~$\dt_\star > 0$ such that, for any~$0 < \dt \leq \dt_\star$, 
  \begin{equation}
    \label{eq:erreur_adl_scheme}
    \begin{aligned}
      \left| \int_{\Xi} (\mathcal{L}_{\mathrm{AdL}, \Sigma_\emph{\vectx}}\varphi)(\tht, p, \xi) \, \nu_{\xi_0,\dt,n}(d\tht \, dp \, d\xi) - \int_{\Xi} (\mathcal{L}_{\rm AdL, \Sigma_\emph{\vectx}}\varphi)(\tht, p, \xi) \, \nu_{\xi_0}(d\tht \, dp \, d\xi) \right| &\\
      \qquad \leq C\left(1+\eps(n)^{3/2}\right)\dt^2.&
    \end{aligned}
  \end{equation}
\end{proposition}

In view of~\eqref{eq:erreur_adl_scheme}, the order of magnitude of the bias on the invariant probability measure is of order~$\eps(n)^{3/2}\dt^2$, which is smaller than the bias~$\eps(n)\dt$ obtained with SGLD or Langevin (recall Propositions~\ref{prop:SGLD_error} and~\ref{prop:Langevin_error}). Even if the estimate is stated for averages of functions~$\mathcal{L}_{\rm AdL, \Sigma_\vectx}\varphi$, this bound allows to uniquely characterize the bias on the invariant measure. To state it for any smooth function~$\varphi$ (and not just functions in the range of the operator~$\mathcal{L}_{\rm AdL, \Sigma_\vectx}$), one would need results ensuring that the inverse of~$\mathcal{L}_{\rm AdL, \Sigma_\vectx}$ is well defined, and that it stabilizes spaces of smooth functions with polynomial growth and whose derivatives also grow at most polynomially. This would be the counterpart of similar estimates for Langevin like dynamics (see~\cite{Kopec14,KopecLangevin}). Such results are however not available, and are presumably rather difficult to prove. At the moment, as discussed at the end of Section~\ref{sec:ppties_AdL_cst_covariance}, the only available ergodicity results on AdL concern the exponential convergence of the semigroup, which ensures that the inverse of the generator is well defined on certain functional spaces, such as~$L^2(\nu)$ (see~\cite{herzog2018exponential,MR4099815}). 

Note also that the error estimate~\eqref{eq:erreur_adl_scheme} holds even when $Z_{\vectx, \Nd, n}$ is not Gaussian (namely for small values of $n$). When $\mathbb{E}[Z_{\vectx, \Nd, n}^3]= 0$, the error estimate~\eqref{eq:erreur_adl_scheme} holds with $C(1+\eps(n))\dt^2$ on the right hand side. When $Z_{\vectx, \Nd, n}$ is Gaussian, the integration~\eqref{eq:elementary_scheme_mb} is in fact exact in law. In any case, the analysis we made crucially depends on the fact that~\eqref{eq:assConst} holds. When it does not hold, there is an additional bias on the invariant measure due to mini-batching of order~$\varepsilon(n)\dt$, much larger than the error of order~$\left(1+\eps(n)^{3/2}\right)\dt^2$ in~\eqref{eq:erreur_adl_scheme}. This is quantified in Section~\ref{subsec:Covariance}.

\subsubsection{Numerical illustration for Gaussian likelihoods}
\label{sec:numerics_AdL}

\begin{figure}
	\centering
	\includegraphics[scale = 0.6]{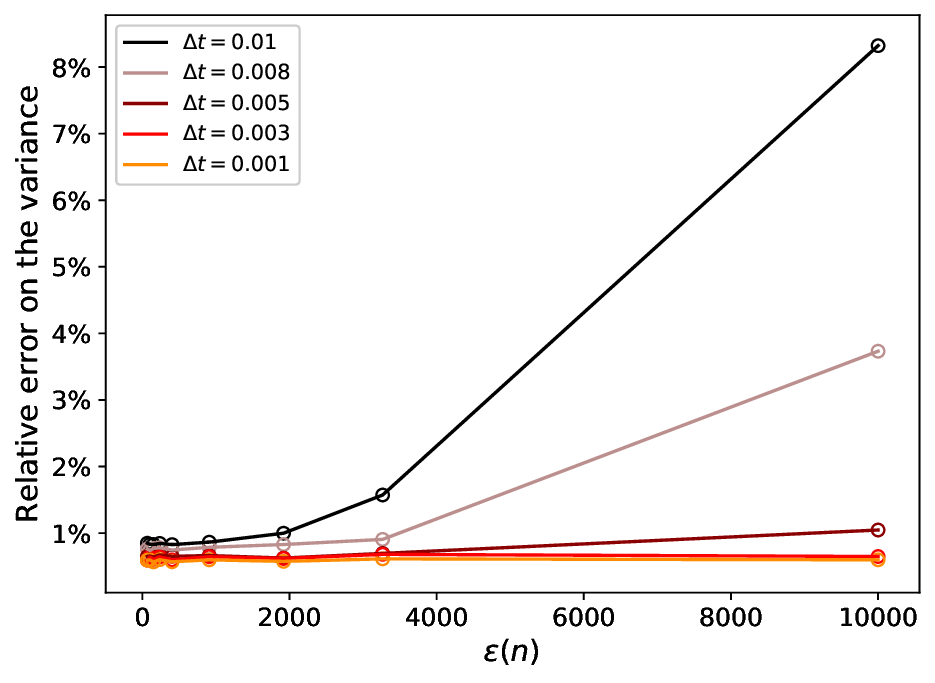}
	\caption{Relative error on the variance of the posterior distribution for various values of $n$ using AdL when the elementary likelihood is Gaussian, and sampling is performed without replacement.} \label{fig:erreur}
\end{figure}

We consider the same setting as in Section~\ref{sec:gaussian_model}. The covariance $\Sigma_\vectx$, given by~\eqref{eq:var_hat_F}, is constant for this model, so that assumption~\eqref{eq:assConst} indeed holds. We run the numerical scheme~\eqref{eq:adl_scheme} with $\gamma = 1$ for an integration time~$T = 10^6$. Similar computations have already been performed in~\cite{NIPS2014_5592} and~\cite{LeSh16}. In Figure~\ref{fig:erreur}, we plot the error on the variance of the marginal posterior distribution in the~$\tht$ variable with respect to~$\eps(n)$. First, we note that, for $\dt$ small enough (namely $\dt \leq 0.005$), the bias is not affected by the value of~$\eps(n)$ (even for the very large value~$\Nd^2$ obtained when~$n=1$), and hence by the size of the mini-batch. The bias is therefore simply due to the discretization time step. For large values of~$\dt$ (namely $\dt \geq 0.008$ in our experiments), the error is affected by the value of~$n$. This is probably due to the fact that this rather large value of $\dt$ is out of the regime where the asymptotic analysis for the bias holds. In any case, a comparison with Figure~\ref{fig:SGLD_gauss} shows that the error obtained with AdL is much smaller than the one obtained with SGLD~\eqref{SGLD} or the numerical scheme~\eqref{GLA_minibatched} (which corresponds to Langevin dynamics with mini-batching), even with much larger time steps. 

\subsection{Impact of a non constant covariance matrix}
\label{subsec:Covariance}

The bias analysis in the previous section does not hold in the case of a non constant covariance matrix $\Sigma_{\vectx}$. We first quantify in Section~\ref{sec:bias_adl} the bias due to mini-batching when assumption~\eqref{eq:assConst} is not satisfied. We illustrate our analysis in Section~\ref{mix_adl} with a numerical example where the elementary likelihood is a mixture of Gaussians. In this case, $\Sigma_\vectx(\tht)$ genuinely depends on~$\tht$. 

\subsubsection{Mini-batching bias for Adaptive Langevin dynamics and non constant covariance}
\label{sec:bias_adl}

We consider the situation where $\Sigma_\vectx(\tht)$ genuinely depends on $\tht$, and prove that the bias on the invariant measure is of order~$\eps(n)\dt + \dt^2$. This is done in two steps. We first characterize the bias of the continuous Adaptive Langevin dynamics with covariance matrix~$\Sigma_\vectx(\tht)$ instead of the constant covariance matrix 
\begin{equation}
\overline{\Sigma}_\vectx = \int_{\Theta} \Sigma_{\vectx}(\tht) \pi(\tht |\vectx) \, d\tht
\label{eq:sigma_bar}
\end{equation}
obtained by averaging over the target probability distribution. We then combine this estimate with~\eqref{eq:erreur_adl_scheme}.

The motivation for introducing the particular constant matrix~\eqref{eq:sigma_bar} is discussed after Lemma~\ref{lemme:f_adl} below. The generator of~\eqref{AdL} can be written in terms of~$\overline{\Sigma}_\vectx$ as
\begin{equation}
\cL_{\mathrm{AdL, \Sigma_\vectx }} = \cL_{\mathrm{AdL, \overline{\Sigma}_\vectx }}  + \eps(n)\dt \widetilde{\cL}_{\Sigma_{\vectx} - \overline{\Sigma}_\vectx},
\qquad
\widetilde{\cL}_\mathscr{M} = \mathscr{M} : \nabla_p^2.
\label{eq:bias_adl}
\end{equation}
Since $\overline{\Sigma}_\vectx $ does not depend on $\tht$, the probability measure~$\nu$ defined in~\eqref{piAdL} is an invariant probability measure of the dynamics with generator~$\cL_{\mathrm{AdL, \overline{\Sigma}_\vectx }} $. It is then easy to prove that the extra bias due to mini-batching on the invariant measure of~\eqref{AdL} when $\Sigma_{\vectx}$ is not constant is of order~$\eps(n)\dt$.

\begin{proposition}
  \label{prop:AdL_non_constant_Sigma}
  Assume that the continuous dynamics~\eqref{AdL} for~$\Sigma_{\emph{\vectx}}$ depending on the~$\tht$ variable, and the same dynamics~\eqref{AdL} with constant covariance set to~$\overline{\Sigma}_\emph{\vectx}$, both admit unique invariant probability measures, respectively denoted by~$\nu_{\xi_0}$ and~$\overline{\nu}_{\xi_0}$; and that the resolvent~$\cL_{\mathrm{AdL, \overline{\Sigma}_\emph{\vectx} }}^{-1}$ is bounded on the subspace of functions in~$L^2(\overline{\nu}_{\xi_0})$ with average~0 with respect to~$\overline{\nu}_{\xi_0}$. Then, for any smooth function~$\varphi$ with compact support, 
\begin{equation}
\begin{aligned}
&\int_{\Xi} \left(\cL_{\mathrm{AdL, \Sigma_\emph{\vectx} }}\varphi\right)(\tht, p, \xi)\, \nu_{\xi_0}(d\tht \, dp \, d\xi)-\int_{\Xi} \left(\cL_{\mathrm{AdL, \Sigma_\emph{\vectx} }} \varphi\right)(\tht, p, \xi)\, \overline{\nu}_{\xi_0}(d\tht \, dp \, d\xi)\\
& \qquad  = -\eps(n) \int_{\Xi} \left(\cL_{\mathrm{AdL, \Sigma_\emph{\vectx} }} \varphi\right)(\tht, p, \xi) f_{\mathrm{AdL, \Sigma_{\emph{\vectx}}}}(\tht, p, \xi) \, \overline{\nu}_{\xi_0}(d\tht \, dp \, d\xi),
\end{aligned}
\label{erreuraAdl}
\end{equation}
where (adjoints being taken on~$L^2(\overline{\nu}_{\xi_0})$)
\begin{equation}
  \label{eq:def_f_AdL_Sigma}
  f_{\mathrm{AdL},\Sigma_{\emph{\vectx}}} = \left(-\cL_{\mathrm{AdL, \overline{\Sigma}_\emph{\vectx} }}^*\right)^{-1} \widetilde{\cL}_{\Sigma_{\emph{\vectx}} - \overline{\Sigma}_\emph{\vectx}}^* \mathbf{1}.
\end{equation}
\end{proposition}

The ergodicity assumption on the dynamics are not trivial statements, as discussed in Remark~\ref{rem:nu_ergodic}. The boundedness of the resolvent has been proved when~$\xi$ is scalar valued, see~\cite{MR4099815}. If this result could be extended to~$\cL_{\mathrm{AdL, \Sigma_\vectx }}$, with further estimates on the derivatives (similary to the ones discussed after Proposition~\ref{prop:AdL_constant_Sigma}), it would be possible to replace~$\cL_{\mathrm{AdL, \Sigma_\vectx }}\varphi$ by~$\varphi$ in~\eqref{erreuraAdl}.

\begin{proof}
  Note first that, by the assumed invariance of~$\nu_{\xi_0}$ by~\eqref{AdL}, it holds
  \[
  \int_{\Xi} \left(\cL_{\mathrm{AdL, \Sigma_\vectx }}\varphi\right)(\tht, p, \xi)\, \nu_{\xi_0}(d\tht \, dp \, d\xi) = 0.
  \]
  Moreover, using the invariance of~$\overline{\nu}_{\xi_0}$ under the dynamics~\eqref{AdL} with constant covariance set to~$\overline{\Sigma}_\vectx$,
  \[
  \begin{aligned}
  \int_{\Xi} \left(\cL_{\mathrm{AdL, \Sigma_\vectx }} \varphi\right)(\tht, p, \xi)\, \overline{\nu}_{\xi_0}(d\tht \, dp \, d\xi)
  & = \int_{\Xi} \left(\cL_{\mathrm{AdL, \overline{\Sigma}_\vectx }} \varphi\right)(\tht, p, \xi)\, \overline{\nu}_{\xi_0}(d\tht \, dp \, d\xi) \\
  & \quad + \varepsilon(n)\dt \int_{\Xi} \left(\widetilde{\cL}_{\Sigma_{\vectx} - \overline{\Sigma}_\vectx} \varphi\right)(\tht, p, \xi)\, \overline{\nu}_{\xi_0}(d\tht \, dp \, d\xi) \\
  & = \varepsilon(n)\dt \int_{\Xi} \varphi \, \widetilde{\cL}_{\Sigma_{\vectx} - \overline{\Sigma}_\vectx}^* \mathbf{1}\, d\overline{\nu}_{\xi_0} \\
  & = -\varepsilon(n)\dt \int_{\Xi} \left[\cL_{\mathrm{AdL, \overline{\Sigma}_\emph{\vectx} }}\varphi\right]\left[\left(-\cL_{\mathrm{AdL, \overline{\Sigma}_\emph{\vectx} }}^*\right)^{-1}\widetilde{\cL}_{\Sigma_{\vectx} - \overline{\Sigma}_\vectx}^* \mathbf{1}\right] d\overline{\nu}_{\xi_0},
  \end{aligned}
  \]
  where leads to the claimed equality in view of the definition of~$f_{\mathrm{AdL},\Sigma_{\vectx}}$.
\end{proof}

The estimate~\eqref{erreuraAdl} shows that the difference between~$\nu_{\xi_0}$ and~$\overline{\nu}_{\xi_0}$ is of order~$\eps(n)\dt$, with a magnitude related to the norm of~$f_{\mathrm{AdL, \Sigma_{\vectx}}}$ in~$L^2(\overline{\nu}_{\xi_0})$. The latter quantity, estimated in the following lemma, is typically smaller than similar quantities for Langevin like dynamics such as~\eqref{eq:f_mb_Langevin}.

\begin{lemma}
  \label{lemme:f_adl}
  Suppose that the same assumptions as in Proposition~\ref{prop:AdL_non_constant_Sigma} hold true, and that the marginal distributions of~$\overline{\nu}_{\xi_0}$ in the~$\tht$ and~$p$ variables are respectively~$\pi(\cdot|\emph{\vectx})$ and~$\tau$. Then, there exists $C \in \R_+$ such that
  \begin{equation}
    \label{eq:f_mb_AdL}
    \left\| f_{\mathrm{AdL, \Sigma_{\emph{\vectx}}}} \right\|_{L^2(\overline{\nu}_{\xi_0})} \leq C \left\| \Sigma_{\emph{\vectx}}-\overline{\Sigma}_\emph{\vectx}\right\|_{L^2(\pi)}.
  \end{equation}
\end{lemma}

Lemma~\ref{lemme:f_adl} provides a motivation for the choice of~$\overline{\Sigma}_\vectx$. Indeed, the error estimate in Lemma~\ref{lemme:f_adl} would be true with~$\overline{\Sigma}_\vectx$ replaced by any constant, positive, symmetric matrix~$M \in \R^{d \times d}$. The choice leading to the sharpest upper bound is the $L^2(\pi)$-orthogonal projection~$\overline{\Sigma}_\vectx$ of~$\Sigma_{\vectx}$ onto constant matrices. 

\begin{proof}
  Note first that $\widetilde{\cL}_{\Sigma_{\vectx} - \overline{\Sigma}_\vectx}^* \mathbf{1} = \left(\Sigma_{\vectx} - \overline{\Sigma}_\vectx \right) : \left(\nabla_p^{2}\right)^* \mathbf{1}$. The $(i,j)$-component of~$\left(\nabla_p^{2}\right)^* \mathbf{1}$ is $\partial_{p_i}^* \partial_{p_j}^* \mathbf{1} = p_ip_j - \delta_{i,j}$. Since
  \[
  \int_{\R^d} \left(p_ip_j - \delta_{i,j}\right)^2 \tau(p) \, dp = 1 + \delta_{i,j},
  \]
  we obtain, by a Cauchy--Schwarz inequality,
  \[
  \left\| \widetilde{\cL}_{\Sigma_{\vectx} - \overline{\Sigma}_\vectx}^* \mathbf{1} \right\|^2_{L^2(\overline{\nu}_{\xi_0})} \leq \left\|\Sigma_{\vectx} - \overline{\Sigma}_\vectx\right\|_{L^2(\pi)}^2 \sum_{i,j=1}^d \int_{\R^d} \left(p_ip_j - \delta_{i,j}\right)^2 \tau(p) \, dp = d(d+1) \left\|\Sigma_{\vectx} - \overline{\Sigma}_\vectx \right\|_{L^2(\pi)}^2.
  \]
  The conclusion then follows from the definition~\eqref{eq:def_f_AdL_Sigma} and the assumed boundedness of the resolvent.
\end{proof}

We finally deduce from~\eqref{eq:erreur_adl_scheme} and~\eqref{erreuraAdl} that the total error between the invariant probability measures for~\eqref{AdL} and~\eqref{eq:adl_scheme} is of order~$\eps(n)\dt + \dt^2$. However, as motivated by Lemma~\ref{lemme:f_adl}, the prefactor for the error term~$\eps(n)\dt$ is much smaller than the one for SGLD and Langevin dynamics, and depends only on the deviation of the covariance of the gradient from its average.

\begin{remark}
  \label{rem:adl_xi_cases}
  Using other versions of the numerical scheme~\eqref{eq:adl_scheme} (see Remark~\ref{rem:other_version_num}) affects the prefactor for the error term $\eps(n) \dt$. If we consider the case where the variable $\xi$ is a scalar, in other words, if we consider~\eqref{eq:xi_scalar} for the numerical scheme, $\overline{\Sigma}_{\emph{\vectx}}$ in~\eqref{eq:sigma_bar} should be replaced by~$s^* \mathrm{Id}_d$, with 
  \[
  s^* = \frac{1}{d}\int_\Theta \Tr \left( \Sigma_{\emph{\vectx}}(\tht) \right) \pi(\tht|\emph{\vectx})\, d \tht.
  \]
  In this case, the prefactor in front of the minibatching error term is larger than~$\| \Sigma_{\emph{\vectx}} - \overline{\Sigma}_{\emph{\vectx}} \| _{L^2(\pi)}$, as it is given by
  \[
  \| \Sigma_{\emph{\vectx}} - s^* \mathrm{I}_d \| _{L^2(\pi)} = \min_{s \in \R} \| \Sigma_{\emph{\vectx}} - s \mathrm{I}_d \| _{L^2(\pi)}.
  \]
  The latter quantity corresponds to the projection error onto the class of isotropic matrices.

  If we consider the variable~$\xi$ to be a diagonal matrix, \emph{i.e.} we consider~\eqref{eq:xi_diag} for the numerical scheme, $\overline{\Sigma}_{\emph{\vectx}}$ in~\eqref{eq:sigma_bar} should be replaced by a diagonal matrix with entries
  \[
  D_i^* = \int_\Theta \left[ \Sigma_{\emph{\vectx}}(\tht) \right] _{i,i}\pi(\tht|\emph{\vectx})\, d \tht ,  \qquad 1 \leq i \leq d.
  \]
  The upper bound in Lemma~\ref{lemme:f_adl} then corresponds to the projection error of~$\Sigma_{\emph{\vectx}}$ onto constant diagonal matrices:
  \[
  \| \Sigma_{\emph{\vectx}} - D^* \|_{L^2(\pi)} = \min_{(D_1,\dots,D_d) \in \R^d} \| \Sigma_{\emph{\vectx}} - \mathrm{diag}(D_1,\dots,D_d) \| _{L^2(\pi)}.
  \]
  
  By interpreting the various $L^2(\pi)$-norms of the difference between~$\Sigma_{\emph{\vectx}}$ and a constant matrix as the distance to finite dimensional subspaces of~$L^2(\pi)$ included in each other, one can conclude that
  \[
  \| \Sigma_{\emph{\vectx}} - \overline{\Sigma}_{\emph{\vectx}} \|_{L^2(\pi)} \leq \| \Sigma_{\emph{\vectx}} - D^* \|_{L^2(\pi)} \leq \| \Sigma_{\emph{\vectx}} - s^* \mathrm{I}_d \| _{L^2(\pi)},
  \]
  \emph{i.e.} the projection error obtained for isotropic matrices is the largest, followed by the projection error for diagonal matrices, the smallest projection error being obtained for genuine matrices. 
\end{remark}

\subsubsection{Mixture of Gaussians}
\label{mix_adl}

\begin{figure}
	\centering
	\begin{subfigure}{\textwidth}
		\centering
		\includegraphics[width=0.49\textwidth]{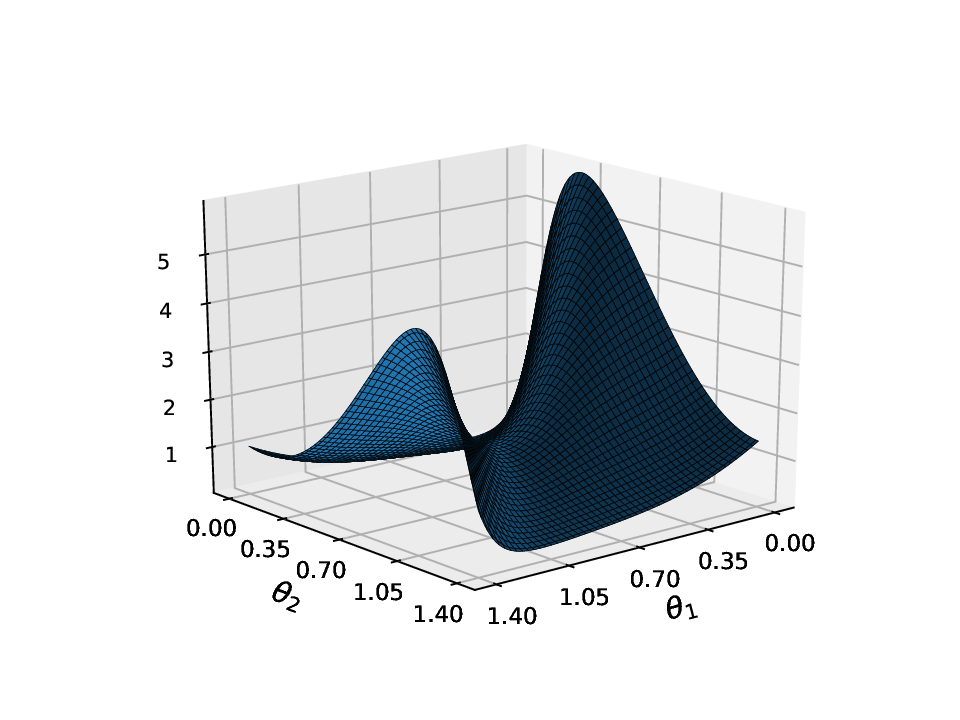}
		\hfill
		\includegraphics[width=0.49\textwidth]{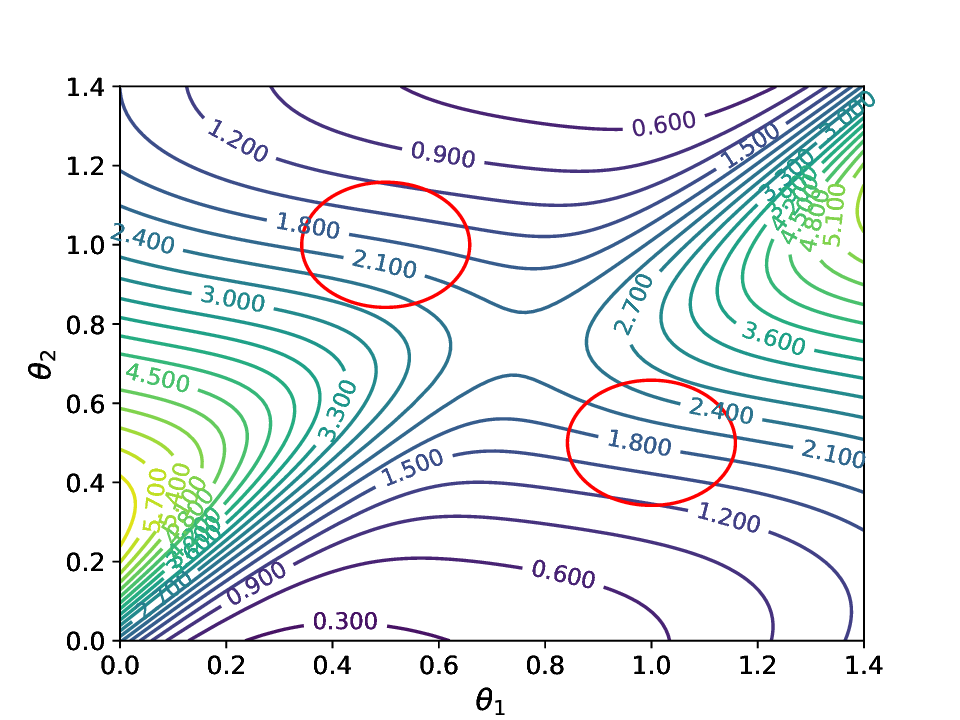}
		\caption[first caption.]{$\Sigma_{1,1}$}
	\end{subfigure}
	\begin{subfigure}{\textwidth}
		\centering
		\includegraphics[width=0.49\textwidth]{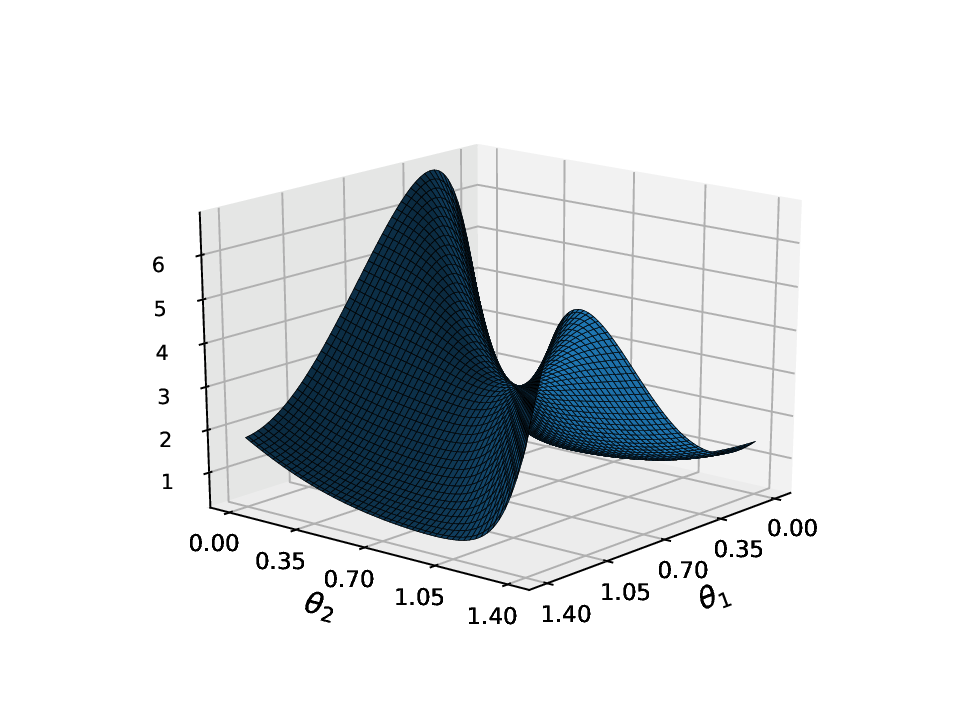}
		\hfill
		\includegraphics[width=0.49\textwidth]{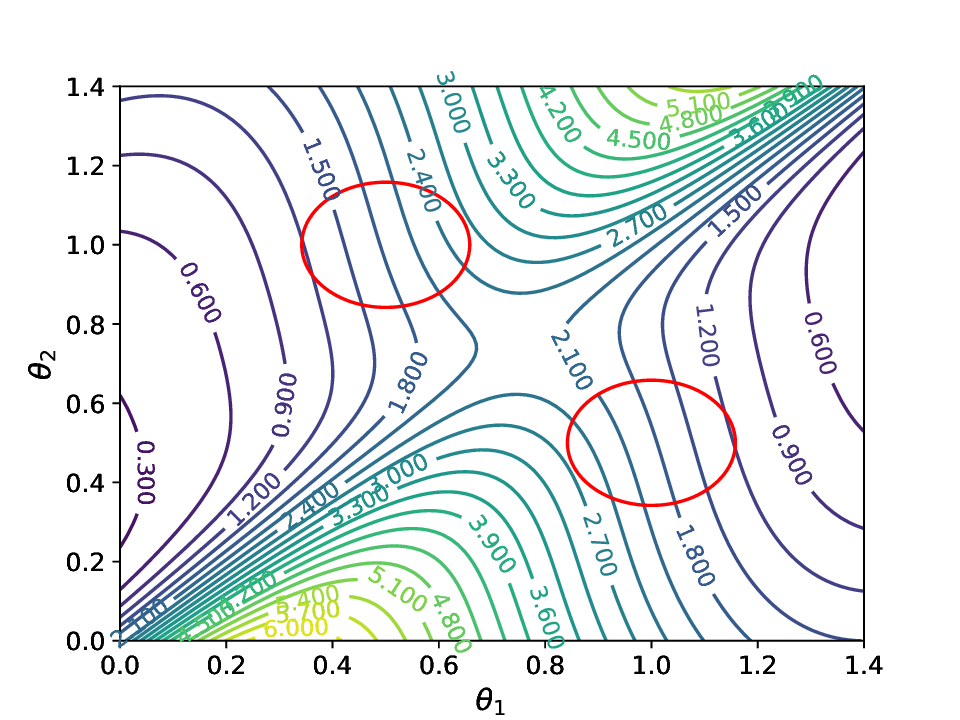}
		\caption[second caption.]{$\Sigma_{2,2}$}
	\end{subfigure}
	\begin{subfigure}{\textwidth}
		\centering
		\includegraphics[width=0.49\textwidth]{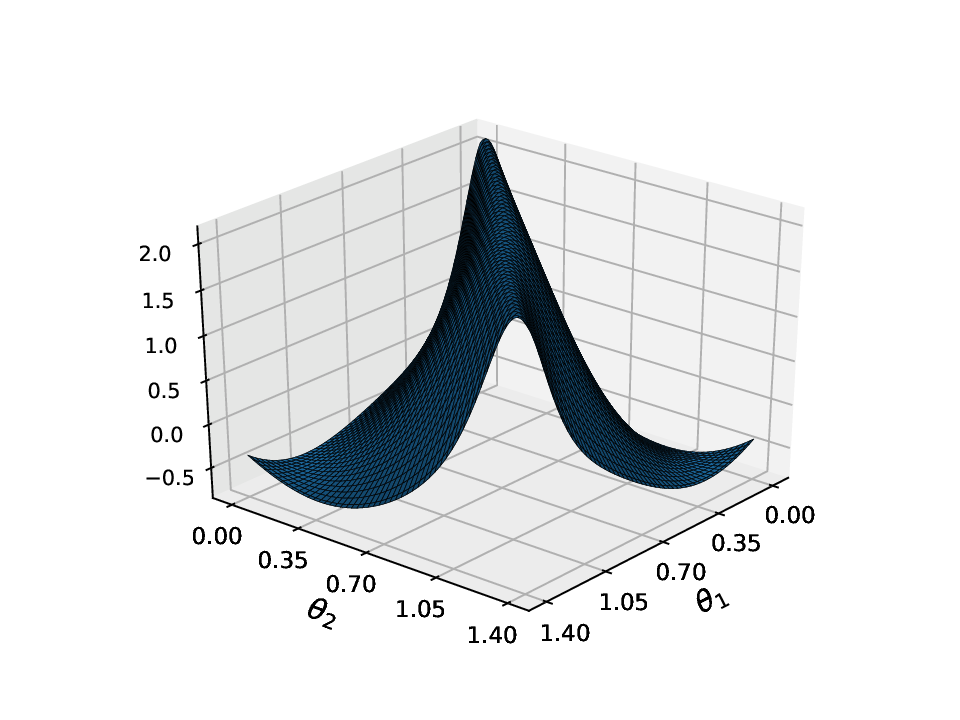}
		\hfill
		\includegraphics[width=0.49\textwidth]{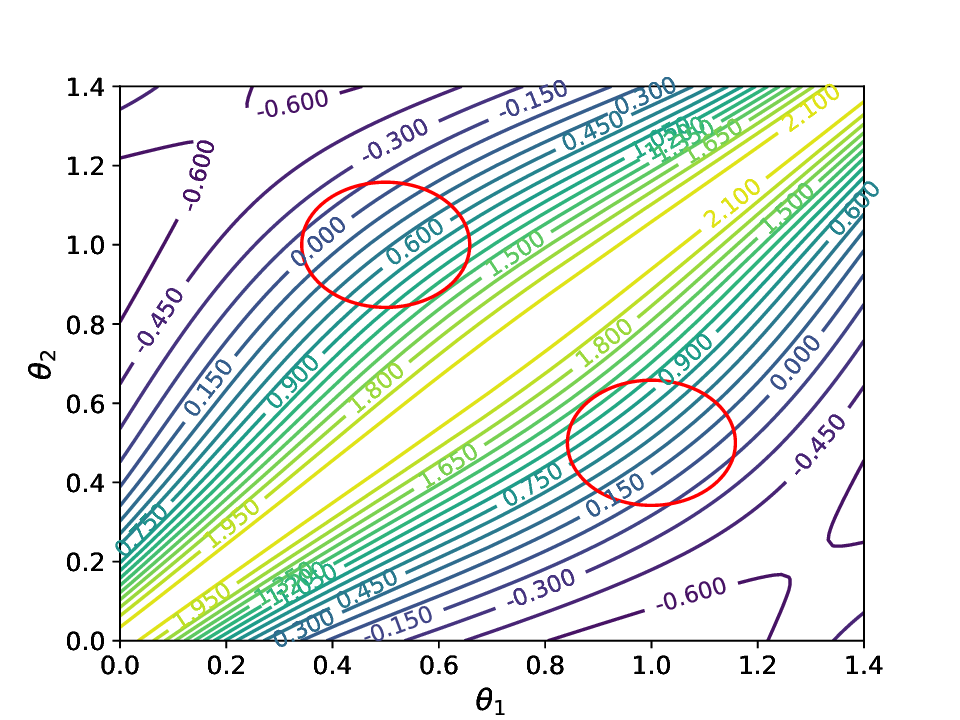}
		\caption[second caption.]{$\Sigma_{1,2}$}
	\end{subfigure}
	
	\caption{The three components of the elementary covariance matrix $\Sigma$ in~\eqref{eq:VarAnalytique} when the elementary likelihood are mixture of Gaussians: surface (left) and contour (right) plots. The modes of the posterior distribution are indicated by orange ellipses.}
	\label{fig:var}
\end{figure}

We numerically illustrate here that the bias on the posterior distribution is indeed of order~$\eps(n)\dt + \dt^2$ when~\eqref{eq:assConst} does not hold. We consider to this end the model described in Section~\ref{sec:gaussian_model} with the same parameters as in that section. Assuming that the data points~$x_i$ are distributed according to $ \PLe(\cdot|\tht_0)$ for some~$\tht_0$, the covariance $\Sigma_\textrm{x}(\tht)$ has the following limit when the number of data points $\Nd$ is large:
\begin{equation}
\begin{aligned}
\Sigma(\tht) &:= \lim_{\Nd \to \infty} \Sigma_{\vectx}(\tht)\\
&= \dps\int\limits_\mathcal{X} \gradT (\log \PLe (x|\tht)) \gradT (\log \PLe (x|\tht))^T \PLe(x|\tht_0) \, dx \\
& -\left(\dps\int\limits_\mathcal{X} \gradT (\log \PLe (x|\tht)) \PLe(x|\tht_0)dx  \right)\left(\dps\int\limits_\mathcal{X} \gradT (\log \PLe (x|\tht) )\PLe(x|\tht_0)dx  \right)^T.
\end{aligned}
\label{eq:VarAnalytique}
\end{equation}
This limit is well defined provided $\gradT( \log\PLe(\cdot|\tht)) \in L^2(\PLe(\cdot|\tht_0))$ for all $\tht \in \Theta$. Computing~$\Sigma(\tht)$ in~\eqref{eq:VarAnalytique} is intractable in general for two reasons: (i) $\tht_0$ is unknown and anyway the data points may even not be distributed according to $\PLe(\cdot|\tht_0)$; (ii) the integral over $x$ is possibly a high dimensional one or requires an evaluation cost of $\mathcal{O}(\Nd)$ to approximate it. We nonetheless use the formula~\eqref{eq:VarAnalytique} to compute the elements of~$\Sigma(\tht)$ in the low-dimensional case considered here to better understand the issues at stake. We plot in Figure~\ref{fig:var} the components $\Sigma_{11}(\tht)$, $\Sigma_{12}(\tht)$ and $\Sigma_{22}(\tht)$ of the symmetric matrix $\Sigma(\tht)$ in~\eqref{eq:VarAnalytique}. It is clear that the variance genuinely depends on the value of the parameter $\tht = (\tht_1, \tht_2)$, so that assumption~\eqref{eq:assConst} fails in this case.

\begin{figure}
	\centering
		\begin{minipage}{0.49\textwidth}
		\centering
		\includegraphics[width=\textwidth]{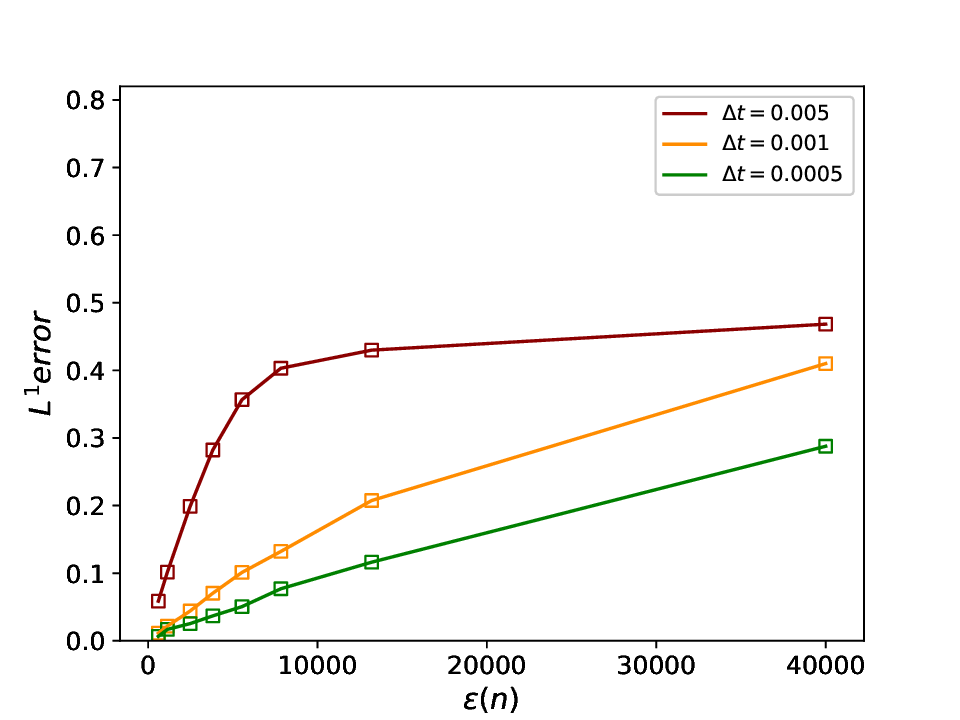}
		\subcaption{$\xi$ matrix}
	\end{minipage}
	\begin{minipage}{0.49\textwidth}
		\centering
		\includegraphics[width=\textwidth]{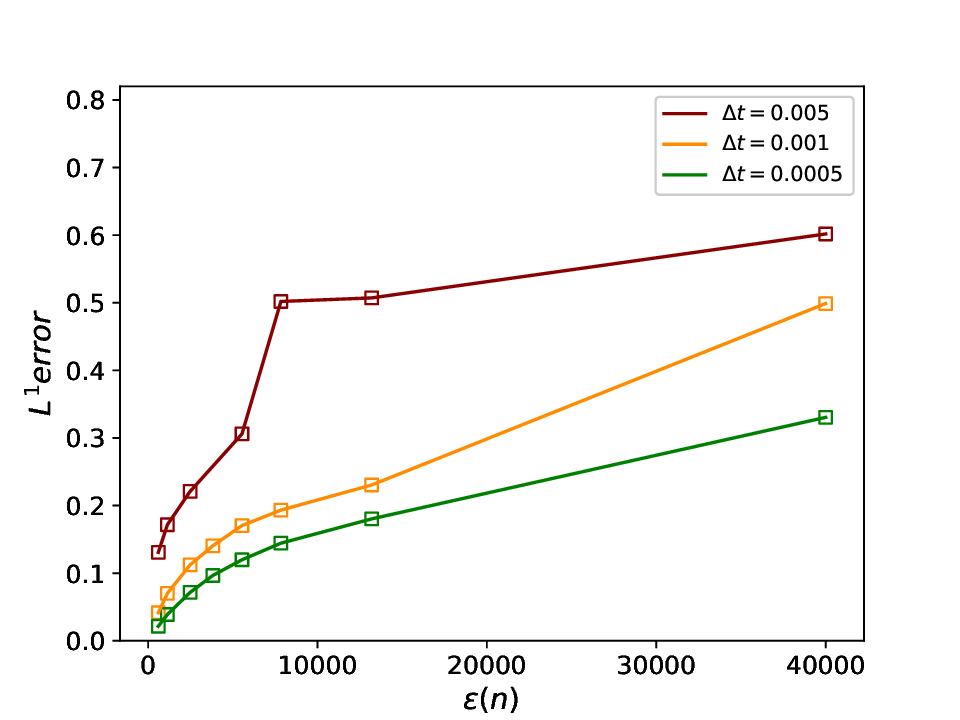}
		\subcaption{$\xi$ scalar}
	\end{minipage}
	\caption{$L^1$ error on the posterior distribution for various values of~$\Delta t$ and~$n$ when sampling from the posterior distribution for the mixture of Gaussians case using AdL, with mini-batching performed without replacement.} \label{fig:AdL_Mixture}
\end{figure}

To check whether the failure of assumption~\eqref{eq:assConst} has an impact on the properties of AdL, we use the dataset described in Section~\ref{sec:mixture_model}. We run the numerical scheme~\eqref{eq:adl_scheme} for the following two cases: when $\xi$ is a $d\times d$ matrix or when it is a scalar. We fix $T= 10^6$, $\gamma = 1$ and $\eta = 0.1$. This last choice is motivated by metastability issues we noticed while using AdL for larger values of~$\eta$. We report in Figure~\ref{fig:AdL_Mixture} the $L^1$ error given by~\eqref{eq:erreur_L1}. Note first that using AdL greatly reduces the $L^1$ error compared to the results of Figure~\ref{fig:SGLD}, obtained with SGLD~\eqref{SGLD} or the numerical scheme~\eqref{GLA_minibatched} (which corresponds to Langevin dynamics with mini-batching). This allows to run simulations with much larger timesteps~$\dt$. However, AdL, in both cases, fails to completely correct the bias due to mini-batching. Consistently with the analysis of Section~\ref{sec:bias_adl}, we observe that the bias seems to scale linearly with~$\eps(n)\dt\| \Sigma_{\vectx}(\tht) - S^*\|_{L^2(\pi)}$ (where~$S^*$ is the $L^2(\pi)$-projection of~$\Sigma_\vectx$ onto the set of admissible matrices) when this quantity is sufficiently small. We numerically compute
\[
\left\|\Sigma_{\vectx}(\tht) - \frac{1}{d}\int_\Theta \Tr \left( \Sigma_{\vectx}(\tht') \right) \pi(\tht'|\vectx)\, d \tht' \right\|_{L^2(\pi)} \approx 1.75
\] 
for the scalar case, and 
\[
\left\|\Sigma_{\vectx}(\tht) -\int_{\Theta} \Sigma_{\vectx}(\tht') \pi(\tht' |\vectx) \, d\tht' \right\|_{L^2(\pi)} \approx  0.57
\]
for the full matrix case. This is consistent with the numerical results for $\eps(n) \dt \leq 100$, where the error on the posterior distribution in the scalar case is roughly $3$ times larger than the error in the full matrix case.  

\subsection{Lack of reduction of the bias for a model of logistic regression}
\label{sec:log_reg}

The estimate of Lemma~\ref{lemme:f_adl} suggests that the reduction in the minibatching error is directly related to the improved quality of the approximation of the covariance matrix. We demonstrate in this section on the example of a Bayesian logistic regression model that the reduction may not be substantial in certain cases when passing from a scalar friction to a genuinely matricial one. This motivates using scalar AdL as this method has a smaller computational cost, and explains why previous numerical results obtained with scalar AdL were of good quality.

\begin{figure}
	\centering
	\begin{minipage}{0.48\textwidth}
		\centering
		\includegraphics[width=\textwidth]{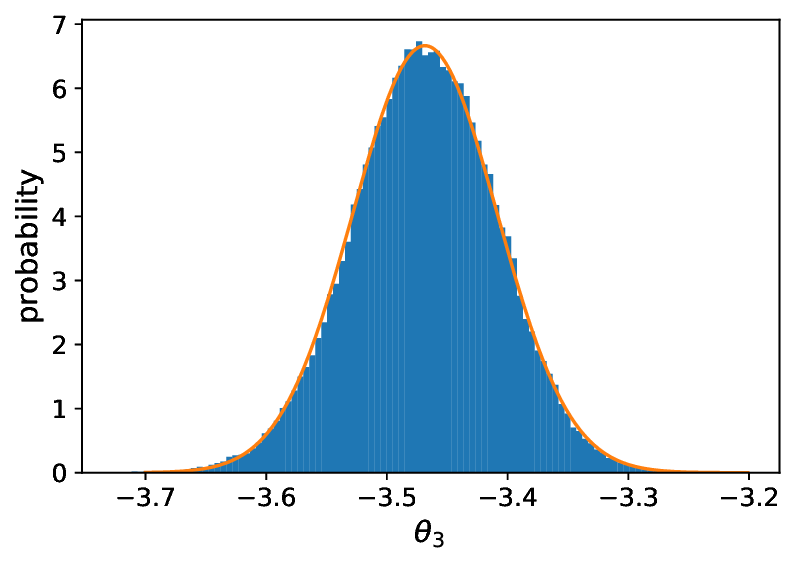}
		\subcaption[second caption.]{Marginal distribution of~$\tht_{3}$}
	\end{minipage}
	\begin{minipage}{0.48\textwidth}
		\centering
		\includegraphics[width=\textwidth]{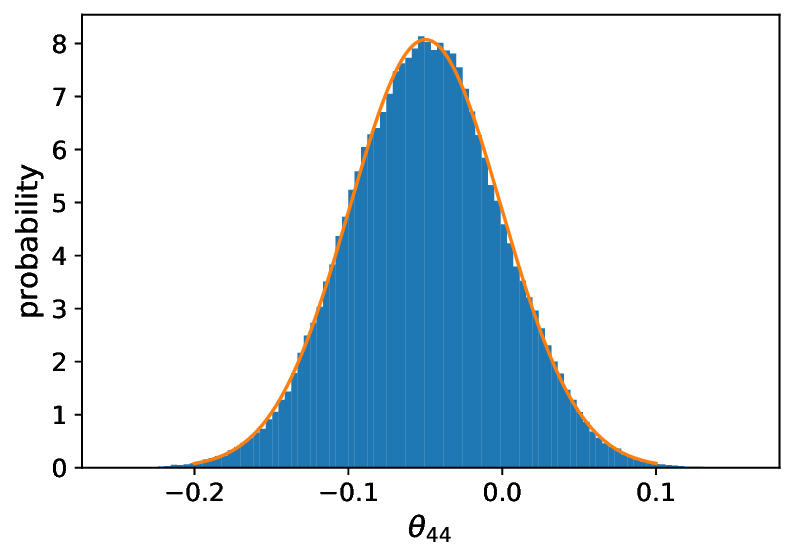}
		\subcaption[first caption.]{Marginal distribution of~$\tht_{44}$}
	\end{minipage}
	\begin{minipage}{0.48\textwidth}
		\centering
		\includegraphics[width=\textwidth]{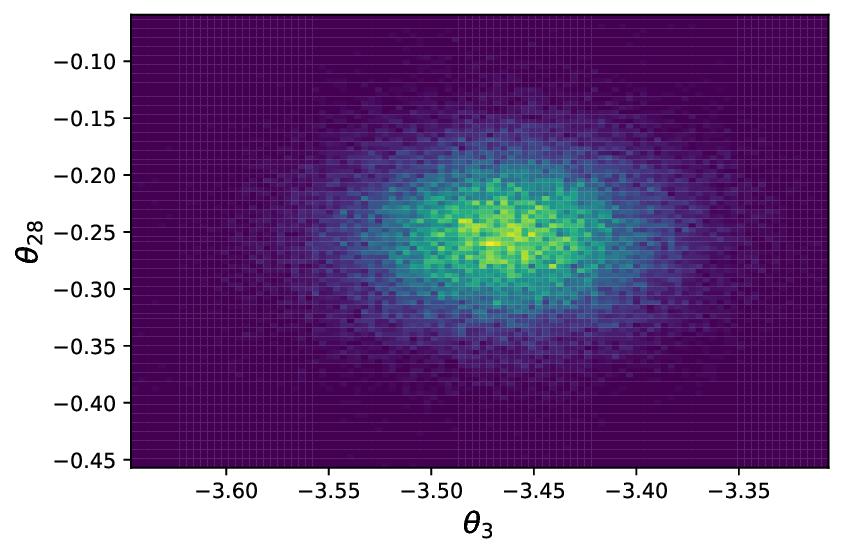}
		\subcaption[first caption.]{Joint marginal distribution of~($\tht_{3}$, $\tht_{29}$)}
	\end{minipage}
	\begin{minipage}{0.48\textwidth}
		\centering
		\includegraphics[width=\textwidth]{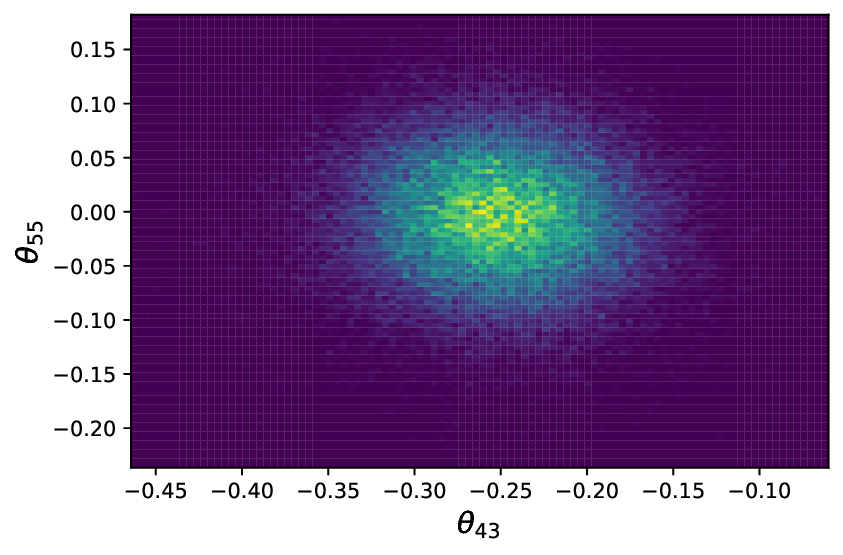}
		\subcaption[first caption.]{Joint marginal distribution of~($\tht_{43}$, $\tht_{55}$)}
	\end{minipage}
	\caption{Marginal posterior distributions of $\tht_3$ and $\tht_{44}$ (top), and of~($\tht_{3}$, $\tht_{28}$) and~($\tht_{43}$, $\tht_{55}$) (bottom).} \label{fig:traj_log_reg}
\end{figure}

\paragraph{Presentation of the model.}
We consider a subset of the MNIST data set containing the digits $7$ and $9$, and which have been pre-processed by a principal component analysis, as described in Section~4.3 of~\cite{MR4099815}. More precisely, there are~$\Nd = 12,251$ scores on the~100 first principal component $\vectz = (z_1, z_2, ..., z_{\Nd}) \subset \R^d$ (with~$d=100$), labeled by $\vecty = (y_1, y_2, ..., y_{\Nd}) \subset \{0,1\}$ (the labels~0 and~1 correspond respectively to digits~7 and~9). We assume that the elementary likelihood on the data point~$x_i=(z_i,y_i)$ is
\[
\PLe(x_i|\tht) = \sigma(\theta^T z_i)^{y_i} (1-  \sigma(\theta^T z_i))^{1-y_i},
\]
where
\begin{equation}
  \label{eq:sigmoid}
  \sigma(z) = \frac{\exp(z)}{1+\exp(z)},
\end{equation}
and $\tht\in \R^d$ is the vector of parameters we want to estimate. Assuming that the prior over the vector of parameter $\tht$ is a centered standard normal distribution, a simple computation based on the identity~$\sigma'(z) = \sigma(z)(1-\sigma(z))$ shows that 
\[
\widehat{F}_n(\tht) = \tht + \frac{\Nd}{n}  \sum_{i\in I_n} (y_i - \sigma(\tht^Tz_i))z_i.
\]
\paragraph{Numerical results.}
We use AdL to sample from the posterior distribution, with~$\gamma = \eta = 1$, $T = 10^3$ and $\dt = 10^{-3}$. We run the numerical scheme~\eqref{eq:adl_scheme} for the following three cases: $\xi$ is a scalar, a diagonal matrix or a full matrix. We plot the marginal distributions of some parameters~$\tht_i$ in Figure~\ref{fig:traj_log_reg}.  The results suggest that the posterior distribution of the parameters is close to a Gaussian distribution. As pointed out in Section~\ref{sec:numerics_AdL}, AdL is sufficient in this case to remove the minibatching error. 

\begin{figure}
	\centering
	\begin{minipage}{0.49\textwidth}
		\includegraphics[scale=0.6]{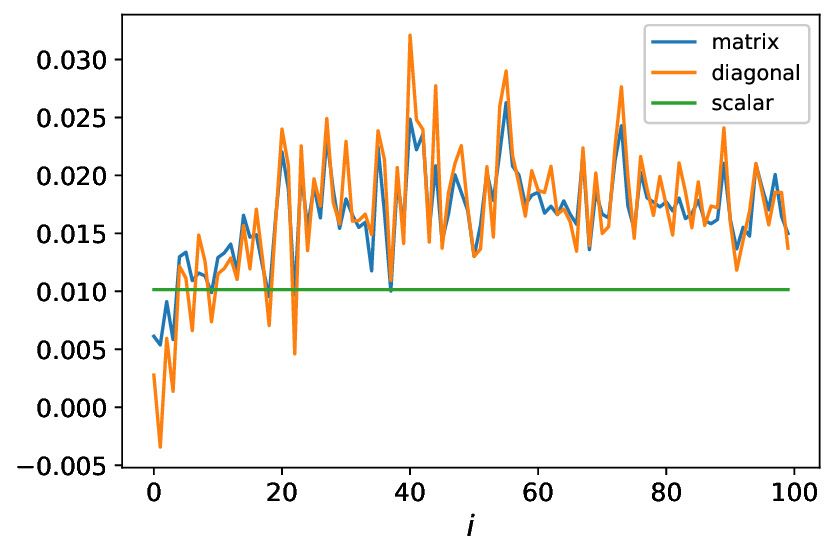}
	\end{minipage}
	\hfill
		\begin{minipage}{0.45\textwidth}
	\includegraphics[scale=0.6]{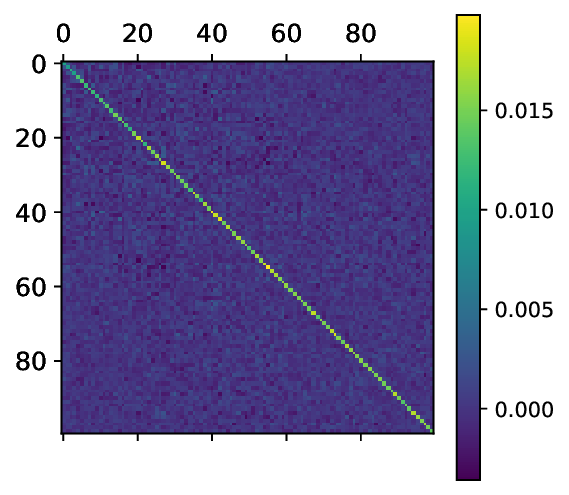}
\end{minipage}
	\caption{Right: Estimated covariance matrix of $\tht$.
		Left: Diagonal components of the estimate of the average covariance matrix of the stochastic gradient as a function of the index $1 \leq i \leq d$.} \label{fig:mat_cov}
\end{figure}

\begin{figure}
	\centering
	\includegraphics[scale=0.6]{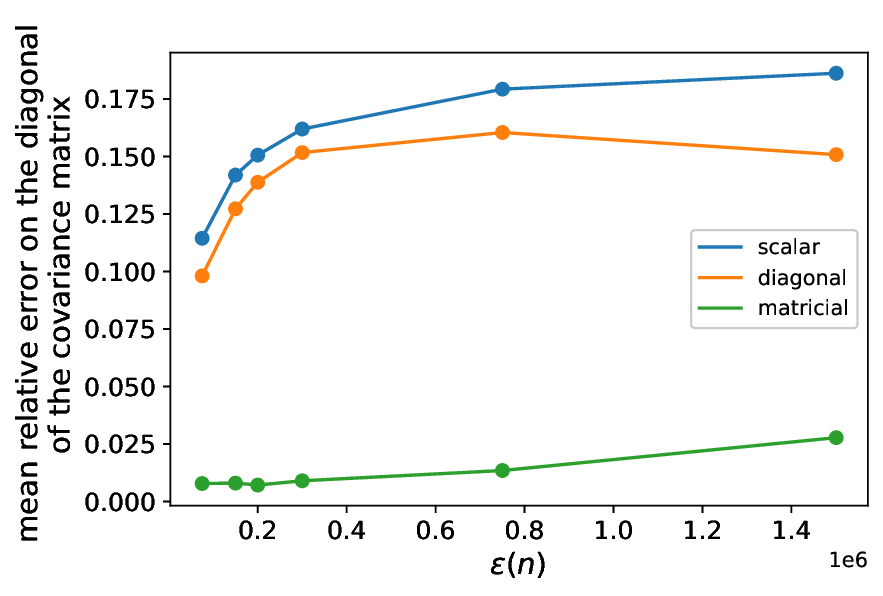}
	\caption{Error on the covariance matrix of the parameters using AdL for~$\xi$ scalar, diagonal matrix or full matrix.} \label{fig:reg_log_error}
\end{figure}

According to Lemma~\ref{lemmeADL}, the posterior distribtion of the variable~$\xi$ is centered on~$A_{\dt,n}$ given by~\eqref{eq:A}, when the latter matrix is constant. An estimate of the average covariance matrix can then be obtained with the estimator
\[
\frac{1}{\eps(n) \dt } \left(\frac{1}{\Ni} \sum_{m = 1}^{\Ni} \xi^m -  \gamma \mathrm{I}_d \right),
\]
where $\gamma \mathrm{I}_d$ is replaced by $\gamma$ in the scalar case. We plot in Figure~\ref{fig:mat_cov} the diagonal components of the estimate of the average of the covariance matrix of the stochastic gradient for the following three cases considered here ($\xi$ scalar, diagonal of full matrix). We can see that $\overline{\Sigma}_{\vectx}$ has a dominant part which is equal to~$s^* \mathrm{I}_d$ (although there are many non zero off diagonal terms). In view of Remark~\ref{rem:adl_xi_cases}, we expect AdL to give comparable results whether the friction variable is a scalar, a diagonal matrix and a full matrix. In order to investigate this, we plot in Figure~\ref{fig:reg_log_error} the mean relative error on the diagonal entries of the covariance matrix of the vector of parameters $\tht$ with respect to $\eps(n)$, given by
\[
\frac{1}{d} \sum_{i = 1}^{d}\left| \frac{[\cov(\tht)_n]_{i, i} - [\cov(\tht)_{\Nd}]_{i, i}}{[\cov(\tht)_{\Nd}]_{i, i}} \right|,
\]
where $\cov(\tht)_n$ and~$\cov(\tht)_{\Nd}$ denotes respectively the covariance of the vector of parameters~$\tht$ along trajectories of AdL with a minibatch of size~$n$, and AdL with a computation of the exact gradient using the~$\Nd$ data points. The error on the covariance matrix is some measure of the bias on the posterior distribution effectively sampled by the numerical scheme. The results confirm that AdL with scalar $\xi$ already leads to acceptable results in this example and performs as well as AdL with diagonal $\xi$.

\subsection{Bayesian neural networks}
\label{sec:BNN}

In this section, we give another numerical example which demonstrates that, when the covariance matrix is not isotropic, there is a reduction of bias when passing from scalar to diagonal AdL. We consider to this end Bayesian neural networks (BNNs) used for binary classification problems. BNNs sample parameters of neural networks and therefore provide a distribution of likelihoods for points outside of the training set. The fact that a distribution of likelihoods has a large variance is an indication that the classification of points outside the training set is uncertain. We consider here such a situation. We assess how the distributions of likelihoods is impacted by the minibatching error, and how this error can be reduced by diagonal versions of AdL.

\begin{figure}
  \centering
    \includegraphics[scale=0.4]{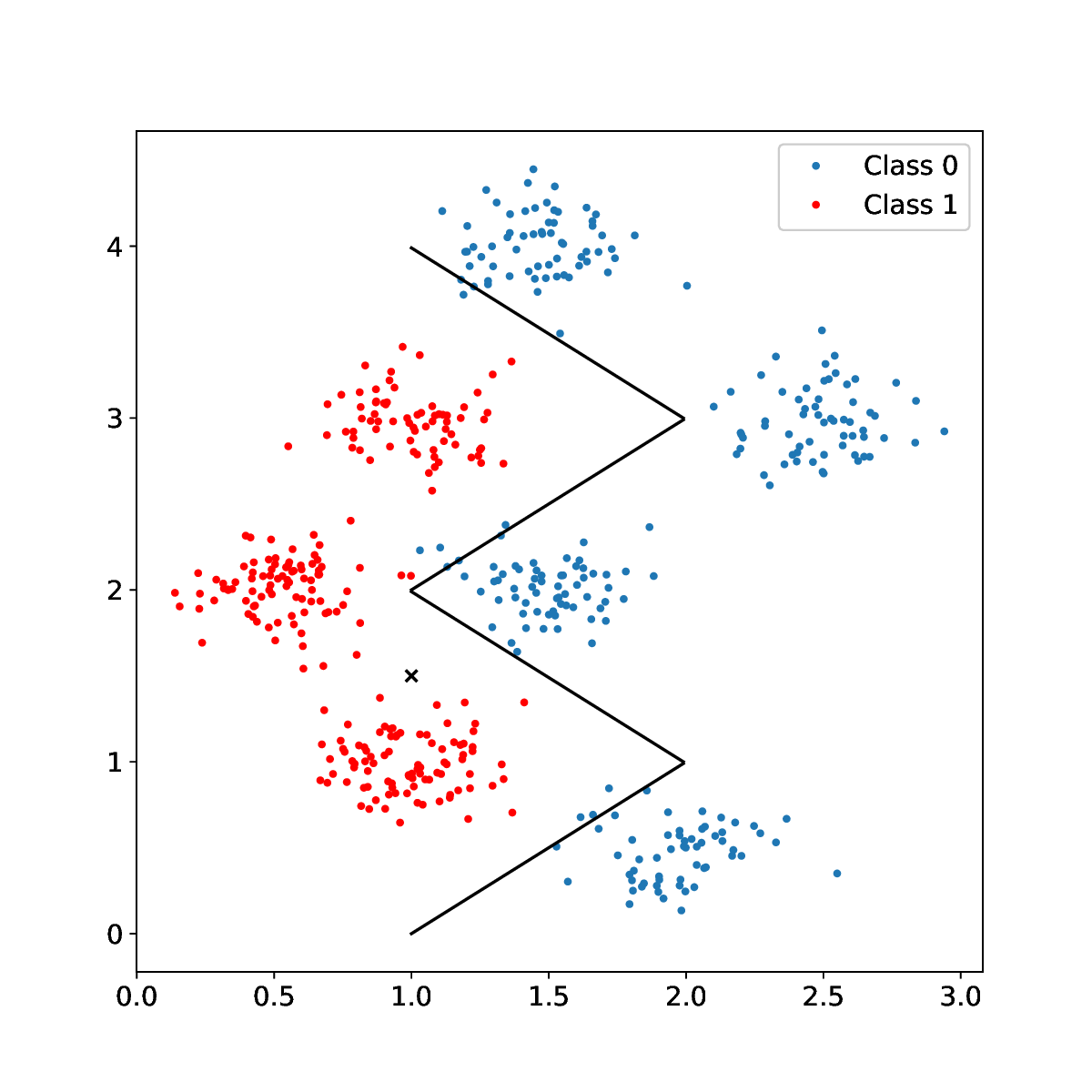}
  \caption{Data set for the classification problem tackled with BNNs. The solid lines are a guide to the eye to indicate possible decision boundaries. The cross represents the point~$x=(1,1.5)$ at which the parameters of the BNN, and hence its outputs, are sampled.} \label{fig:data_BNN}
\end{figure}

\paragraph{Presentation of the model.} We use a synthetic data set created by generating points using a mixture of Gaussians. Points in the class~$0$ (resp.~$1$) are generated by considering realizations of a random variable~$X_0 \in \mathbb{R}^2$ (resp.~$X_1$) 
\[
X_0 = x_{0, \mathcal{I}} + c G_0, \qquad X_1 =x_{1, \mathcal{J}} + c G_1,
\]
where~$G_0,G_1$ are two independent 2-dimensional standard Gaussian random variables, while~$\mathcal{I} \sim \mathcal{U}\{1,2,3\}$, $\mathcal{J}\sim \mathcal{U}\{1,2,3, 4\}$ and  
\[\left\{ \begin{aligned}
&x_{0, 1} = (1, 1),\\
&x_{0, 2} = (0, 2) ,\\
&x_{0, 3} =(1, 3).
\end{aligned} \right. \qquad
\left\{ \begin{aligned}
&x_{1, 1} =(2, 0.5),\\
&x_{1, 2} = (1, 2) ,\\
&x_{1, 3} =(2, 3),\\
&x_{1, 4} =(1, 4),
\end{aligned}
\right.\]
with $c^2 =0.03$. The first class is therefore the superposition of~3 Gaussian distributions, while the second one is the superposition of~4 Gaussian distributions. The total number of data points is~$\Nd = 500$ ($250$ points in each class); see Figure~\ref{fig:data_BNN}.
The probability that a point~$x$ has label~1 is estimated with a 2-layer neural network, with a hidden layer of size~$d_1$. More precisely, 
\[
N_\tht(x) = \sigma(W^2r(W^1x + b)),
\]
where $W^1 \in \R^{d_1 \times 2}$, $b \in \R^{d_1}$ and $W^2 \in \R^{d_1 \times 1}$. The activation functions~$r$ is the ReLU function~$r(z) = \max(0,z)$ which acts componentwise, while~$\sigma$ is the sigmoid function~\eqref{eq:sigmoid}. The vector of parameters is therefore~$\tht = (W^1, b, W^2) \in \R^{d}$ with $d=4 d_1$, where the matrices $W^1$ and $W^2$ are reshaped into vectors. For the numerical simulations, we set $d_1 = 8$.

The loss function is the binary cross entropy loss
\begin{equation}
L_f(x, y) = y \log(f(x)) + (1-y)\log(1-f(x)),
\label{eq:loss_BCE}
\end{equation} 
where~$x \in \mathbb{R}^2$ is a point from the data set and $y \in \{0,1\}$ its corresponding label. The likelihood distribution can be written in terms of the loss function as
\begin{equation*}
\PLe(x, y|\tht) \propto \exp(-L_{N_\theta}(x, y)).
\end{equation*}
The prior distribution on the parameters~$\theta$ is a standard $d$-dimensional Gaussian distribution.

\paragraph{Shape of the friction matrix in Adaptive Langevin dynamics.}
We restrict ourselves to diagonal friction matrices, and do not consider the full matricial case since it is computationally too expensive and cannot be considered as a practical sampling algorithm in the BNN framework. It is convenient to reshape~$\xi$ as a vector of dimension~$4d_1$.  As in~\eqref{eq:xi_diag}, the first~$2d_1$ entries of~$\xi$ correspond to the scalar coefficients associated with the sampling of the entries of~$W^1$, the following~$d_1$ components to the components of~$b$, and the final~$d_1$ components to those of~$W^2$. This corresponds more formally to the choice
\[
\xi = \left( \begin{array}{c|c|c}
\xi_1 & 0 &0 \\
\hline
0 & \xi_2 &0 \\
\hline 
0 & 0 & \xi_3 
\end{array} \right),
\]
where $\xi_1, \xi_2, \xi_3$ are diagonal matrices of respective sizes~$2d_1,d_1$ and~$d_1$. The number of coefficients can be reduced by considering a single scalar variable for~$W^1,b,W^2$, which corresponds to the choice 
\[
\xi = \left( \begin{array}{c|c|c}
\xi_1{\rm I}_{2  d_1} & 0 &0 \\
\hline
0 & \xi_2 {\rm I}_{d_1} &0 \\
\hline 
0 & 0 & \xi_3 {\rm I}_{d_1} \\ 
\end{array} \right),
\]
with $\xi_1, \xi_2, \xi_3 \in \R$. 

\begin{figure}
  \centering
  \begin{minipage}{0.49\textwidth}
    \includegraphics[scale=0.6]{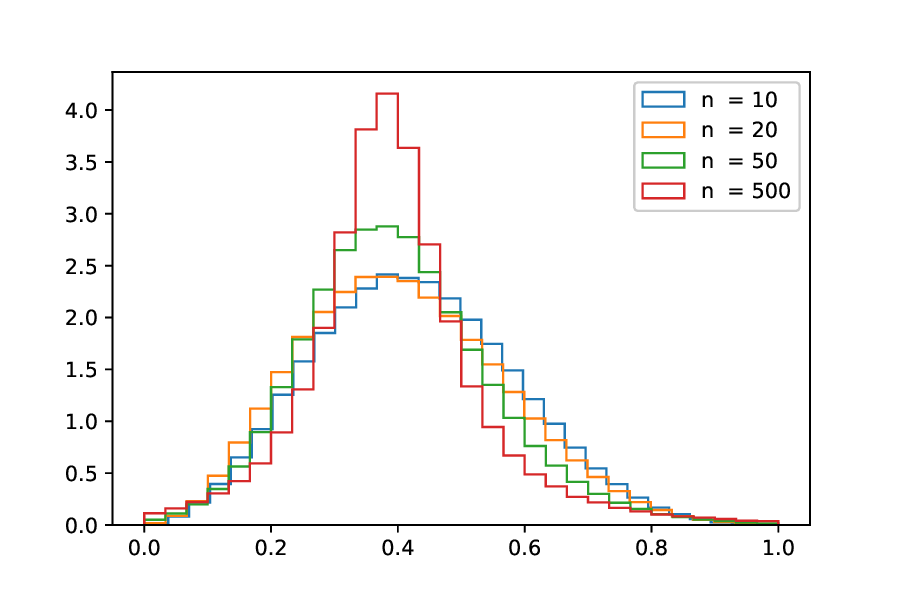}
    \subcaption{Scalar AdL}
  \end{minipage}
  \hfill
  \begin{minipage}{0.45\textwidth}
    \includegraphics[scale=0.6]{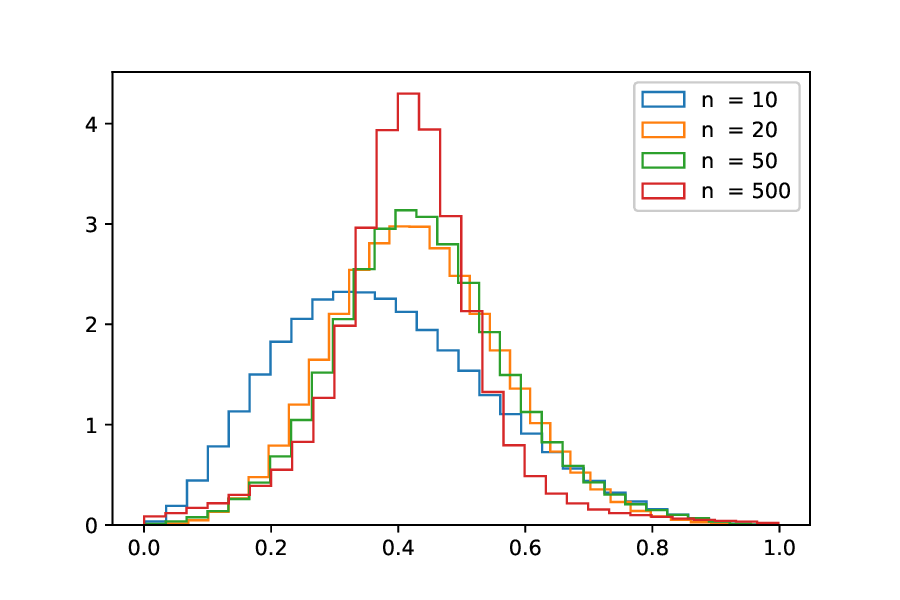}
    \subcaption{diagonal AdL}
  \end{minipage}
  \caption{Distribution of $N_{\tht}(x)$ for various size of minibatch for $\dt = 10^{-2}$.} \label{fig:N_tht}
\end{figure}

\begin{figure}
  \centering
  \begin{minipage}{0.49\textwidth}
    \includegraphics[scale=0.6]{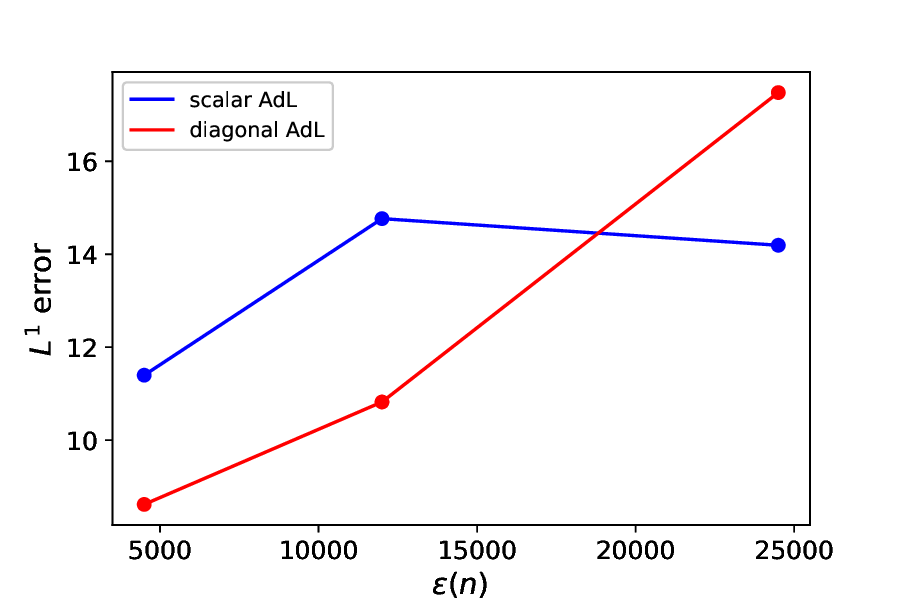}
    \subcaption{$\dt = 10^{-2}$}
  \end{minipage}
  \hfill
  \begin{minipage}{0.45\textwidth}
    \includegraphics[scale=0.6]{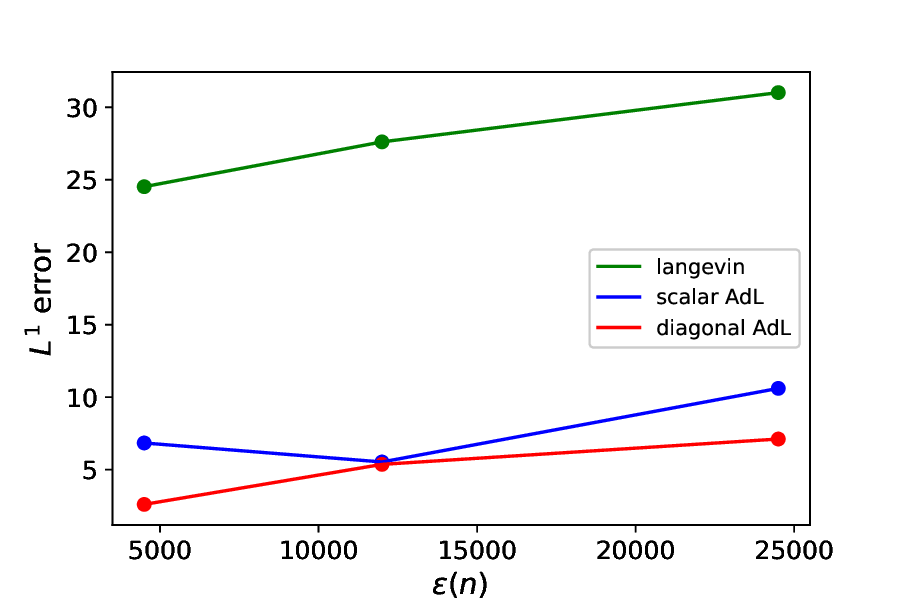}
    \subcaption{$\dt = 10^{-3}$}
  \end{minipage}
  \caption{$L^1$ error on the posterior distribution of $N_\tht(x)$ for various minibatching sizes and two values of the timestep~$\dt$.} \label{fig:mat_cov_BNN}
\end{figure}

\paragraph{Numerical results.}
We fix $\gamma = 1$ and $\eta=200$, and use the numerical scheme~\eqref{eq:adl_scheme} for AdL both in the diagonal and scalar cases to sample from the posterior distribution on~$\theta$. The numerical scheme is run for $4 \times 10^5$ epochs (each epoch being composed of~$\Nd/n$ iterations of the numerical method), for various sizes of minibatch and various time steps. We plot in Figure~\ref{fig:N_tht} the distribution of $N_{\tht}(x)$ for various size of minibatch in the case of diagonal and scalar AdL; and in Figure~\ref{fig:mat_cov_BNN} the $L^1$ error on the distribution of~$N_\tht(x)$ for a given point~$x = (1, 1.5)$ outside of the dataset. The exact posterior distribution is considered to be the one corresponding to AdL without minibatching. 

The results show that the diagonal version of AdL reduces the minibatching error compared to the scalar version of AdL for values of~$\varepsilon(n)\Delta t$ sufficiently small. To confirm the effect of the non isotropic matrix on the reduction of the bias, we approximate the mean and the variance of the diagonal element of the covariance matrix~$\Sigma_\vectx$ using $\xi$ for the diagonal AdL and compare them to the mean of $\xi$ for the scalar case. We confirm that $$\frac{1}{\Ni}\sum_{i = 1}^{\Ni} \frac{1}{d} \sum_{j = 1}^{d} [\xi_{\rm {diagonal}, i}]_{j, j}\approx \frac{1}{\Ni} \sum_{i = 1}^{\Ni} \xi_{{\rm scalar}, i},$$ where $\xi_{\rm diagonal}$ and $\xi_{\rm scalar}$ are respectively the $\xi$ variables for the diagonal and scalar AdL. The variance on the diagonal elements is about~$2$ (for each set of NN parameters), while the average of these elements is of the order of~$\gamma=1$.

\section{Extended Adaptive Langevin Dynamics}
\label{sec:eAdL}

AdL corrects for the bias due to mini-batching when the covariance of the stochastic gradient is constant. This is unfortunately not always the case as we demonstrated in Section~\ref{subsec:Covariance}. In this section, we suggest an extension of AdL which allows to remove the bias when the covariance is not constant but can be decomposed on a finite basis of functions. This approach in fact generally allows to reduce the minibatching bias even if the covariance matrix cannot be decomposed in a finite basis, as the residual minibatching bias, still of order~$\varepsilon(n)\dt$, has a prefactor related to some projection error of the covariance matrix onto the approximation space under consideration, see~\eqref{eq:proj_sigma} below.

We start by presenting the modified AdL dynamics in Section~\ref{sec:presentation_eADL}, under the assumption that the covariance of the force estimator can be decomposed on a finite basis of functions. This key assumption guarantees that the marginal distribution in the~$\tht$ variable is the target posterior~\eqref{pi_bayes}. We next propose a numerical scheme for extended AdL in Section~\ref{sec:disEADL}, where we also quantify the bias on the invariant measure arising from the use of finite time steps, and possibly from mini-batching in situations where the covariance of the force estimator cannot be decomposed as a finite sum for the chosen basis of functions. We then discuss in Section~\ref{sec:choice} a crucial point of the method, namely the choice of basis functions. We finally present in Section~\ref{Mixture} numerical results for the example of Section~\ref{sec:mixture_model} where the elementary likelihoods are given by a mixture of Gaussians, and for which AdL fails to fully remove the bias because the covariance matrix is not constant. We demonstrate that with a reduced number function basis, the bias is significantly reduced even for the smallest values of~$n$, including $n=1$.

\subsection{Presentation of the dynamics}
\label{sec:presentation_eADL}

Using the same notation as in Section~\ref{sec:AdL}, let us consider the case where $ A_{\dt,n}(\tht)$ in~\eqref{eq:A} genuinely depends on $\tht$ in the following manner.

\begin{assumption}
	The matrix-valued function $A_{\dt,n}$ can be decomposed on a finite basis of functions $f_0,\dots,f_K$ as
	\begin{equation}
	\label{eq:def_Aq}
	A_{\dt,n}(\tht) = \sum_{k=0}^K A_{\dt,n}^k f_k(\tht),
	\end{equation}
	where $A_{\dt,n}^0, ..., A_{\dt,n}^K \in \R^{d \times d}$ are symmetric matrices. Moreover,  $A_{\dt,n}(\tht)$ is a symmetric positive matrix for any $\tht \in \Theta$.
	\label{assumption:A}
\end{assumption}

The choice of the basis of function is of great importance for the numerical performance of the method and is discussed more precisely in Section~\ref{sec:choice}. The situation when~\eqref{eq:def_Aq} does not hold is considered at the end of Section~\ref{sec:disEADL}. Note that the matrices $A_{\dt,n}^0, ..., A_{\dt,n}^K$ need not be positive as long as $A_{\dt,n}(\tht)$ is. In accordance with~\eqref{eq:def_Aq}, we choose the variable $\xi_t(\tht)$ to be of the following form:
\begin{equation}
\label{eq:xi_eAdL}
\xi_t(\tht) = \sum_{k=0}^K \xi_{k,t} f_k(\tht), \qquad \xi_{k,t} \in \R^{d\times d}.
\end{equation}
If $K = 0$ and $f_0 = \mathbf{1}$, then~\eqref{eq:def_Aq} coincides with assumption~\eqref{eq:assConst}, in which case AdL is sufficient to remove the bias due to mini-batching. In practice, the expression~\eqref{eq:def_Aq} can be thought of as a truncation of the expansion of the function $A_{\dt,n}(\tht)$ on a complete basis.

We are now in position to write the following extended Adaptive Langevin dynamics (eAdL), for which we introduce $K+1$ additional (matrix) equations on the coefficients~$\xi_{k,t}$ in~\eqref{eq:xi_eAdL}:
\begin{equation}
\left\{ \begin{aligned}
d\tht_t & = p_t \, dt, \\
dp_t & = \gradT (\log\pi(\tht_t|\vectx)) \, dt - \xi_t(\tht_t) p_t \, dt + \sqrt{2} A_{\dt,n}(\tht_t)^{1/2}\, dW_t, \\
d[\xi_{k,t}]_{i,j} & = \frac{f_k(\tht_t)}{\eta_k} \left( p_{i,t} p_{j,t} -  \delta_{i,j}\right), \qquad 1 \leq i , j \leq d,  \quad 0 \leq k \leq K,
\end{aligned} \right.
\label{eq:EadL}
\end{equation}
where $A_{\dt,n}$ is given by~\eqref{eq:A}, $[\xi_{k,t}]_{i,j}$ is the $(i,j)$ component of $\xi_{k,t} \in \R^{d \times d}$, and $\eta_k$ are positive scalars for $0 \leq k \leq K$. The interest of eAdL is the following consistency result on the existence of an invariant probability measure with the correct marginal distribution in the~$\tht$ variable.

\begin{theorem}
  \label{th:1}
  Suppose that Assumption~\ref{assumption:A} holds. Then, the eAdL dynamics~\eqref{eq:EadL} admits the invariant probability measure
  \begin{equation}
    \nu_K(d\tht\,dp\,d\xi_0\dots d\xi_K) = \pi(\tht|\emph{\vectx}) \tau(dp) \rho_K(d\xi) \, d\tht,
    \label{eq:nu_K}
  \end{equation}
  where $\tau(dp)$ is defined in~\eqref{eq:tau}, and 
  \[
  \rho_K(d\xi_0\dots d\xi_K) = \prod_{k=0}^K \prod_{i , j = 1}^{d} \sqrt{\frac{\eta_k}{2 \pi}} \exp\left(-\frac{\eta_k}{2}\left([\xi_{k}]_{i,j}- [A_{\dt,n}^k]_{i,j}\right)^2\right)  d[\xi_{k}]_{i,j}, 
  \]
  with $[A_{\dt,n}^k]_{i,j}$ the $(i,j)$ component of~$A_{\dt,n}^k \in \R^{d\times d}$.
\end{theorem}

As for AdL (see the discussion following Lemma~\ref{lemmeADL}), Theorem~\ref{th:1} suggests that we recover the target posterior distribution~$\pi(\cdot|\vectx)$ whatever the extra noise due to mini-batching, since the marginal in the variable~$\tht$ of the probability measure~$\nu_K$ in~\eqref{eq:nu_K} is $\pi(\cdot|\vectx)$. However, from a discussion similar to the one in Remark~\ref{rem:nu_ergodic}, the dynamics cannot be ergodic for the extended measure~$\nu_K$.


\begin{proof}
  We follow the same approach as for the proof of Lemma~\ref{lemmeADL}. The generator of the dynamics reads
  \[
  \mathcal{L}_{\rm eAdL, \Sigma_\vectx}  =  \cLham + \mathcal{L} _{\mathrm{FD}} + \widetilde{\mathcal{L}} _{\mathrm{NH}},
  \]
  where $\cLham$ and $\mathcal{L} _{\mathrm{FD}}$ are the same operators as in~\eqref{eq:lham} and~\eqref{eq:cLfd} respectively (even if $A_{\dt,n}$ depends on $\tht$), while 
  \[
  \widetilde{\mathcal{L}} _{\mathrm{NH}} = \sum\limits_{k=0}^K f_k\mathcal{L}_{\mathrm{NH}, k},
  \]
  where 
  \[
  \mathcal{L}_{\mathrm{NH}, k} = - \sum_{i , j = 1}^{d} p_i \left([\xi_k]_{i,j} - [A_{\dt,n}^k]_{i,j}\right) \partial^*_{p_j} + \frac{1}{\eta_k}(p_ip_j - \delta_{i,j})\partial_{[\xi_k]_{i,j}}.
  \]
  A computation similar to the one performed in the proof of Lemma~\ref{lemmeADL} shows that $\mathcal{L}_{\mathrm{NH}, k}^* = -\mathcal{L}_{\mathrm{NH}, k}$, with adjoints taken with respect to~$L^2(\nu_K)$; while $\cLham$ and $\mathcal{L} _{\mathrm{FD}}$ are respectively antisymmetric and symmetric on~$L^2(\nu_K)$. This implies that $\mathcal{L}_{\rm eAdL, \Sigma_\vectx}^* \mathbf{1} = 0$, from which the claimed invariance of~$\nu_K$ follows.
\end{proof}

\subsection{Numerical scheme and estimates on the bias}
\label{sec:disEADL}

We present in this section a numerical integrator for the eAdL dynamics~\eqref{eq:EadL} based on a Strang splitting similar to the one considered in Section~\ref{sec:numerical_scheme_AdL} for AdL. We consider the same elementary SDEs~\eqref{eq:elementary_SDE_tht}-\eqref{eq:elementary_SDE_xi} as for the discretization of AdL, except that there are now~$K+1$ elementary SDEs in~\eqref{eq:elementary_SDE_xi}, indexed by~$0 \leq k \leq K$. The associated numerical scheme obtained for $\Gamma = \gamma \mathrm{I}_d$ reads
\begin{equation}
\left\{ \begin{aligned}
\xi^m & = \sum_{k=0}^{K} \xi^{m}_k  f_k(\tht^{m}),\\
p^{m+\frac{1}{2}} & = \rme^{-\Delta t \xi^m/2} p^m +  \left[\gamma\left(\xi^m\right)^{-1} \left(\mathrm{I}_d-\rme^{- \Delta t \xi^m}\right) \right]^{1/2} G^m, \\
\xi_k^{{m+\frac{1}{2}}}  & = \xi^{m}_k+\frac{ \Delta t }{2 \eta_k }f_k(\tht^m) \left[ \left(p^{m+\frac{1}{2}}\right) \left( p^{m+\frac{1}{2}}\right)^T - \mathrm{I}_d\right], \qquad k = 0, ..., K,\\
\tht^{m+\frac{1}{2}} & = \tht^m + \frac{ \Delta t }{2} p^{m+\frac{1}{2}}, \\
\widetilde{p}^{m+\frac{1}{2}} & = p^{m+\frac{1}{2}} + \Delta t \widehat{F_n} \left( \tht^{m+\frac{1}{2}} \right),\\
\tht^{m+1} & =  \tht^{m+\frac{1}{2}}  +   \frac{ \Delta t }{2} \widetilde{p}^{m+\frac{1}{2}}, \\
\xi^{m+1}_k  & =  \xi_k^{m+\frac{1}{2}}  + \frac{ \Delta t }{2 \eta_k } f_k(\tht^{m+1})\left[ \left(  \widetilde{p}^{m+\frac{1}{2}} \right) \left( \widetilde{p}^{m+\frac{1}{2}} \right)^T- \mathrm{I}_d\right], \qquad k = 0, ..., K,\\
\xi^{m+1} & = \sum_{k=0}^{K} \xi^{m+1}_k  f_k(\tht^{m+1}), \\
p^{m+1} & = \rme^{-\Delta t \xi^{m+1}/2} \widetilde{p}^{m+\frac{1}{2}}+\left[\gamma\left(\xi^{m+1}\right)^{-1} \left(\mathrm{I}_d-\rme^{- \Delta t \xi^{m+1}}\right) \right]^{1/2}  G^{m+ \frac{1}{2}},
\end{aligned} \right.
\label{eq:sheme_eadl}
\end{equation} 
where $(G^m)_{m \geq 0}$ and $(G^{m+ \frac{1}{2}})_{m \geq 0}$ are two independent families of i.i.d.~standard $d$-dimensional Gaussian random variables.

\paragraph{Characterization of the bias.}
We assume as before that the numerical scheme~\eqref{eq:sheme_eadl} admits a unique invariant probability measure, which may depend on the initial condition~$(\xi_0^0,\dots,\xi_K^0)$. Error estimates on this invariant probability measure can be obtained by following the approach of in Section~\ref{sec:bias_adl}. Given the high level of similarity between the estimates in Section~\ref{sec:bias_adl} and the ones obtained here, we only make precise what needs to be changed.

The main conclusion is that the bias is still of order~$\dt^2 + \eps(n)\dt$, but with a smaller prefactor for the dominant term~$\eps(n)\dt$. Let us emphasize that we do not require Assumption~\ref{assumption:A} to hold for this analysis. To make this precise, the first step is to write the generator as
\[
\mathcal{L}_{\rm eAdL, \Sigma_\vectx} = \mathcal{L}_{\rm eAdL, \overline{\Sigma}_\vectx^K} + \eps(n)\dt \widetilde{\cL}_{\Sigma_{\vectx} - \overline{\Sigma}_\vectx^K},
\]
where~$\widetilde{\cL}_{\mathscr{M}}$ is defined in~\eqref{eq:bias_adl}, and~$\overline{\Sigma}_\vectx^K$ is a symmetric positive matrix which belongs to the vector space generated by~$f_0,\dots,f_K$. The specific choice of this matrix for the theoretical analysis relies again on an orthogonal projection and the associated projection error, see~\eqref{eq:proj_sigma} below. One can then state an inequality similar to~\eqref{erreuraAdl}, where~$\overline{\nu}_{\xi^0}$ is replaced by the invariant probability measure~$\overline{\nu}_{K,(\xi_0^0,\dots,\xi_K^0)}$ of the eADL dynamics associated with the generator~$\mathcal{L}_{\mathrm{eAdL}, \overline{\Sigma}_{\vectx}^{K}}$, and~$\nu_{\xi^0}$ is replaced by the invariant probability measure~$\nu_{K,(\xi_0^0,\dots,\xi_K^0)}$ of the eAdL dynamics with generator~$\mathcal{L}_{\rm eAdL, \Sigma_\vectx}$. The prefactor of the dominant error term, proportional to~$\eps(n)\dt$, depends on the function
\[
f_{\mathrm{eAdL},\Sigma_{\vectx},\overline{\Sigma}_\vectx^K} =  \left(-\cL_{\mathrm{eAdL},\overline{\Sigma}_\vectx^K}^*\right)^{-1} \widetilde{\cL}_{\Sigma_{\vectx} - \overline{\Sigma}_\vectx^K}^* \mathbf{1},
\]
where adjoints are taken on~$L^2(\overline{\nu}_{K,(\xi_0^0,\dots,\xi_K^0)})$. Under appropriate conditions on the structure of~$\overline{\nu}_{K,(\xi_0^0,\dots,\xi_K^0)}$ and resolvent bounds for~$\mathcal{L}_{\mathrm{eAdL}, \overline{\Sigma}_\vectx^K}$, it is possible to upper bound the norm of~$f_{\mathrm{eAdL},\Sigma_{\vectx}}$ in~$L^2(\overline{\nu}_{K,(\xi_0^0,\dots,\xi_K^0)})$ by~$\left\| \Sigma_{\vectx}  - \overline{\Sigma}_\vectx^K\right\|_{L^2(\pi)}$. By optimizing upon the matrix valued function~$\overline{\Sigma}_\vectx^K$, one can conclude that the prefactor of the error term proportional to~$\eps(n)\dt$ is bounded, up to a constant, by
\begin{equation}
  \min_{\mathscr{M}_0,\dots,\mathscr{M}_K \in \mathbb{R}^{d \times d}} \left\| \Sigma_{\vectx} - \sum_{k=0}^K \mathscr{M}_k f_k \right\|_{L^2(\pi)},
  \label{eq:proj_sigma}
\end{equation}
which corresponds to the $L^2(\pi)$ projection error of~$\Sigma_{\vectx}$ onto the vector space of symmetric matrices generated by the basis. This shows that the better the approximation of the covariance~$\Sigma_{\vectx}$ is for the chosen basis~$(f_0,\dots,f_K)$, the smaller the bias is.

\paragraph{Discussion on the computation cost.}
As for the numerical discretization of AdL~\eqref{eq:adl_scheme} (see Remark~\ref{rem:other_version_num}), we can also use the numerical scheme~\eqref{eq:sheme_eadl} while considering each variable $\xi_k$ as a scalar or a diagonal matrix. Observations similar to the ones made in Remark~\ref{rem:adl_xi_cases} apply here as well. More precisely, the prefactor of the error term proportional to~$\eps(n) \dt$ is bounded by the $L^2(\pi)$-projection of~$\Sigma_{\vectx}$ onto: (i) the vector space generated by functions with values in the vector space of isotropic matrices, \emph{i.e.} the variables~$\xi_k$ are scalars for each~$k$, so that $\mathscr{M}_k = \sigma_k \mathrm{I}_d$ where $ \sigma_k \in \mathbb{R}$ in~\eqref{eq:proj_sigma}; (ii) the vector space of diagonal matrices generated by the basis when the variables $\xi_k$ are diagonal matrices for each $k$, meaning that $\mathscr{M}_0,\dots,\mathscr{M}_K$ are diagonal matrices in~\eqref{eq:proj_sigma}. For the numerical simulations reported in Section~\ref{Mixture}, we consider the variables~$\xi_k$ as full~$d\times d$ symmetric matrices.

The extra computational cost associated with extended Adaptive Langevin dynamics depends on the choice made for the form of the friction matrix. When~$\xi$ is a scalar, the extra computational cost compared to standard Langevin like dynamics is of order~$K$, while it is of order~$Kd$ when~$\xi$ is a diagonal matrix. When~$\xi$ is a matrix, the computational cost is much higher because of the matrix exponential which appears in the first and last steps of the numerical scheme~\eqref{eq:sheme_eadl}. The computation of such exponentials involves in general a matrix diagonalization, which has a cost of order~$d^3$. Additionally, one needs to update the components of~$K$ matrices of size~$d \times d$, which has a cost of order~$Kd^2$. Overall, the extra computational cost of extended Adaptive Langevin is therefore of order~$(K+d)d^2$ in the matrix case, to be compared to the cost of order~$d^3$ of usual Adaptive Langevin with constant friction matrices. This cost could however be mitigated to~$C K d^2$ upon using a midpoint scheme to integrate the Ornstein--Uhlenbeck process, and solving the resulting matrix problem using iterative methods. In any case, when the dimension~$d$ of the parameters is large, it is usually too expensive from a computational viewpoint to use matrix versions of Adaptive Langevin. This may also be unnecessary, as all the~$d^2$ entries of the covariance matrix are not necessarily relevant; recall the exampled of Section~\ref{sec:log_reg} and see the discussion in Section~\ref{sec:discussion_perspectives}.

\subsection{Choice the basis functions}
\label{sec:choice}

The choice of the basis functions~$f_0,\dots,f_K$ is a key point in eAdL, since there is a trade-off between a good approximation of the matrix $A_{\dt,n}(\tht)$ such that the projection error~\eqref{eq:proj_sigma} is small, and the number of additional degrees of freedom (which is equal to the number of functions $K$ introduced multiplied by $d(d+1)/2$ if the unknown are matrices). One typically prefers to introduce only a limited number of degrees of freedom. One option towards this goal is for instance to consider only isotropic matrices $\xi_k = \sigma_k \mathrm{I}_d$ with $\sigma_k \in \R$. Such a choice is advocated in~\cite{NIPS2014_5592} for~$K=0$ and~$f_0 = \mathbf{1}$.

\paragraph{Spatial decomposition.} Numerical experiments suggest that the covariance matrix of the estimator of the gradient may change rapidly in certain regions of the parameter space (see in particular Figure~\ref{fig:var}). A convenient approach to approximating this matrix is to partition the domain~$\Theta$ into $K+1$~subdomains, denoted by~$\mathcal{D}_0,\dots,\mathcal{D}_K$, and to consider the functions~$f_k$ to be indicator functions of these domains, \emph{i.e.} $f_k = \mathbf{1}_{\mathcal{D}_k}$. This corresponds to a piecewise constant approximation of the covariance matrix. If the domain is simple and the dimension~$d$ is sufficiently small, one can think of simple geometric decompositions, using \emph{e.g.} rectangles if~$\Theta$ is itself rectangular, or rings around the various modes of the posterior distribution. For more complicated domains in possibly high dimension, one can decompose the parameter space with a Voronoi tessellation, where the centers of the Voronoi tesselation are for instance the local maxima of the posterior distribution. Such points can be located by performing preliminary SGLD or AdL runs. This amounts to considering a constant friction matrix in each mode of the probability distribution. 

\paragraph{Polynomial approximation.} If the domain $\Theta$ is sufficiently simple from a geometric viewpoint, a more sophisticated method is to couple a spatial decomposition with some spectral approximation, by introducing basis functions on each subdomain. For instance, for a rectangular decomposition of the domain, a polynomial basis can be defined on each subdomain
\begin{equation}
\label{eq:domains_Dk}
\mathcal{D}_k = [M^-_{k,1}, M^+_{k,1}] \times ... \times [M^-_{k,d}, M^+_{k,d}],
\end{equation}
by tensorization of monomial functions of the form
\begin{equation}
e_{k,j,i}(\tht_j) = \left( \frac{\tht_j - M^-_{k,j}}{M^+_{k,j}- M^-_{k,j}}\right)^i.
\label{rescale}
\end{equation}
The integer~$k$ indexes the domain under consideration, $j$~characterizes the degree of freedom under consideration, and~$i$ is the power of the monomial. The normalization we choose ensures that each of the term in the tensor product defining the basis function has values in~$[0,1]$. Defining for instance the degree of the tensor product~${\tt deg}$ to be the maximum of the degree of the individual polynomials, the total number of degrees of freedom is given by $K({\tt deg}+1)^d$. When ${\tt deg}$ is large, the elementary polynomials~\eqref{rescale} assume non negligible values only close to the boundary of the domain, which can result in numerical instabilities. We therefore rescale the basis functions so that their $L^2(\pi)$ norm is~1 (as the normalization factor being estimated using preliminary AdL runs). 

\subsection{Numerical illustration on a mixture of Gaussians}
\label{Mixture}

\begin{figure}
  \centering
  \begin{minipage}{0.49\textwidth}
    \includegraphics[width=\textwidth]{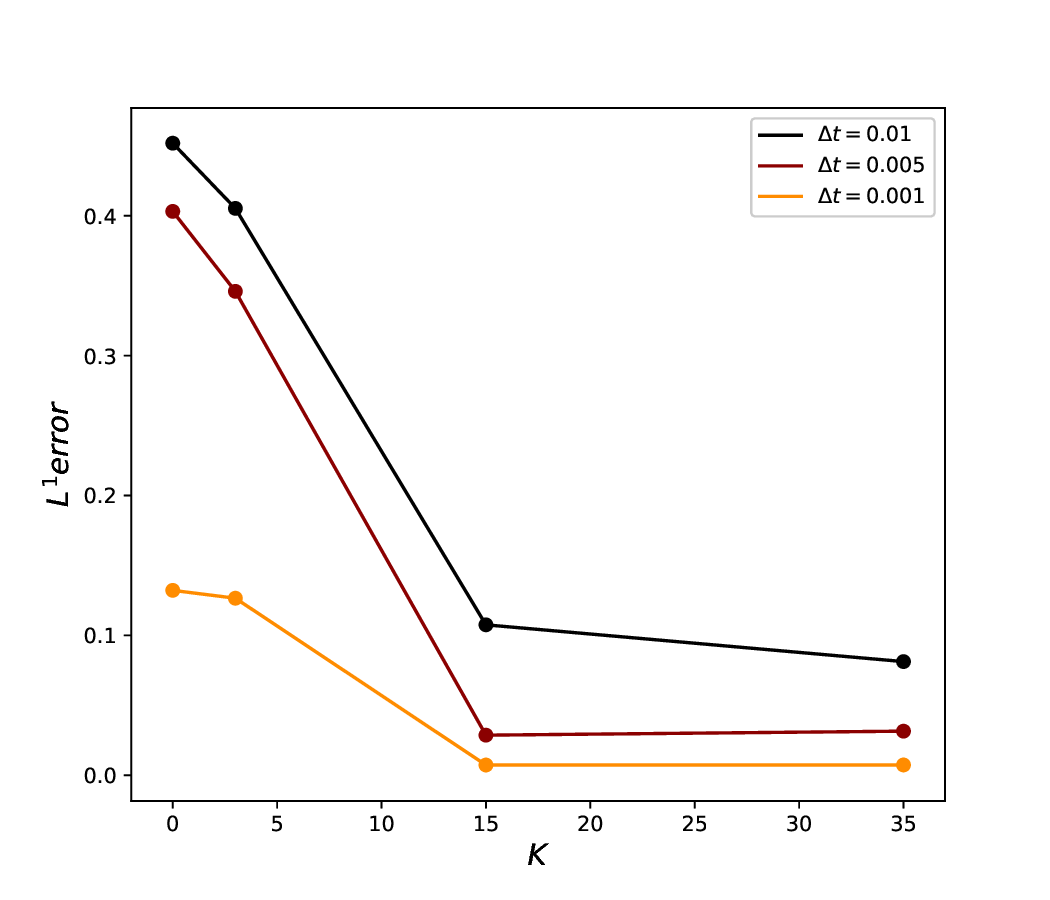}
    \subcaption[second caption.]{$n  =5$}
  \end{minipage}
  \begin{minipage}{0.49\textwidth}
    \includegraphics[width=\textwidth]{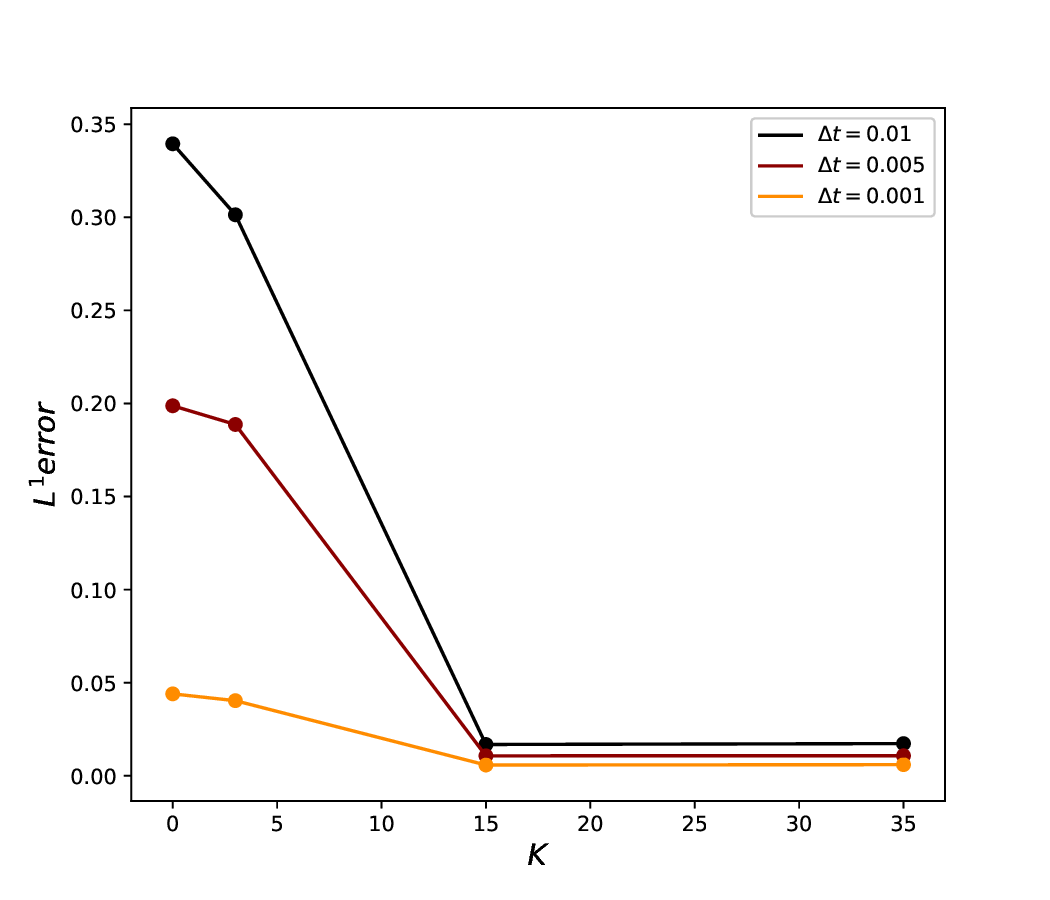}
    \subcaption[second caption.]{$n  =15$}
  \end{minipage}
  \caption{$L^1$ error on the the marginal over $\tht_1$ of the posterior distribution with eAdL for various values of~$K$ and~$\dt$ ($K=0$ corresponds to AdL results).} \label{fig:eadl_mixture1}
\end{figure}

\begin{figure}
  \centering
  \includegraphics[scale=0.6]{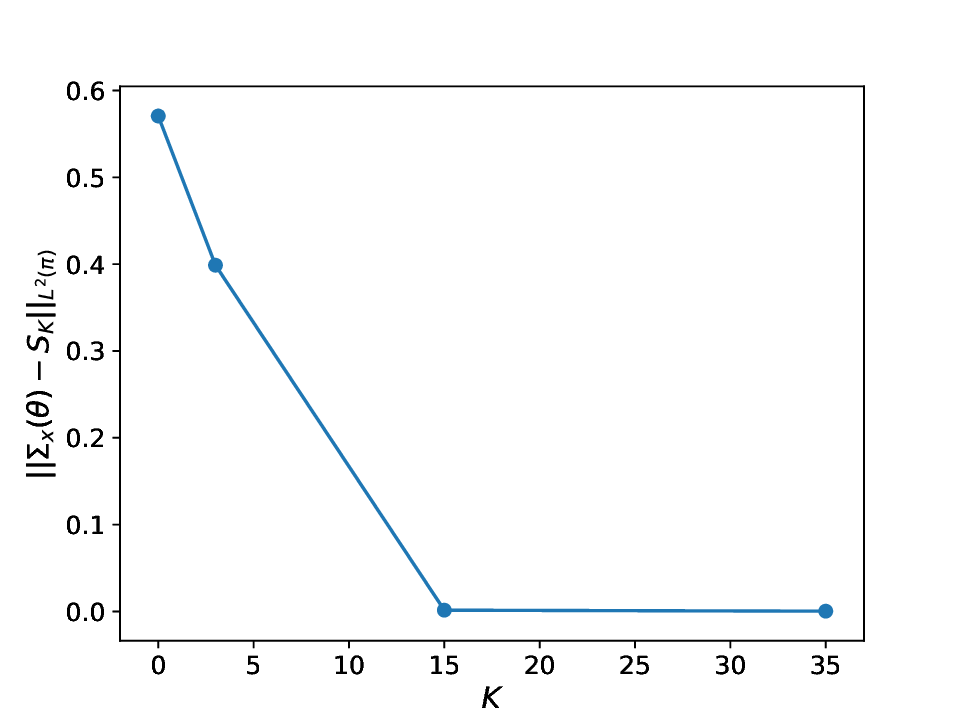}
  \caption{Projection error~$\|\Sigma_{\vectx}- S_K \|_{L^2(\pi)}$, where~$S_K$ is the  $L^2(\pi)$ projection of~$\Sigma_{\vectx}$ onto the vector space of symmetric matrices generated by the basis for each value of $K$.} \label{fig:approx_sigma}
\end{figure}

We present in this section numerical results for the example introduced in Section~\ref{sec:mixture_model}. The covariance matrix in this case is not constant, and AdL fails to correct for the extra noise due to mini-batching; see Section~\ref{mix_adl} and Figure~\ref{fig:AdL_Mixture}. We consider the same parameters as in Section~\ref{mix_adl}. To define the basis functions, we first consider a symmetric rectangular partition on the domain, with 4~meshes of the form~\eqref{eq:domains_Dk}, with 
$M^-_{0,1} = M^-_{0,2} = M^-_{1,2} = M^-_{2,1} = 0$, $M^+_{0,1} = M^+_{0,2} = M^-_{1,1} = M^+_{1,2} = M^+_{2,1} = M^-_{2,2} = M^-_{3,1} = M^-_{3,2} = 0.7$ and $M^+_{1,1} = M^+_{2,2} = M^+_{3,1} = M^+_{3,2}  = 1.4$. We define polynomials on each mesh as discussed in Section~\ref{sec:choice}. We run the numerical scheme~\eqref{eq:sheme_eadl} for eAdL with $\eta_k = 1$ for all values of~$0 \leq k \leq K$, for a final time $T =10^{6}$, various values of~$\Delta t$, and various degrees of polynomials. The case $K=0$ corresponds to standard AdL, while $K=3$ corresponds to an estimation of the covariance matrix by a piecewise constant matrix valued function on the 4~domains under consideration, $K=15$ to products of affine functions in each degree of freedom on the 4~domains, and $K=35$ to products of second order polynomials in each variable on each domain. We fix the size of the minibatch~$n$ to~$5$ and~$15$, which corresponds respectively to $\eps(n) = 7800$ and $\eps(n) = 2466.67$.
We report in Figure~\ref{fig:eadl_mixture1} the $L^1$ error on the marginal distribution over~$\tht_1$ of the posterior distribution sampled by the numerical scheme for various values of $K$. Let us first emphasize that the bias decreases compared to AdL even for the smallest value~$K=3$ (compare with Figure~\ref{fig:AdL_Mixture}). A dramatic decrease is observed for~$K=15$. A small bias remains for~$K \geq 15$, probably mostly due to the time step error. We plot in Figure~\ref{fig:approx_sigma} the quantity $\| \Sigma_{\vectx} - S_K\|_{L^2(\pi)}$ where $S_K$ is the $L^2(\pi)$ projection of~$\Sigma_{\vectx}$ onto the vector space of symmetric matrices generated by the basis for each value of $K$. The key observation here is that the projection error, which quantifies the quality of the approximation of $\Sigma_{\vectx}$, decays similarly to the bias on the invariant measure when~$K$ is increased. This therefore confirms the bias analysis of Section~\ref{sec:disEADL}, which predicts that the bias should be controlled by the projection error.

\section{Discussion and perspectives}
\label{sec:discussion_perspectives}

We quantified in this work the bias on the posterior distribution arising from mini-batching for (kinetic or underdamped) Langevin dynamics, both for standard and adaptive Langevin dynamics. As highlighted in~\eqref{eq:erreur_gen}, the bias is dictated by the quality of the approximation of the covariance matrix~$\Sigma_\vectx(\tht)$ of the gradient estimator by the average (at~$\tht$ fixed) of the extra friction variable~$\xi$ introduced in AdL. The latter average is independent of the parameter~$\tht$ in the original formulation of AdL, and this is sufficient to remove the bias if the target distribution is Gaussian, and allows to strongly reduce the bias for posteriors close to Gaussian (as for the MNIST example considered in Section~\ref{sec:log_reg}). The latter situation is typical when the number of data points~$\Nd$ is large, as the posterior distribution concentrates in a Gaussian manner around one mode (up to symmetries) due to the Bernstein--von Mises theorem (see Section~10.2 in~\cite{vdV98}). In this case, it is likely that AdL already captures most of the mini-batching bias. Our analysis however allows to go one step further, and make precise when a reduction to diagonal, or even isotropic friction matrices, does not degrade the quality of the sampling.

Standard AdL is however not sufficient in various situations, in particular when~$\Sigma_\vectx$ genuinely depends on~$\tht$, which is the case when~$\Nd$ is of moderate size. In this case, the posterior distribution can strongly deviate from a Gaussian distribution. Extended AdL provides a framework to systematically reduce the mini-batching bias in such situations, by adding extra degrees of freedom to better tune the friction variable. Striking reductions in the bias can be observed already for not too many additional degrees of freedom, even for the extreme situation where the batch size is set to its minimal value~$n=1$. In any case, eAdL always leads to a smaller bias than AdL.

Let us also emphasize here that AdL and eAdL can be used in conjunction with other techniques to reduce the variance of the stochastic estimator of the force, such as control variate or extrapolation techniques (see for instance~\cite{articleN17, brosse:hal-01934291, MR3969063, NIPS2016_03f54461}). AdL and eAdL should therefore not be seen as alternatives to these techniques, but as complements.

The choice of the dimensionality of the vector space for the friction variable is a key element for standard AdL and its extension, as this directly impacts the computational cost of the method; see the discussion at the end of Section~\ref{sec:disEADL}. One would like the number of degrees of freedom to be as small as possible. From a theoretical perspective, this motivates further characterizing the structure of the covariance matrix~$\Sigma_\vectx$, which have been observed to be low rank in certain situations related to neural network training; see~\cite{chaudhari2018stochastic}. The low rank structure suggests that most of the variance~$\Sigma_\vectx(\theta)$ can be captured by relying on frictions~$\xi(\theta)$ parametrized by a few degrees of freedom only, potentially much smaller than~$d$. Better characterizing the structure of~$\Sigma_\vectx$ could also shed some light on the current active research efforts on understanding the so-called implicit bias of neural network training.


\subsubsection*{Acknowledgements}The authors thank Ben Leimkuhler and Tiffany Vlaar for stimulating discussions on AdL and Matthias Sachs for kindly providing the code used to perform numerical experiments of Section~\ref{sec:log_reg}. I.~Sekkat gratefully acknowledges financial support from Université Mohammed VI Polytechnique. The work of G.~Stoltz in funded in part by the European Research Council (ERC) under the European Union's Horizon 2020 research and innovation programme (grant agreement No~810367), and by the Agence Nationale de la Recherche, under grants ANR-19-CE40-0010-01 (QuAMProcs) and~ANR-21-CE40-0006 (SINEQ).


\newpage

\appendix
\section{Consistency estimates}
\label{sec:consistency}

We provide in this appendix the algebraic manipulations for various consistency estimates. To simplify the notation, we simply write~$Z$ in all this section for the random variable which appears in~\eqref{eq:hypothese}. The analysis we perform is asymptotic since it relies on the assumption that~$\varepsilon(n) \geq 1$ and~$\dt \leq 1$ are such that~$\varepsilon(n)\dt \leq 1$ is sufficiently small. Let us also recall that the notation~$\mathcal{O}(\dt^{s+1})$ in equalities such as~\eqref{eq:local_consistency_general} is made precise in~\eqref{eq:write_remainders_rigorously}. We explicitly write the first estimates in Section~A.1 using the full notation from~\eqref{eq:write_remainders_rigorously} to give a clear meaning to all equalities, but then return to the notation~$\mathcal{O}(\dt^{s+1})$ for the sake of readability, hoping that the first estimates and~\eqref{eq:write_remainders_rigorously} allow the reader to see how to give a precise meaning to all the equalities we write. 

\subsection{Proof of~\eqref{T_op}}
\label{app_1}

We prove here that the evolution operator for SGLD satisfies~\eqref{T_op} for~$\dt$ and~$\eps(n)\dt$ small. The equality~\eqref{T_op} should be understood as follows: for any given smooth function~$\tht \mapsto \varphi(\tht)$ with compact support, there exist~$\dt_\star>0$ and~$K \in \mathbb{R}_+$ such that, for any~$\dt \in (0,\dt_\star]$, there is a function~$R_{\varphi,n,\dt}$ for which 
\[
\left(\widehat{P}_{\Delta t,n}\varphi\right)(\tht) = \left(\mathrm{e}^{\Delta t (\cL_\mathrm{ovd} + \Delta t \mathcal{A}_{\mathrm{ovd},n} )}\varphi\right)(\tht) + (1+\eps(n)^{3/2})\Delta t^3 R_{\varphi,n,\dt}(\tht),
\]
with
\[
\sup_{\dt \in (0,\dt_\star]} \sup_{1 \leq n \leq \Nd} \sup_{\theta \in \Theta} |R_{\varphi,n,\dt}(\theta)| \leq K.
\]
We rewrite to this end the SGLD scheme~\eqref{SGLD} as
\[
\tht^{m+1} = \Phi_{\dt, n}(\tht^m, Z^m, G^m),
\]
with
\[
\Phi_{\dt, n}(\tht, Z, G) = \tht + \sqrt{\eps(n)} \dt \Sigma^{1/2}_{\vectx}(\tht) Z + \sqrt{2\dt} G + \dt \gradT (\log \pi (\tht|\vectx).
\]
Consider a given smooth function~$\tht \mapsto \varphi(\tht)$ with compact support. Then, there exists~$C \in \mathbb{R}_+$ such that
\begin{equation*}
\begin{alignedat}{1}
&\varphi(\Phi_{\dt, n}(\tht^m, Z^m, G^m)) = \varphi\left(\Phi_{\dt, n}(\tht^m, 0, G^m)+ \sqrt{\eps(n)} \dt \Sigma^{1/2}_{\vectx}(\tht^m) Z^m \right)\\
&\qquad \qquad = \varphi(\Phi_{\dt, n}(\tht^m, 0, G^m))  +\sqrt{\eps(n)} \dt \Sigma^{1/2}_{\vectx}(\tht^m) Z^m \cdot (\gradT\varphi)(\Phi_{\dt, n}(\tht^m, 0, G^m))\\
&\qquad \qquad  \quad + \frac{1}{2} \eps(n)\dt^2 (\gradT^2 \varphi)(\Phi_{\dt, n}(\tht^m, 0, G^m)): \Sigma^{1/2}_{\vectx}(\tht^m)Z^m\otimes  \Sigma^{1/2}_{\vectx}(\tht^m)Z^m\\
& \qquad \qquad\quad + \frac{1}{6} \eps(n)^{3/2} \dt^3 D^3 \varphi(\Phi_{\dt, n}(\tht^m, 0, G^m)): \left(\Sigma^{1/2}_{\vectx}(\tht^m)Z^m\right)^{\otimes 3}\\
&\qquad \qquad \quad + \eps(n)^2\dt^4 \mathscr{R}_{\varphi,n,\dt}(\theta^m,G^m,Z^m),
\end{alignedat}
\end{equation*}
with
\[
\sup_{\dt \in (0,\dt_\star]} \sup_{1 \leq n \leq \Nd} \sup_{\theta \in \Theta} \left|\mathscr{R}_{\varphi,n,\dt}(\theta,G,Z)\right| \leq C \left(1+|Z|^4\right).
\]
In the above expansion in powers of~$\dt$, we use the notation
\[
D^3 \varphi(\tht): v_1 \otimes  v_2 \otimes v_3 = \sum\limits_{i, j, k = 1}^d [v_1]_i [v_2]_j [v_3]_k (\partial^3_{\tht_i, \tht_j, \tht_k} \varphi)(\tht),
\]
and $v^{\otimes 3} = v\otimes   v \otimes v$. Using classical results on the evolution operator of the Euler--Maruyama scheme to compute the expectation of~$ \varphi(\Phi_{\dt, n}(\tht^m, 0, G^m))$ with respect to~$G^m$ (see for example~\cite{JMLR:v17:15-494} for details), we can compute the expectation of~$\varphi(\Phi_{\dt, n}(\tht^m, Z^m, G^m))$ with respect to the variables $G^m$ and $Z^m$ as 
\begin{equation*}
\begin{alignedat}{1}
\left(\widehat{P}_{\dt,n} \varphi\right) (\tht^m) = \mathrm{I}_d & + \dt \cL _{\rm ovd} \phi(\tht^m)+ \dt^2 \left(\frac{1}{2}\cL_{\rm ovd}^2 + \mathcal{A}_{\rm disc}\right)\varphi (\tht^m) \\ 
& + \dt^2 \eps(n) \mathbb{E}_{G}\left[(\mathcal{A}_{\rm mb} \varphi) (\Phi_{\dt, n}(\tht^m, 0, G))\right] + \left(1+\eps(n)^{3/2}\right) \dt^3 \mathcal{R}_{\varphi,n,\dt}(\tht^m),
\end{alignedat}
\end{equation*}
with
\[
\sup_{\dt \in (0,\dt_\star]} \sup_{1 \leq n \leq \Nd} \sup_{\theta \in \Theta} \left|\mathcal{R}_{\varphi,n,\dt}(\tht)\right| < +\infty.
\]
For the end of the proof, for the sake of readability, we no longer make explicit that the operator equalities should be understood as applied to a smooth and compactly supported functions~$\varphi$, with remainder terms multiplying functions of~$\tht$ which are uniformly bounded with respect to~$1 \leq n \leq \Nd$, $\dt \leq \dt_\star$ and~$\tht \in \Theta$. Since $\mathbb{E}_{G}\left[\mathcal{A}_{\rm mb} \varphi(  \Phi_{\dt, n}(\tht^m, 0, G) )\right] = \mathcal{A}_{\rm mb} \varphi(\tht^m) + \mathcal{O}(\dt)$, we obtain on the one hand that
\begin{equation}
\begin{alignedat}{1}
\widehat{P}_{\dt, n} & = \mathrm{I}_d + \dt \cL _{\rm ovd} + \dt^2 \left(\frac{1}{2}\cL_{\rm ovd}^2 + \mathcal{A}_{\rm disc} + \eps(n)\mathcal{A}_{\rm mb}\right) + \mathcal{O}\left((1+\eps(n)^{3/2})\dt^3\right).
\end{alignedat}
\label{eq:P_dev}
\end{equation}
On the other hand, notice that 
\begin{equation}
\begin{alignedat}{1}
\mathrm{e}^{\dt (\cL_{\rm ovd} + \eps(n)\dt \mathcal{A}_{\rm mb} + \dt \mathcal{A}_{\rm disc} )} = \mathrm{I}_d & + \dt \cL_{\rm ovd} + \dt^2\left(\frac{1}{2}\cL_{\rm ovd}^2 + \mathcal{A}_{\rm disc} + \eps(n)\mathcal{A}_{\rm mb}\right)\\
& + \mathcal{O}((1+\eps(n))\dt^3) .
\label{eq:e_dev}
\end{alignedat}
\end{equation}
Using~\eqref{eq:P_dev} alongside with~\eqref{eq:e_dev}, we deduce the desired result~\eqref{T_op}.

\begin{remark}
	\label{rem:erreur_sgld_Z_moment}
	Note that when~$\mathbb{E}[Z^3] = 0$, the remainder term is in fact of order~$\mathcal{O}\left((1+\eps(n)^2\dt) \dt^3\right)$ in~\eqref{eq:P_dev}. This leads to a remainder of order~$\mathcal{O}\left((1+\eps(n)) \dt^3\right)$ in~\eqref{T_op}.
	
\end{remark}
\subsection{Proof of~\eqref{eq:op_Q},~\eqref{eq:op_lan_mini} and~\eqref{eq:widehatP_Langevin_mb}}
\label{app_2}

To prove~\eqref{eq:op_Q}, we consider a smooth function~$(\tht,p)\mapsto \varphi(\tht,p)$ with compact support, and rewrite the operator as
\begin{equation*}
\begin{alignedat}{1}
\left(Q_{\Delta t}^{\cLa}\varphi\right)(\tht^m,p^m)
& = \mathbb{E}\left[\varphi\left(\tht^m,p^m + \Delta t \widehat{F_n}(\tht^m)\right)\right] \\
& = \mathbb{E} \left[ \varphi\left(\tht^m,p^m +\dt \gradT(\log \pi(\tht^m|\vectx)) + \sqrt{\eps(n)} \dt \Sigma_\vectx^{1/2}(\tht^m) Z^m \right) \right].
\end{alignedat}
\end{equation*}
We then have
\[
\begin{alignedat}{1}
&\left(Q_{\Delta t}^{\cLa}\varphi\right)(\tht^m,p^m)  =  \mathbb{E} \left[ \varphi\left(\tht^m,p^m + \dt\gradT(\log \pi(\tht^m|\vectx)\right) \right] \\
&\qquad \qquad \quad +  \sqrt{\eps(n)} \dt \mathbb{E} \left[ \Sigma_\vectx^{1/2}(\tht^m) Z^m \cdot \nabla_p \varphi(\tht^m,p^m + \gradT(\log \pi(\tht^m|\vectx)) \right]\\
&\qquad \qquad  \quad + \frac{1}{2} \eps(n)\dt^2  \mathbb{E} \left[ \nabla_p^2  \varphi(\tht^m,p^m +\dt \gradT(\log \pi(\tht^m|\vectx)):\Sigma^{1/2}_{\vectx}(\tht^m)Z^m  \otimes  \Sigma^{1/2}_{\vectx}(\tht^m)Z^m \right] \\
&\qquad \qquad \quad + \mathcal{O}\left(\eps(n)^{3/2}\dt^3\right),
\end{alignedat}
\]
so that
\begin{equation}
\begin{alignedat}{1}
\left(Q_{\Delta t}^{\cLa}\varphi\right)(\tht^m,p^m)
& = \left(\mathrm{e}^{\dt \cLa} \varphi\right)(\tht^m, p^m)  + \eps(n) \dt^2 (\mathcal{A}_{\rm lan}\varphi)(\tht^m,p^m+ \dt\gradT (\log \pi(\tht^m|\vectx))) \\
& \quad +  \mathcal{O}\left(\eps(n)^{3/2}\dt^3 \right)\\
& = \left(\mathrm{e}^{\dt \cLa} \varphi\right)(\tht^m,p^m)+ \eps(n) \dt^2 \left(\mathcal{A}_{\rm lan}\varphi\right)(\tht^m,p^m)+\mathcal{O}\left(\eps(n)^{3/2}\dt^3\right), 
\end{alignedat}
\label{eq:Q_L1}
\end{equation}
from which the result directly follows.

\begin{remark}
	\label{rem:erreur_Q_non_gauss}
	If $\mathbb{E}[Z^3] = 0$, the error term $\mathcal{O}\left(\eps(n)^{3/2}\dt^3\right)$ can be replaced by $\mathcal{O}\left(\eps(n) \dt^3\right)$. 
\end{remark}

To prove~\eqref{eq:op_lan_mini}, we use the Baker–Campbell–Hausdorff formula on the operator of the numerical scheme~\eqref{GLA_minibatched}:
\begin{equation*}
\begin{alignedat}{1}
\widehat{P}_{\Delta t, n} &= \mathrm{e}^{\Delta t \cLc/2}\mathrm{e}^{\Delta t \cLb/2} Q_{\Delta t}^{\cLa} \mathrm{e}^{\Delta t \cLb/2}\mathrm{e}^{\Delta t \cLc/2}\\
& =  \mathrm{e}^{\Delta t \cLc/2}\mathrm{e}^{\Delta t \cLb/2} \mathrm{e}^{\dt \cLa} \mathrm{e}^{\Delta t \cLb/2}\mathrm{e}^{\Delta t \cLc/2} + \eps(n) \dt^2 \mathrm{e}^{\Delta t \cLc/2}\mathrm{e}^{\Delta t \cLb/2} \mathcal{A}_{\rm lan} \mathrm{e}^{\Delta t \cLb/2}\mathrm{e}^{\Delta t \cLc/2} \\
& \quad + \mathcal{O}\left(\eps(n)^{3/2}\dt^3 \right) \\
& = \mathrm{e}^{\dt \cL_{\rm lan}} + \mathcal{O}\left((\dt + \eps(n))\dt^2\right).
\end{alignedat}
\end{equation*}
Using computations similar to the ones leading to~\eqref{T_op}, we can prove the following estimate on~$Q_{\Delta t}^{\cLa}$: 
\begin{equation}
\begin{alignedat}{1}
\mathrm{e}^{\dt (\cLa + \eps(n)\dt \mathcal{A}_{\rm lan})} = Q_{\Delta t}^{\cLa} + \mathcal{O}\left((1+\eps(n)^{3/2})\dt^3\right).
\end{alignedat}
\label{eq:estim_Q}
\end{equation}
The desired result~\eqref{eq:widehatP_Langevin_mb} follows by using the BCH formula.

\subsection{Proof of~\eqref{eq:P_adL_erreur}}
\label{app_3}

The evolution operator of the numerical scheme of AdL is given by 
\begin{equation*}
\begin{alignedat}{1}
\widehat{P}_{\Delta t, n} &= \mathrm{e}^{\Delta t \cLd/2}\mathrm{e}^{\Delta t \cLc/2}\mathrm{e}^{\Delta t \cLb/2} Q_{\Delta t}^{\cLa} \mathrm{e}^{\Delta t \cLb/2}\mathrm{e}^{\Delta t \cLc/2}\mathrm{e}^{\Delta t \cLd/2},
\end{alignedat}
\end{equation*}
where~$\cLa = \gradT \log(\pi(\cdot|\vectx))^T \nabla_p$ and~$\cLb$ are respectively the generators of~\eqref{eq:elementary_SDE_mb} when~$\eps(n) = 0$ and~\eqref{eq:elementary_SDE_tht} (which coincide with the operators in~\eqref{eq:L_op}), while~$\cLc$ and~$\cLd$ are the generators of~\eqref{eq:elementary_SDE_xi} and~\eqref{eq:elementary_SDE_OU}. Replacing $Q_{\Delta t}^{\cLa}$ by its expression in~\eqref{eq:Q_L1} and using the BCH formula, we obtain, by computations similar to the ones leading to~\eqref{eq:estim_Q}:
\begin{equation*}
\begin{alignedat}{1}
\widehat{P}_{\Delta t, n}\varphi
& = \mathrm{e}^{\Delta t \cLd/2}\mathrm{e}^{\Delta t \cLc/2}\mathrm{e}^{\Delta t \cLb/2} \mathrm{e}^{\Delta t \cLa} \mathrm{e}^{\Delta t \cLb/2}\mathrm{e}^{\Delta t \cLc/2}\mathrm{e}^{\Delta t \cLc/2} +\eps(n) \dt^2 \mathcal{A}_{\rm lan}\varphi + \mathcal{O}\left(\eps(n)^{3/2}\dt^3\right) \\
& = \mathrm{e}^{\Delta t (\cLa + \cLb + \cLc + \cLd)} \varphi +\eps(n) \dt^2 \mathcal{A}_{\rm lan}\varphi + \mathcal{O}\left((1+\eps(n)^{3/2})\dt^3\right) \\
& = \mathrm{e}^{\Delta t (\cLa + \cLb + \cLc + \cLd +\eps(n) \dt  \mathcal{A}_{\rm lan} )}\varphi + \mathcal{O}\left(\eps(n)\dt^3\right) + \mathcal{O}\left((1+\eps(n)^{3/2})\dt^3 \right) \\
& = \mathrm{e}^{\Delta t \cL_{\rm AdL, \Sigma_{\vectx} }}\varphi + \mathcal{O}\left((1+\eps(n)^{3/2})\dt^3\right).
\end{alignedat}
\end{equation*}
When $\mathbb{E}[Z^3] = 0$, the error term becomes~$\mathcal{O}\left((1+\eps(n))\dt^3\right)$.

\section{Unbiasedness of the mean of Gaussian distributions for Langevin dynamics}
\label{app:unbiased_Langevin}

We prove that, in the case of one dimensional Gaussian likelihoods, there is no bias on the mean of the posterior distribution when using the discretization of Langevin dynamics~\eqref{GLA_minibatched} (see Section~\ref{sec:gaussian_model}). We start by rewriting the numerical scheme as 
\begin{equation*}
\begin{pmatrix}
\tht^{m+1} \\
p^{m+1}
\end{pmatrix}
=
M_1 
\begin{pmatrix}
\tht^{m} \\
p^{m}
\end{pmatrix}
+ \left( 1-\alpha_{\Delta t} \right)^{1/2} M_2 
\begin{pmatrix}
G^{m} \\
G^{m+1/2}
\end{pmatrix}
+ V_3^m,
\end{equation*}
where
\begin{equation*}
M_1 = 
\begin{pmatrix}
\dps 1-a  \frac{\Delta t^2}{2}  & \dps \alpha_{\Delta t/2} \left( \Delta t  - a\frac{\Delta t^3}{4}\right)  \\[4pt]
-a \dt \alpha_{\dt/2} & \dps \alpha_{\dt}\left(1 - a \frac{\Delta t^2}{2}\right)  
\end{pmatrix}
, \quad
M_2 =  
\begin{pmatrix}
\dps \Delta t \left(1- a\frac{\Delta t^2}{4}\right) & 0\\[4pt]
\dps \alpha_{\dt/2}\left(1 - a \frac{\Delta t^2}{2}\right)  & 1
\end{pmatrix},
\end{equation*}
\begin{equation*}
V_3^m = 
\begin{pmatrix}
\dps\frac{\dt^2}{2}b^m\\
\alpha_{\dt/2} \dt b^m
\end{pmatrix},
\end{equation*}
and
\[
a = \frac{1}{ \sigma_\tht^2} +  \frac{\Nd}{ \sigma_x^2} ,
\qquad
b^m =  \frac{\Nd}{n}\sum\limits_{i\in I_n^m} \frac{x_i}{\sigma_x^2}.
\]
Since
\[
\mathbb{E}[b^m] = \dps\frac{1}{\sigma_x^2} \sum_{i = 1}^{\Nd} x_i := b,
\]
we obtain, by first taking expectations with respect to $G^m$ and $G^{m+1/2}$, and then with respect to realizations of~$I_n^m$ in~$b^m$:
\begin{equation}
\mathbb{E}\left[\begin{pmatrix}
\tht^{m} \\
p^{m}
\end{pmatrix}\right]
=
M_1^m
\mathbb{E}\left[\begin{pmatrix}
\tht^{0} \\
p^{0}
\end{pmatrix}\right]
+ \sum_{j=0}^{m-1} M_1^j V_3^j,
\label{eq:schema_matr_GLA}
\end{equation}
with~$V_2 = (\dt^2 b/2,\alpha_{\dt/2} \dt b)^T$.
We note that (recalling that, for the one dimensional case considered here, the friction~$\Gamma>0$ is a scalar)
\[
M_1 = \mathrm{I}_d - \dt \begin{pmatrix}
0  & -1 \\
a & \Gamma  
\end{pmatrix}
+ \mathcal{O}(\dt^2),
\]
which shows that the eigenvalues of~$M_1$ have real parts which are strictly smaller than~1 provided~$\dt$ is sufficiently small.
Therefore, $M_1^m \to 0$ as~$m \to +\infty$, and the matrix~$\mathrm{I}_d - M_1$ is invertible.
Moreover, a simple computation shows that 
\[
(\mathrm{I}_d - M_1)\begin{pmatrix}
b/a\\
0
\end{pmatrix} = V_3,
\]
so that
\[
(\mathrm{I}_d - M_1)^{-1} V_3 = \begin{pmatrix}
b/a\\
0
\end{pmatrix}.
\]
By letting~$m$ go to infinity in~\eqref{eq:schema_matr_GLA}, and using the above equality, we finally conclude that
\[
\lim_{m \to +\infty} \mathbb{E}\left[\begin{pmatrix}
\tht^m \\
p^m
\end{pmatrix}\right]
=
\begin{pmatrix}
b/a\\
0
\end{pmatrix} = \begin{pmatrix}
\mu_{\rm post}\\
0
\end{pmatrix},
\]
which shows that the mean posterior distribution is unbiased.

\vskip 0.2in
\bibliographystyle{plain}
\bibliography{biblio_MD}

\end{document}